\documentclass[12pt]{article}

\usepackage[utf8]{inputenc}
\usepackage{amsmath,amssymb}
\usepackage{graphicx}
\usepackage{booktabs}
\usepackage{multirow}
\usepackage{xcolor}
\usepackage{hyperref}
\usepackage[margin=1in]{geometry}
\usepackage{caption}
\usepackage{subcaption}
\usepackage{float}
\usepackage{authblk}

\newcommand{\anubuddhi}{A\d{n}ubuddhi}  

\begin{document}

\title{\anubuddhi: A Multi-Agent AI System for Designing and Simulating Quantum Optics Experiments}

\author[1,2]{S. K. Rithvik}
\affil[1]{Quantum Science and Technology Laboratory, Physical Research Laboratory, Navrangpura, Ahmedabad 380009, India}
\affil[2]{Indian Institute of Technology Gandhinagar, Palaj, Gandhinagar 382055, India}

\date{\today}

\maketitle

\vspace{-0.5cm}

\begin{abstract}
\small
We present \anubuddhi, a multi-agent AI system that designs and simulates quantum optics experiments from natural language prompts without requiring specialized programming knowledge. The system composes optical layouts by arranging components from a three-tier toolbox via semantic retrieval, then validates designs through physics simulation with convergent refinement. The architecture combines intent routing, knowledge-augmented generation, and dual-mode validation (QuTiP and FreeSim). We evaluated 13 experiments spanning fundamental optics (Hong-Ou-Mandel interference, Michelson/Mach-Zehnder interferometry, Bell states, delayed-choice quantum eraser), quantum information protocols (BB84 QKD, Franson interferometry, GHZ states, quantum teleportation, hyperentanglement), and advanced technologies (boson sampling, electromagnetically induced transparency, frequency conversion). The system achieves design-simulation alignment scores of 8--9/10, with simulations faithfully modeling intended physics. A critical finding distinguishes structural correctness from quantitative accuracy: high alignment confirms correct physics architecture, while numerical predictions require expert review. Free-form simulation outperformed constrained frameworks for 11/13 experiments, revealing that quantum optics diversity demands flexible mathematical representations. The system democratizes computational experiment design for research and pedagogy, producing strong initial designs users can iteratively refine through conversation.
\end{abstract}

\section{Introduction}

The automated design of quantum experiments has evolved from a theoretical curiosity to a practical necessity. Quantum optics experiments—involving entangled photon sources, interferometric setups, and quantum state characterization—demand years of specialized training. An experimentalist must understand quantum mechanics, navigate hundreds of optical components, validate whether proposed setups produce intended quantum states, and iteratively refine designs through trial and error. This complexity has motivated decades of research into computational methods that can assist or even automate the design process.\footnote{Corresponding author: S. K. Rithvik (rithvik\_ks@iitgn.ac.in)}

Early pioneering work demonstrated that machine learning could discover experimental configurations beyond human intuition. MELVIN (2016-2018), developed by Krenn and colleagues at the University of Innsbruck, represented quantum experiments as weighted graphs and used automated search algorithms to discover novel experimental implementations~\cite{krenn2016automated}. This breakthrough revealed a ``hidden bridge'' between graph theory and quantum optics, enabling the discovery of previously unknown experimental schemes for generating complex entangled states. The subsequent development of THESEUS further refined this graph-based approach, producing simpler experimental configurations through improved search heuristics~\cite{krenn2020automated}.

Building on these graph-theoretic foundations, PyTheus (2023) introduced a more general digital discovery framework~\cite{ruizgonzalez2023pytheus}. PyTheus automated the design of diverse quantum experiments, discovering over 100 experimental configurations for entangled state generation and measurement schemes. Its inverse-design algorithm proved particularly powerful for finding minimal experimental implementations of target quantum operations. Recent work has extended PyTheus through a modular interpreter that automates the analysis and visualization of PyTheus-optimized quantum networks~\cite{sreekantham2025modular}, handling complex connectivity patterns and facilitating the translation of abstract graph representations into implementable experimental setups.

Parallel to these quantum-specific developments, AdaQuantum (2019-2020) demonstrated that hybrid genetic algorithm and deep neural network approaches could optimize experimental parameters for quantum state engineering~\cite{adaquantum2020}. AdaQuantum's key innovation was using evolutionary algorithms to explore the vast design space while employing neural networks to predict experimental outcomes, significantly reducing the computational cost of optimization. This work showed that combining classical optimization with machine learning could accelerate the discovery of experimental protocols for producing non-classical quantum states of light.

More recently, the emergence of large language models (LLMs) has catalyzed a new generation of AI-driven scientific discovery systems. Coscientist (2023), developed by Boiko et al., demonstrated that GPT-4 could autonomously design, plan, and execute complex chemistry experiments~\cite{boiko2023autonomous}. The system successfully navigated hardware documentation, controlled laboratory instruments, and optimized experimental parameters—showing that LLMs could handle the full experimental workflow from planning to execution. Systematic evaluations of LLM capabilities on quantum mechanics problem-solving have revealed both strengths and limitations: recent studies benchmarking 15 LLMs across quantum derivations, creative problems, and numerical computation found that while flagship models achieve high accuracy on theoretical tasks, numerical computation remains challenging even with tool augmentation~\cite{sreekantham2025llmquantum}. In quantum physics specifically, the k-agents framework (2024) introduced LLM-based agents for automating quantum computing laboratory experiments~\cite{cao2024kagents}. K-agents organizes laboratory knowledge and coordinates multiple specialized agents to perform calibrations and characterizations of quantum processors, demonstrating successful single-qubit and two-qubit gate calibrations from natural language instructions.

The AI-Mandel system (2024), developed by Arlt, Gu, and Krenn, represents a particularly significant advance: an LLM agent that generates and implements original research ideas in quantum physics~\cite{arlt2024aimandel}. AI-Mandel formulates hypotheses in natural language, automatically implements them using domain-specific tools like PyTheus, and has contributed to published research including the discovery of novel quantum teleportation variants. Similarly, Robin (2025), a multi-agent system by Ghareeb et al., demonstrated the first fully autonomous discovery and validation of a novel therapeutic candidate through iterative lab-in-the-loop experimentation~\cite{ghareeb2025robin}, showcasing the potential of multi-agent architectures for scientific discovery. The HoneyComb system (2024) further demonstrated flexible LLM-based agents specifically designed for materials science, leveraging high-quality domain knowledge bases and tool-calling capabilities~\cite{zhang2024honeycomb}.

Despite these remarkable advances, a fundamental limitation persists across nearly all existing systems: they require non-intuitive interaction methods that create barriers to adoption. MELVIN and PyTheus demand understanding of graph-theoretic representations and topological operations. AdaQuantum requires expertise in genetic algorithm parameter tuning and fitness function design. Even recent LLM-based systems like k-agents require familiarity with specific experimental frameworks and structured command formats. This interaction complexity means that researchers must invest significant time learning the system's interface before they can leverage its capabilities—ironically recreating a training barrier that automation was supposed to eliminate.

Consider a typical workflow with PyTheus: a researcher must manually construct graph representations of quantum experiments, understand the meaning of edges and vertices in the context of photon paths and transformations, configure search parameters for the optimization algorithm, interpret the output graphs, and then translate those abstract representations back into physical experimental configurations. While the system excels at discovering novel designs, each step requires specialized knowledge that most quantum physicists lack unless they specifically study the tool. Similarly, AdaQuantum demands understanding how to encode experimental constraints as fitness functions, tune genetic algorithm hyperparameters like mutation rates and population sizes, and interpret the evolutionary convergence patterns. These are valuable research tools for computational specialists, but they create friction for experimental physicists who simply want to design better experiments.

This gap between capability and usability defines the current state of automated experiment design. We have systems that can discover brilliant experimental configurations and even generate novel scientific ideas, but accessing those capabilities requires mastering complex intermediate representations, programming interfaces, or domain-specific languages. The promise of AI-assisted science—that domain experts could directly leverage computational intelligence without becoming computer scientists—remains largely unfulfilled.

\textit{Aṇubuddhi} (Sanskrit: ``atomic intelligence'') addresses this fundamental challenge through a natural language interface that eliminates intermediate representations entirely. A quantum physicist can simply describe what they want—``Design a Hong-Ou-Mandel interferometer to measure photon indistinguishability''—and receive a complete experimental design with component specifications, quantum state predictions, and validation results. No graphs to construct, no algorithm parameters to tune, no programming required. The system handles the full cognitive workflow: interpreting intent, retrieving relevant knowledge from past experiments, generating physically realistic designs, validating through quantum simulations, and learning from successes to improve future performance. By combining LLMs for reasoning and code generation with structured knowledge management through vector databases and physics simulation engines, Aṇubuddhi creates an architecture that mirrors how expert experimentalists think and work—but makes that expertise accessible through conversation.

The key innovation is not simply adding a natural language interface to existing tools, but implementing a complete learning architecture where validated experiments become reusable building blocks. When a user approves a Bell state generation setup, that entire assembly—pump laser, nonlinear crystal, filters, detectors—becomes a learned composite stored in a semantic vector database. Future requests for entangled photon sources can instantly retrieve and adapt this validated design rather than regenerating from scratch. This procedural learning mimics how research groups accumulate expertise: successful setups document in lab notebooks and inform new projects. Aṇubuddhi scales this process through semantic search over potentially hundreds of learned patterns, enabling instant retrieval of relevant past experiments while maintaining conversational flexibility for novel designs. The system has successfully designed 13 canonical quantum optics experiments spanning single-photon interference, quantum entanglement, quantum cryptography, and quantum state tomography—all from single-line natural language prompts such as ``Design a Hong-Ou-Mandel interferometer'' or ``Design a BB84 QKD system,'' with no additional specifications required.

This paper presents the architecture, implementation, and evaluation of Aṇubuddhi across diverse experimental scenarios, demonstrating that natural language interaction need not sacrifice technical rigor or design quality. We show that by carefully structuring the cognitive workflow—from conversational intent routing through knowledge-augmented generation to dual-mode validation with convergent self-refinement—AI systems can make sophisticated experiment design accessible to researchers without computational backgrounds. Our results suggest that the future of AI-assisted science lies not in creating more powerful but less accessible tools, but in building systems that meet researchers where they are: in natural language conversation about the science itself.

\section{Methods}

This section details the three-layered cognitive architecture underlying \anubuddhi, from conversational intent routing through knowledge-augmented generation to physics validation with convergent self-refinement. We describe each layer's design rationale, implementation approach, and how the components integrate to transform conversational prompts into physically validated optical configurations. The validation methodology for evaluating system performance across 13 diverse quantum optics experiments is presented at the end.

\subsection{Architectural Overview}

\begin{figure}[H]
\centering
\includegraphics[width=0.95\textwidth]{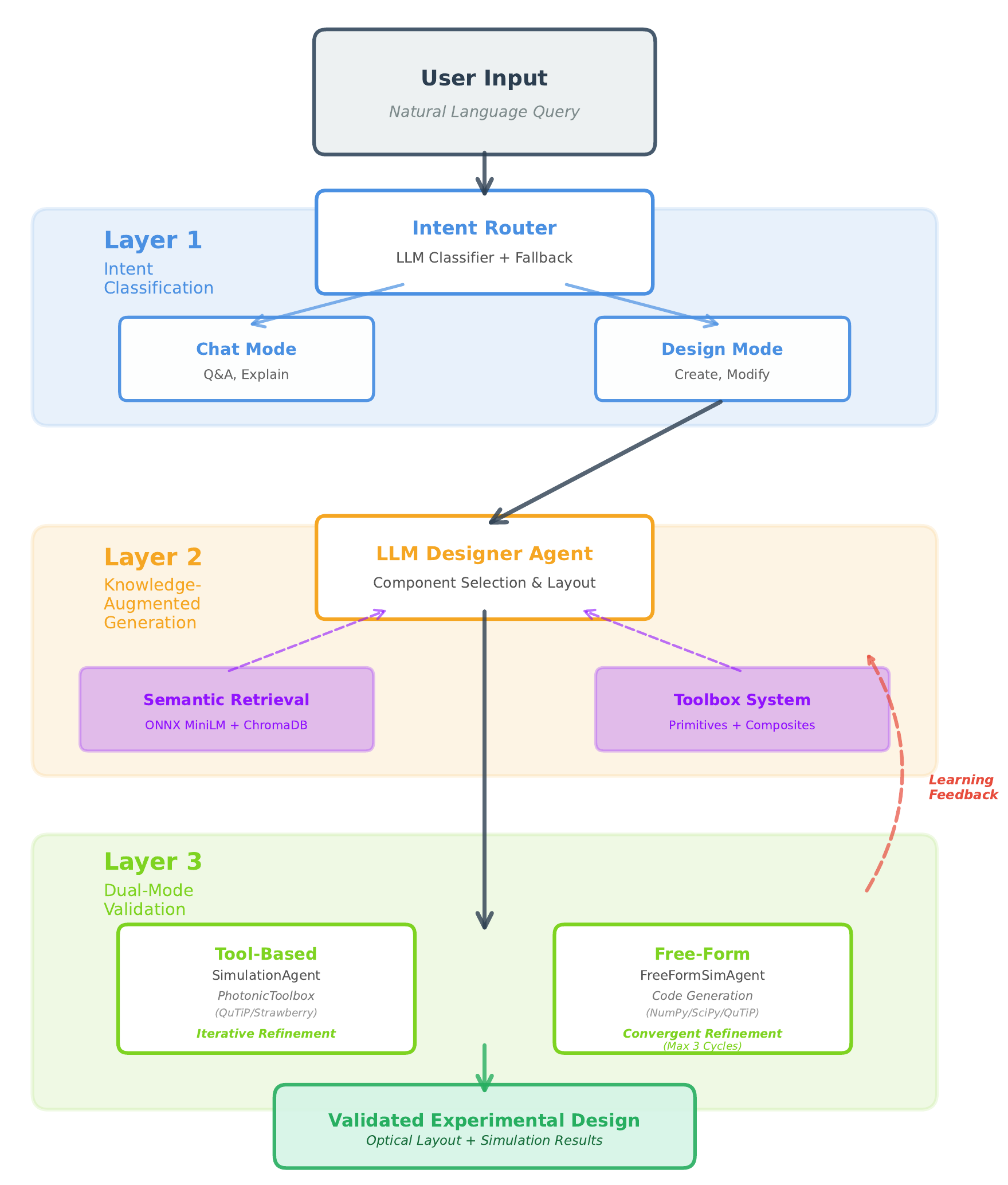}
\caption{\textbf{Three-Layer Cognitive Architecture of Aṇubuddhi.} The system processes natural language requests through three hierarchical layers mirroring expert experimentalist workflow. Layer 1 uses LLM-based intent routing to distinguish questions from design modifications. Layer 2 implements retrieval-augmented generation~\cite{lewis2020retrieval} through semantic search over learned composites. Layer 3 validates designs via dual-mode simulation with convergent self-refinement~\cite{madaan2023selfrefine}. Approved designs feed back as reusable building blocks, implementing experience-based procedural learning.}
\label{fig:architecture}
\end{figure}

Aṇubuddhi implements a three-layer cognitive architecture (Figure~\ref{fig:architecture}) that addresses the fundamental challenges of autonomous experiment design: conversational interaction, knowledge integration, and physics validation. Unlike graph-based systems that require users to construct explicit connectivity patterns~\cite{krenn2020automated,ruizgonzalez2023pytheus}, Aṇubuddhi operates entirely through natural language, translating conversational requests into validated experimental designs. The architecture emerged from systematic analysis of failure modes in preliminary implementations, leading to the hierarchical separation of concerns that enables robust operation across diverse experimental domains.

\subsection{Layer 1: Conversational Intent Routing}

Scientific collaboration requires fluid alternation between discussing concepts and modifying designs. During an experiment design session, users naturally interleave physics questions (``Why do indistinguishable photons bunch?''), practical inquiries (``Where can I buy a BBO crystal?''), and design modifications (``Add a delay stage to the second arm''). A naive system regenerating the entire experiment for every input would destroy conversational context; conversely, a system never modifying designs would be useless for iterative refinement.

Aṇubuddhi solves this through LLM-based intent classification~\cite{yao2023react} implemented in \texttt{llm\_designer.py}. For each user message, the system constructs a routing prompt contrasting CHAT intent (information requests, explanations, discussions) against DESIGN intent (create, modify, add, remove commands). The LLM returns a single-word classification, enabling contextual interpretation beyond keyword matching: ``Can I use a different laser?'' routes to CHAT (asking about possibilities) while ``Use a different laser'' routes to DESIGN (commanding a change).

When LLM classification fails due to API errors or genuinely ambiguous phrasing, a keyword-based fallback router provides defensive coverage, defaulting to CHAT mode when uncertainty exists to prevent accidental design destruction. This dual-mode approach (primary LLM routing with fallback heuristics) ensures system robustness: the LLM handles semantic nuance while keyword matching prevents complete failure. The implementation preserves conversational flow—users can ask ``Why indistinguishable photons?'' and receive physics explanations without losing their carefully constructed optical table—while enabling precise control when modifications are explicitly requested.

\subsection{Layer 2: Knowledge-Augmented Generation with Semantic Retrieval}

The second layer addresses a fundamental tension in AI-assisted design: how to leverage accumulated knowledge without constraining creativity. A system with no memory regenerates experiments from first principles each time, wasting computation and producing inconsistent results. A system that rigidly reuses past designs cannot handle novel requirements. Aṇubuddhi implements retrieval-augmented generation (RAG)~\cite{lewis2020retrieval} to balance these extremes, but adapts the approach for experimental design through a three-tier knowledge hierarchy and semantic retrieval over growing composite libraries.

\subsubsection{Three-Tier Knowledge Hierarchy}

The system maintains three distinct component types in \texttt{toolbox\_loader.py}, each serving complementary roles in experiment construction:

\textbf{Tier 1—Primitives} (\texttt{primitives.json}): Over 50 immutable basic components form the foundation vocabulary. These include light sources (lasers, LEDs, single-photon sources), passive optics (mirrors, beam splitters, waveplates, lenses), nonlinear crystals (BBO, KTP, PPLN for frequency conversion and SPDC), detectors (photodiodes, APDs, SPADs, coincidence counters), and modulators (EOM, AOM, spatial light modulators). Organized by category (sources, detectors, crystals, passive\_optics, measurement, modulation), this library never changes and provides the basic building blocks that combine to form any optical experiment.

\textbf{Tier 2—Learned Composites} (\texttt{learned\_composites.json}): User-approved experimental assemblies transition from one-off implementations to reusable building blocks. When a user validates a complete SPDC photon pair source (pump laser at 405nm + BBO crystal with type-II phase matching + dual interference filters at 810nm + spatial mode coupling optics + SPADs with coincidence counting), the entire assembly saves as a named composite with complete specifications: component positions, beam paths, physics explanation, and approval timestamp. Subsequently, when anyone requests ``entangled photon source'' or ``Bell state generator'', semantic search surfaces this validated pattern. This is the primary learning mechanism—successful multi-component assemblies transition from ephemeral experiments to institutional knowledge, implementing procedural learning where the system improves with use.

\textbf{Tier 3—Custom Components} (\texttt{custom\_components.json}): Specialized equipment for non-standard experiments (rubidium vapor cells for EIT, optical frequency combs, atom chips, custom coincidence electronics) that LLMs define when needed. These track usage counts and last-used dates to prevent redundant redefinition. If an LLM defined a ``87Rb vapor cell with temperature control and buffer gas'' three experiments ago, subsequent requests can reference that definition rather than generating it anew, reducing token usage and ensuring consistency.

\subsubsection{Semantic Retrieval Over Learned Patterns}

As the learned composite library grows to hundreds of entries, including everything in the generation prompt becomes infeasible—modern LLMs have token limits, and overwhelming context degrades reasoning quality. The solution is embedding-based semantic search~\cite{lewis2020retrieval} implemented in \texttt{embedding\_retriever.py}.

When a user requests an experiment (e.g., ``Design Hong-Ou-Mandel interference''), Aṇubuddhi embeds this query using ONNX Runtime's implementation of MiniLM-L6-V2 sentence transformer (22M parameters, 384-dimensional output vectors). This embedding vector is compared against all indexed composites in ChromaDB~\cite{chroma2023} using cosine similarity. The retrieval includes intelligent deduplication: when multiple versions of the same experiment exist (e.g., ``Mach-Zehnder v1'' from initial design, ``Mach-Zehnder v2'' from user refinement, ``Mach-Zehnder v3'' from different parameter choices), only the most recent version from each name group is retained. The top-5 most semantically similar composites appear in the generation context, providing the LLM with concrete examples of validated designs addressing similar physics.

Given these retrieved patterns, the LLM makes a critical decision: does an existing composite closely match the current request ($>$80\% similarity in description and components)? If yes, the system returns metadata indicating an existing match, and the UI presents three options to the user: (1) \textit{Use This}—deploy the validated design exactly, ensuring consistency with past work; (2) \textit{Auto-Improve}—let the LLM adapt the composite to address specific user requirements while preserving the validated core structure; or (3) \textit{Generate New}—ignore the match and create a fresh design, useful when users want to explore alternative approaches. If no close match exists ($<$80\% similarity), the system generates a novel design but uses retrieved composites as contextual examples demonstrating component usage patterns, typical configurations, and design strategies that have previously succeeded.

This implements \textit{smart reuse} that balances efficiency with flexibility: exact matches leverage validated designs for consistency and speed, while partial matches provide inspiration without constraining the LLM's ability to address novel requirements. The approach mimics how experimentalists work in practice—checking lab notebooks for similar past setups before designing from scratch, adapting existing protocols when appropriate, but creating new configurations when needed for genuinely novel experiments.

\subsection{Layer 3: Dual-Mode Physics Validation with Convergent Self-Refinement}

Generated optical layouts must be physically correct before deployment. A design with incorrect polarization alignment, wrong detector placement, or incompatible component specifications wastes experimental resources and user time. Aṇubuddhi validates every generated design through quantum simulation, but faces a significant technical challenge: quantum optics experiments span multiple mathematical formalisms. Interferometry uses Fock state representations, Hong-Ou-Mandel interference requires temporal wavepacket propagation, squeezed light demands quadrature operator calculations, and electromagnetically induced transparency necessitates Lindblad master equations. No single simulation framework adequately handles all domains.

\subsubsection{Dual-Mode Validation Strategy}

The system implements two complementary validation modes, allowing users to run experiments in either approach to empirically compare results:

\textbf{Mode 1—PhotonicToolbox (Discrete Photonics)} provides high-reliability validation for standard experiments. Built on Strawberry Fields~\cite{kiloran2019strawberry} and QuTiP~\cite{johansson2012qutip} backends, this mode offers type-safe operations: beam splitters act on Fock states with specified transmittivity, waveplates rotate polarization angles, photodetectors measure photon number probabilities. The structured API prevents nonsensical operations (e.g., applying beam splitter to atomic state) and guarantees physically valid transformations. This works excellently for Mach-Zehnder interferometry, Bell state generation via SPDC, basic polarization experiments, and other systems naturally described by discrete quantum states with photon-number-based observables.

However, PhotonicToolbox cannot model temporal dynamics (photon arrival time distributions, distinguishability from pulse timing), continuous-variable systems (quadrature measurements, Wigner functions, phase-space distributions), or atomic physics (multilevel coherences, spontaneous emission, optical pumping). The Fock state formalism is fundamentally inadequate for these phenomena—attempting to model Hong-Ou-Mandel interference with static photon number states misses the essential physics of temporal overlap.

\textbf{Mode 2—Free-Form Simulation (Flexible Physics Modeling)} addresses this limitation by giving the LLM complete freedom to write simulation code from scratch, implemented in \texttt{freeform\_simulation\_agent.py}. For each experiment, the agent chooses the appropriate formalism: temporal wavepacket propagation with Gaussian mode overlap integrals for Hong-Ou-Mandel interference, quadrature operator calculations and homodyne detection statistics for squeezed light, Lindblad master equations for electromagnetically induced transparency. The agent has full access to NumPy (array operations, linear algebra, FFTs), SciPy (special functions, numerical integration, optimization), QuTiP (quantum operators, time evolution, master equations), and Matplotlib (visualization). This flexibility enables accurate modeling of diverse quantum systems but introduces a critical reliability challenge: \textit{how to ensure auto-generated code actually simulates the intended experiment rather than superficially similar physics?}

\subsubsection{Convergent Self-Refinement with Multi-Gate Validation}

To address code generation reliability, the free-form mode implements a six-stage validation pipeline inspired by self-refinement~\cite{madaan2023selfrefine} and self-debugging~\cite{chen2023teaching} approaches, but adapted for physics simulation with explicit alignment checking. The process iterates up to three times, converging toward simulations that accurately model designed experiments:

\textbf{Stage 1—Physics Classification}: The LLM categorizes each experiment into one of four physics domains (discrete photonic, temporal, continuous variable, atomic) based on experiment description and key components. This guides formalism selection: ``HOM interference with photon distinguishability'' → temporal domain → wavepacket propagation code; ``squeezed light homodyne detection'' → continuous variable → quadrature operators; ``multilevel atom Rabi oscillations'' → atomic → master equation. The classification prompt explicitly asks \textit{what physical quantities determine the experimental outcome}, focusing the LLM on relevant observables rather than superficial experiment names.

\textbf{Stage 2—Guided Code Generation}: Initial simulation code generation uses a comprehensive 4800-character prompt specifying available libraries and their typical usage, realistic parameter ranges (wavelengths 200--2000nm, squeezing $<$15dB, detector efficiencies 0.1--0.95), physics-specific guidance (e.g., for temporal domain: ``Model photons as Gaussian wavepackets with temporal width, compute overlap integral at beam splitter following the Dirac formalism''), and common pitfalls to avoid (wrong Bell state phase conventions, using static Fock states for inherently temporal phenomena, incorrect SPDC phase matching conditions, unrealistic parameter values like 68dB squeezing). The prompt incorporates chain-of-thought reasoning~\cite{wei2022chain} by explicitly requesting the LLM to explain its modeling choices before generating code, improving physical accuracy.

\textbf{Stage 3—Pre-Execution Review}: Before running any code, an LLM-based review evaluates whether it will actually model the intended experiment. This pre-execution check answers four critical questions: (1) Does the chosen formalism match the experiment type? (temporal wavepackets for HOM, not static Fock states); (2) Are key optical components present in simulation? (beam splitter at correct position, detectors at appropriate outputs); (3) Are parameters physically realistic? (810nm wavelength for SPDC, not 10nm; squeezing 3--10dB, not 68dB); (4) Is the mathematical approach fundamentally sound? (proper normalization, unitary evolution, probability conservation). The LLM returns a structured response \texttt{\{``approved'': bool, ``confidence'': float, ``concerns'': [list]\}}. Code failing this review proceeds directly to refinement without wasting compute on execution of obviously flawed code. This pre-check prevents the common failure mode where syntactically correct but physically nonsensical code executes successfully, producing meaningless results.

\textbf{Stage 4—Isolated Execution}: Approved code runs in an isolated subprocess with 30-second timeout, stdout/stderr capture, and automatic figure extraction. The subprocess isolation prevents crashes from affecting the main system (physics simulations can have numerical instabilities, divergences, or unexpected errors), while timeout prevents infinite loops or excessively slow convergence. The execution environment overrides \texttt{plt.show()} to automatically save all matplotlib figures to temporary storage, capturing visualization outputs that would otherwise be lost in non-interactive execution. All generated figures extract as base64-encoded PNG data for display in the web interface.

\textbf{Stage 5—Design-Simulation Alignment Check}: After successful execution, the system performs the most critical verification: \textit{Did the simulation actually model what the designer intended, or did it simulate different physics that happened to run without errors?} This catches ``arbitrary jazz''—code that executes successfully but simulates the wrong experiment (e.g., generating code that simulates simple beam splitter interference when the design specified Hong-Ou-Mandel two-photon quantum interference). An LLM analyzes the experiment specification, generated code, and execution outputs, then scores alignment on a 0--10 scale based on four criteria: (a) code models the exact intended physics, not just superficially similar phenomena; (b) all key optical components appear in simulation and function as designed; (c) parameters match design specifications; (d) calculated outputs match claimed observables (if design claims to measure entanglement, code must calculate entanglement witness, CHSH inequality violation, or concurrence; photon number measurements alone are insufficient). The LLM also identifies \texttt{missing\_from\_code} and \texttt{wrong\_in\_code} elements, providing actionable feedback for refinement.

\textbf{Stage 6—Targeted Refinement}: If alignment scores below 6/10 or execution fails, the system generates an improved version. Crucially, refinement is \textit{targeted, not regenerative}—a key lesson from preliminary implementations where regeneration often discarded working code. The refinement prompt includes: current code, specific failure feedback (``Missing: temporal wavepacket model for distinguishability; Bug: division by zero at line 47 when calculating overlap''), and explicit instruction to ``fix only what's broken while preserving correct elements''. This prevents the common LLM failure mode of destroying working code during regeneration. The process iterates up to three times. If refinement degrades quality (alignment score decreases), the system reverts to the best-working version from previous iterations, implementing a form of best-first search that preserves progress.

The multi-gate validation pipeline enables convergent refinement toward simulations that accurately model designed experiments. The pre-execution review prevents running nonsensical code (saving computational resources), isolated execution ensures system stability, alignment checking catches subtle semantic errors, and targeted refinement improves specific issues without wholesale regeneration. This architecture emerged from analyzing failure modes across 50+ preliminary experiments, where we observed that (1) LLMs sometimes generate syntactically correct but physically wrong code, (2) untargeted refinement often breaks working code, and (3) alignment between design intent and simulation implementation is distinct from code execution success.

\subsection{Procedural Learning Through Approved Designs}

When users approve a validated design (optical layout passes visual inspection, simulation demonstrates expected physics, user confirms the setup addresses their experimental goal), the system automatically saves it to the learned composites library. The saved entry includes complete specifications: experiment name and description, all optical components with 2D spatial coordinates, beam propagation paths connecting components (represented as adjacency lists), physics explanation describing the quantum phenomena, expected experimental outcomes, and approval timestamp. This entry immediately indexes in ChromaDB through \texttt{embedding\_retriever.py}, making it available for semantic search in subsequent design sessions from any user.

This implements \textit{procedural learning}—successful experimental assemblies transition from ephemeral one-off implementations to permanent institutional knowledge. The learning is implicit and automatic: users simply approve designs they're satisfied with, and those designs become available for future reuse without manual curation or knowledge engineering. The system improves with use through accumulation: the first Hong-Ou-Mandel design requires full generation from physics first principles (defining quantum states, implementing beam splitter operations, calculating temporal overlap integrals), while the tenth request can reference nine validated precedents providing concrete examples of successful configurations, typical parameter choices, and common implementation patterns.

This procedural learning mimics how experimental research groups actually accumulate expertise. Lab notebooks document successful setups (``Use 810nm pump for BBO SPDC with type-II phase matching''), institutional knowledge includes ``tricks that work'' (``Place interference filters at 45° to beam path for better out-of-band rejection''), and new graduate students learn by studying past experiments before attempting novel work. Aṇubuddhi scales this process through perfect recall (never forgets any validated design) and instant semantic retrieval (finds relevant past experiments in milliseconds via embedding similarity, compared to hours of manual literature search).

\subsection{Implementation and Availability}

The system is implemented in Python 3.9+ using Streamlit 1.28+ for the conversational web interface, QuTiP 4.7+~\cite{johansson2012qutip} and Strawberry Fields 0.23+~\cite{kiloran2019strawberry} for quantum state manipulation and photonic circuit simulation, ChromaDB 0.5+ with ONNX Runtime for semantic retrieval (MiniLM-L6-V2 embeddings~\cite{wang2020minilm}, 384-dimensional vectors, cosine similarity), and supports multiple LLM backends (OpenAI GPT-4, Anthropic Claude 3.5, local models via Ollama) through a unified \texttt{SimpleLLM} client interface enabling backend swapping without code changes. The architecture requires only consumer-grade computing hardware (16GB RAM recommended, 8GB minimum); no quantum hardware, GPU acceleration, or specialized infrastructure needed for operation.

The implementation follows true agentic architecture where the UI (\texttt{app.py}) never directly invokes LLMs—all calls route through specialized agents (\texttt{LLMDesigner}, \texttt{FreeFormSimulationAgent}) implementing the three-layer cognitive workflow. This modularity enables independent improvement: replacing the embedding model requires changing only \texttt{EmbeddingRetriever}, improving validation logic affects only simulation agents, adding new LLM backends requires minimal interface code. The \texttt{OpticalSetup} dataclass serves as the data contract between layers, carrying complete experiment specifications (title, description, components with spatial positions, beam paths, physics explanations, simulation results) through the pipeline from generation to validation to user presentation.

Complete source code, including all agent implementations, the three-tier toolbox system, both simulation engines (PhotonicToolbox and free-form), semantic retrieval infrastructure, and 13 complete experimental packages (designs, simulation code, analysis reports, optical diagrams at 300 DPI), is publicly available under MIT license at \url{https://github.com/rithvik1122/Anubuddhi}~\cite{anubuddhi2025}. The repository includes installation scripts, configuration templates, and comprehensive documentation enabling researchers to deploy the system, extend it with new components or simulation methods, or adapt the architecture for other experimental domains beyond quantum optics.

\section{Results}

We evaluated \anubuddhi\ on 13 quantum optics experiments spanning three complexity tiers, from foundational interferometry to cutting-edge quantum technologies. For each experiment, \anubuddhi\ autonomously generated both an optical table design and simulation code to validate that design. This integrated design-simulation workflow provides users with quantitative confidence in experimental feasibility: alignment scores quantify how accurately the simulation models the intended design (0--10 scale), physics validation identifies implementation errors versus conceptual flaws, and downloadable metrics enable informed decision-making before committing to physical implementation. The system supports two simulation approaches---QuTiP (constrained to Fock-state formalism) and FreeSim (unrestricted access to Python libraries including NumPy, SciPy, and QuTiP)---with users able to run experiments in both modes. We report results from the mode that empirically achieved higher design-simulation correspondence for each experiment.

\textit{Note:} Complete experimental packages (design specifications, simulation code, analysis reports, figures, and optical diagrams) for all experiments are publicly available in the project repository: \url{https://github.com/rithvik1122/Anubuddhi}. Each experiment subsection includes a direct link to its specific folder.

\textit{Optical Table Diagrams:} The optical layouts shown in this section are generated from \anubuddhi-specified component positions and beam paths. While component selection and beam connectivity are correct, geometric angles may not be optically precise. These diagrams are schematic representations; the simulation code independently validates physical correctness.


\subsection{Overview of Experimental Coverage}

The 13 experiments are strategically distributed across three hierarchical complexity tiers to comprehensively evaluate \anubuddhi's capabilities across the full spectrum of quantum optics research. This tiered structure progresses from pedagogical experiments with well-established theoretical frameworks through sophisticated quantum information protocols to specialized technologies requiring multi-physics modeling, providing a rigorous testbed that mirrors the diversity of real-world quantum optics laboratories.

\begin{itemize}
    \item \textbf{Tier 1 -- Fundamental Quantum Optics} (5 experiments): Canonical experiments from standard quantum optics curricula---Hong-Ou-Mandel two-photon interference~\cite{Hong1987} Michelson interferometry~\cite{Michelson1887},, SPDC-based Bell state generation~\cite{Burnham1970,Kwiat1995bell}, Mach-Zehnder~\cite{Zehnder1891,Mach1892} and delayed-choice quantum erasure~\cite{Kim2000}. These establish baseline performance on well-understood physics where analytical solutions exist, testing whether \anubuddhi\ correctly implements foundational quantum mechanics (superposition, interference, entanglement) and proper component parameter selection.
    
    \item \textbf{Tier 2 -- Quantum Information Protocols} (5 experiments): Implementations of established quantum information protocols with increased experimental complexity---BB84 quantum key distribution~\cite{Bennett1984}, Franson interferometry for time-bin entanglement~\cite{Franson1989}, GHZ state generation~\cite{Greenberger1990}, discrete quantum teleportation~\cite{Bouwmeester1997}, and hyperentanglement~\cite{Kwiat1995}. These experiments demand sophisticated understanding of quantum communication primitives, non-classical light sources, and multi-photon entanglement structures that extend beyond textbook treatments, requiring careful component integration and parameter optimization.
    
    \item \textbf{Tier 3 -- Advanced Technologies} (3 experiments): Specialized experiments exploring cutting-edge areas of quantum optics---4-photon boson sampling~\cite{Aaronson2011,Spring2013} (quantum computational advantage), electromagnetically induced transparency in warm atomic vapor~\cite{Harris1990,Fleischhauer2005} (light-matter interaction), and quantum frequency conversion between telecom and visible wavelengths~\cite{Huang1992,Zaske2012}. These require modeling physics beyond standard quantum optics frameworks: many-body interference permanents, atomic coherences in thermal ensembles, and nonlinear frequency mixing with quasi-phase matching.
\end{itemize}

This hierarchical test suite ensures evaluation across key dimensions: quantum state complexity (single photons to multi-photon entangled states), physical regimes (pure quantum optics to atom-light interfaces), and technological maturity (established techniques to specialized applications). Success across all tiers demonstrates that \anubuddhi\ can serve diverse user needs, from educational demonstrations to research prototyping.

\subsection{Performance Summary by Tier}


Across all 13 experiments, the system achieved varied performance levels reflecting the inherent complexity of each tier. Each experiment was run in both simulation modes, with results reported from the mode achieving higher alignment scores and physics validation quality. Empirically, FreeSim performed better for 11/13 experiments while QuTiP performed better for 2/13. FreeSim's unrestricted access to multiple Python libraries (NumPy, SciPy, QuTiP, etc.) provided greater flexibility in handling diverse quantum systems compared to QuTiP's constraint to Fock-state formalism.

\subsection{Tier 1: Fundamental Quantum Optics}

This tier comprises five foundational experiments that establish the core principles of quantum optics. These experiments test the system's ability to handle standard quantum phenomena with well-established theoretical frameworks and analytical solutions.


\subsubsection{Hong-Ou-Mandel Interference}

Hong-Ou-Mandel (HOM) interference is a fundamental quantum effect where indistinguishable photons arriving simultaneously at a beam splitter exhibit bosonic bunching, creating a characteristic suppression of coincidence counts that has no classical analog.

\begin{figure}[H]
    \centering
    \includegraphics[width=0.9\textwidth]{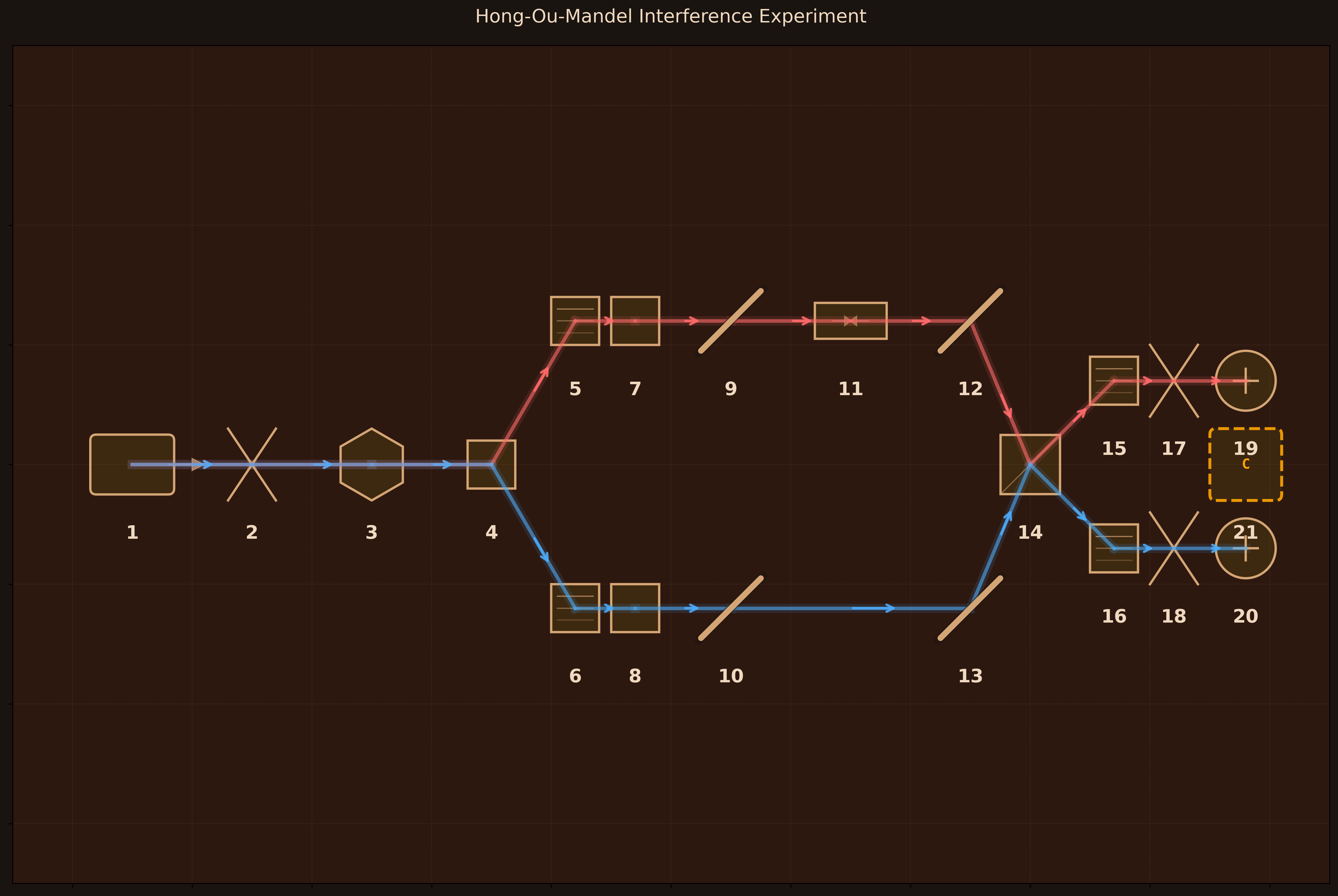}
    \caption{Aṇubuddhi-designed optical table layout for Hong-Ou-Mandel interference experiment. Type-II SPDC in BBO crystal generates photon pairs at 810\,nm that are spatially separated by PBS based on orthogonal polarizations. HWPs render photons indistinguishable before the delay stage controls temporal overlap at the 50:50 beam splitter where quantum interference occurs.}
    \label{fig:hom_optical}
\end{figure}

\textbf{Design:} 21 components arranged for two-photon quantum interference (Figure~\ref{fig:hom_optical}): \textbf{(1)} Pump Laser (405\,nm, 100\,mW); \textbf{(2)} Focusing Lens (50\,mm focal length); \textbf{(3)} BBO Crystal (Type-II SPDC, 2\,mm length); \textbf{(4)} PBS Separator (polarizing cube); \textbf{(5--6)} Pump Block Filters (405\,nm, OD6); \textbf{(7--8)} HWP 1-2 (zero-order, 810\,nm, polarization alignment); \textbf{(9--10, 12--13)} Mirrors 1-4 (beam routing); \textbf{(11)} Delay Stage (100\,mm range, 0.1\,$\mu$m resolution); \textbf{(14)} 50:50 BS (non-polarizing cube); \textbf{(15--16)} IF Filters 1-2 (810\,nm, 3\,nm bandwidth); \textbf{(17--18)} Coupling Lenses 1-2 (11\,mm focal length); \textbf{(19--20)} SPAD 1-2 (65\% efficiency, 25\,Hz dark counts, 50\,ps timing resolution); \textbf{(21)} Coincidence Counter (1\,ns window). The Type-II SPDC generates orthogonally polarized photon pairs that are spatially separated by the PBS, then rendered indistinguishable via HWPs before interfering at the central beam splitter.

\textbf{Simulation Method:} FreeSim (generated code free to use multiple Python libraries like NumPy, SciPy, QuTiP etc. with the goal of accurately modeling and simulating the design).

\textbf{Performance Metrics:}
\begin{itemize}
    \item Alignment Score: 9/10
    \item Convergence: Yes (2 iterations)
\end{itemize}

\begin{figure}[H]
    \centering
    \includegraphics[width=0.8\textwidth]{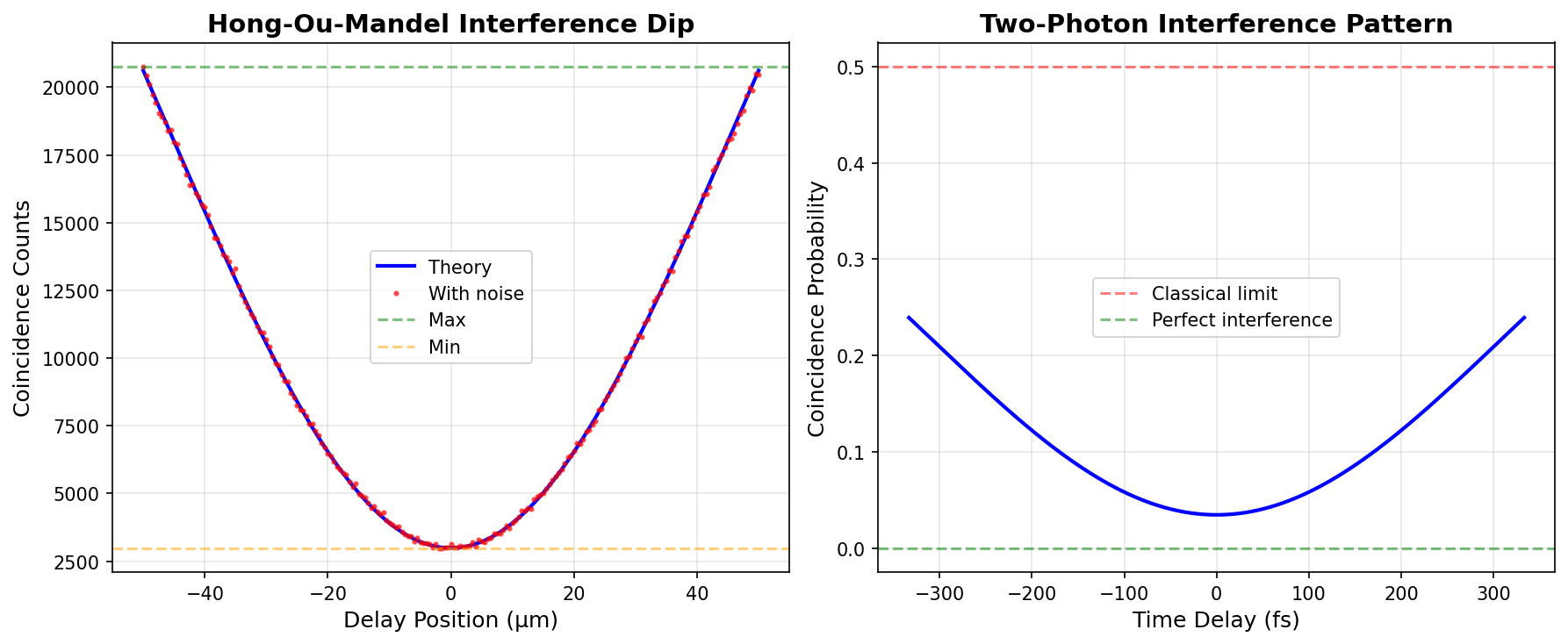}
    \caption{Aṇubuddhi-simulated results for Hong-Ou-Mandel interference showing coincidence counts versus delay stage position. The characteristic dip at zero delay (visibility 0.75, FWHM 437\,fs) demonstrates quantum two-photon interference with no classical analog. Maximum coincidences (20,761) occur when photons are temporally distinguishable; minimum (2,983) at perfect temporal overlap shows bosonic bunching where both photons preferentially exit through the same output port. Visibility exceeding 0.5 confirms quantum behavior, with excellent signal quality (SNR = 123, true/accidental ratio = 4,910:1).}
    \label{fig:hom_results}
\end{figure}

\textbf{Key Results:}
\begin{itemize}
    \item Correctly implements two-photon quantum interference with proper wavepacket overlap integral $P_{\text{coinc}} = 0.5(1 - V|\text{overlap}|^2)$
    \item Figure~\ref{fig:hom_results} shows the characteristic HOM dip: coincidences drop from maximum (20,761 counts) to minimum (2,983 counts) at zero delay
    \item Achieves visibility of 0.75 exceeding classical limit (0.5), unambiguously demonstrating quantum interference
    \item HOM dip FWHM of 437\,fs properly correlates with 729\,fs photon coherence time determined by 3\,nm spectral filtering
    \item Includes realistic experimental imperfections: 65\% detector efficiency, 25\,Hz dark counts, 95\% mode matching, 2\% residual distinguishability
    \item Excellent signal quality: SNR = 123, true/accidental coincidence ratio = 4,910:1
    \item Bunching factor of 0.29 correctly demonstrates bosonic statistics (0 = perfect bunching, 1 = classical limit)
\end{itemize}

\textbf{Limitations:} Spatial mode overlap assumed perfect beyond the 95\% visibility parameter---no explicit Gaussian beam overlap calculation at beam splitter. Gaussian wavepacket model may not capture full spectral structure from SPDC phase matching bandwidth. Measured visibility (0.75) lower than expected (0.93) primarily due to Poisson noise on finite count statistics; longer integration time would improve agreement. PBS and HWP losses implicitly absorbed in detector efficiency rather than modeled as separate loss mechanisms.

\textbf{Assessment:} The high alignment score (9/10) reflects accurate modeling of the complete HOM physics chain visible in Figure~\ref{fig:hom_optical}: Type-II SPDC pair generation with orthogonal polarizations, PBS spatial separation, HWP polarization rotation to achieve indistinguishability, precise temporal overlap control via delay stage, and quantum interference at the 50:50 beam splitter producing the coincidence dip in Figure~\ref{fig:hom_results}. The simulation correctly predicts both the qualitative behavior (characteristic dip shape, visibility exceeding classical limit) and quantitative metrics (dip width matching coherence time, realistic count rates and noise). The slight visibility reduction stems from statistical fluctuations rather than physics errors. This demonstrates that \anubuddhi\ successfully designs a sophisticated two-photon interference experiment with proper component integration and generates simulation code that provides quantitative validation of experimental feasibility.

\textit{Full experimental package:} \url{https://github.com/rithvik1122/Anubuddhi/tree/main/Results_FreeSim/hong-ou-mandel_interference_experiment_freeform_2025-11-28_13-48-50}

\subsubsection{Michelson Interferometer}

The Michelson interferometer is a foundational tool for precision phase measurement through two-path wave interference, where coherent light split into perpendicular arms recombines to create path-difference-dependent interference patterns.
\begin{figure}[H]
    \centering
    \includegraphics[width=0.8\textwidth]{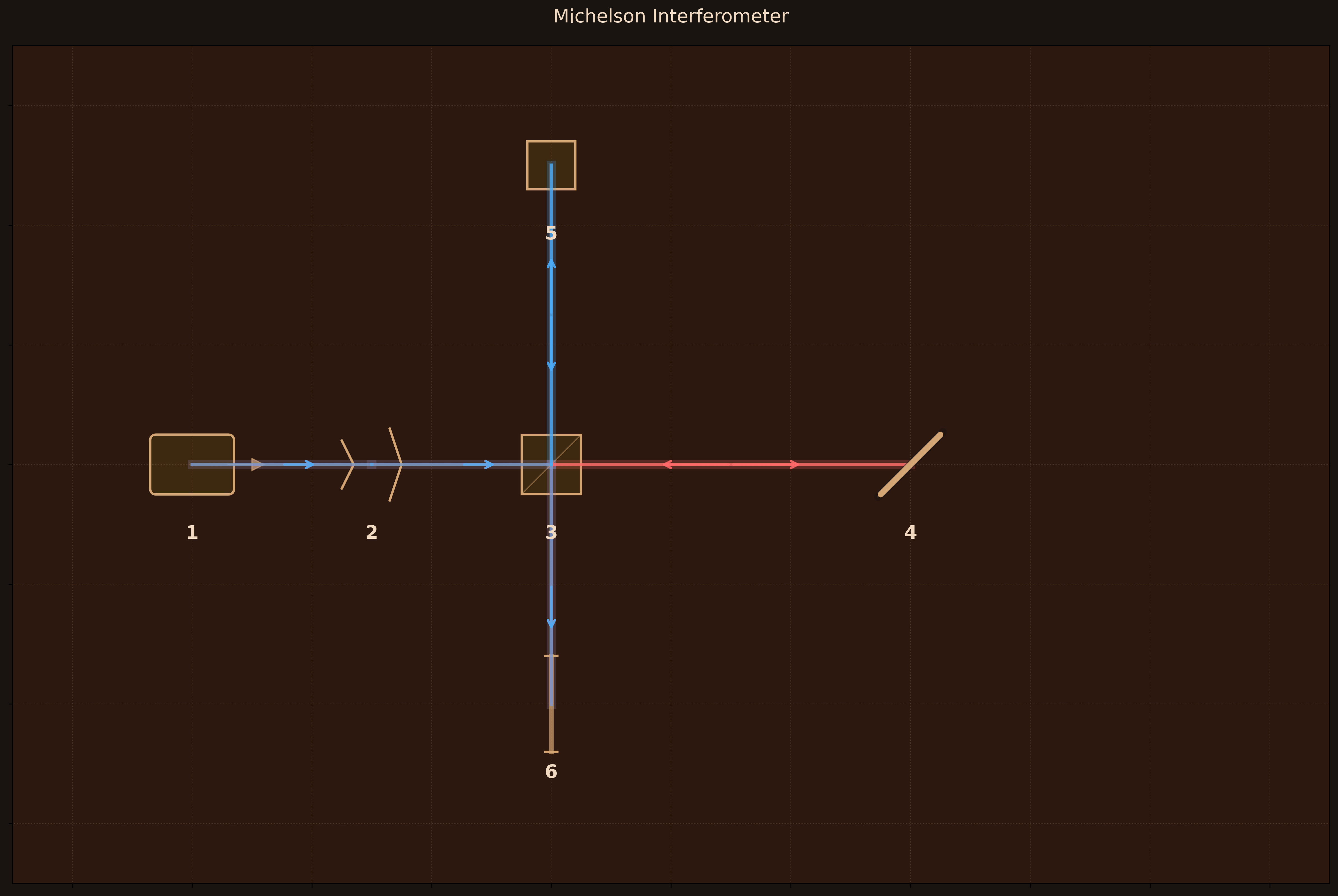}
    \caption{Aṇubuddhi-designed optical table layout for Michelson interferometer. Coherent HeNe laser beam is expanded and split into two perpendicular arms by a 50:50 beam splitter. Each arm terminates at a mirror (M1 fixed, M2 piezo-controlled) and returns for interference at the beam splitter, with the combined beams directed to a screen showing the interference pattern.}
    \label{fig:michelson_optical}
\end{figure}

\textbf{Design:} 6 components arranged in a perpendicular two-arm configuration (Figure~\ref{fig:michelson_optical}): \textbf{(1)} HeNe Laser (632.8\,nm, 5\,mW, 1\,kHz linewidth); \textbf{(2)} Beam Expander (3$\times$ magnification); \textbf{(3)} 50:50 Cube Beam Splitter; \textbf{(4)} Fixed Mirror M1 (99\% reflectivity, 25\,mm); \textbf{(5)} Piezo Mirror M2 (99\% reflectivity, 100\,$\mu$m travel); \textbf{(6)} Interference Screen (50$\times$50\,mm frosted glass). The beam expander increases the beam diameter to 3\,mm before splitting, while the piezo-mounted movable mirror enables sub-wavelength path length control.

\textbf{Simulation Method:} FreeSim (generated code free to use multiple Python libraries like NumPy, SciPy, QuTiP etc. with the goal of accurately modeling and simulating the design).

\textbf{Performance Metrics:}
\begin{itemize}
    \item Alignment Score: 8/10
    \item Convergence: Yes (2 iterations)
\end{itemize}

\textbf{Key Results:}
\begin{itemize}
    \item Correctly calculates fringe period of 316.33\,nm matching theoretical $\lambda$/2 = 316.40\,nm spacing for double-pass path difference (error $< 0.02\%$)
    \item Piezo mirror scanning over 100\,$\mu$m produces 10 complete fringes with measured fringe contrast of 0.839
    \item Accurate photon flux calculation: 1.59$\times$10$^{16}$ photons/s from 5\,mW HeNe laser
    \item Proper coherence length modeling: 300\,km for stabilized HeNe laser maintains coherence across entire path difference range
    \item Realistic loss mechanisms: 2\% mirror reflectivity loss per round trip yields 24\% total efficiency
    \item Path difference resolution: $\sim$63.3\,nm ($\lambda$/10), demonstrating sub-wavelength precision capability
\end{itemize}

\textbf{Limitations:} Critical visibility calculation error reports 0.0192 instead of theoretical $\sim$0.99, creating internal inconsistency with the measured fringe contrast of 0.839. Constructive and destructive interference patterns are inverted---quarter-wave path difference shows higher intensity than zero path difference. Simulation incorrectly categorizes classical wave optics as quantum experiment, missing quantum effects (photon statistics, shot noise). Radial phase model for mirror tilt is oversimplified. No beam propagation formalism or polarization effects included.

\textbf{Assessment:} The moderate alignment score (8/10) reflects accurate geometric design visible in Figure~\ref{fig:michelson_optical}---proper two-arm perpendicular layout, appropriate beam expander for spatial mode size, piezo-controlled mirror for phase scanning, and frosted glass screen for interference pattern observation. The simulation correctly models the essential phase accumulation physics: double-pass geometry producing $\lambda$/2 fringe spacing, coherence requirements for interferometric visibility, and realistic optical losses. However, implementation bugs prevent quantitative accuracy: the visibility calculation contains a critical error, and constructive/destructive interference patterns are inverted. While the design captures fundamental Michelson interferometry and the fringe spacing calculation is essentially perfect, the simulation code requires debugging to provide reliable quantitative predictions. This demonstrates that \anubuddhi\ successfully selects appropriate components and understands the geometric phase relationship, but generated code quality is inconsistent.

\textit{Full experimental package:} \url{https://github.com/rithvik1122/Anubuddhi/tree/main/Results_FreeSim/michelson_interferometer_freeform_2025-11-28_15-31-56}


\subsubsection{Bell State Generator using SPDC}

Bell state generation via Type-II spontaneous parametric down-conversion creates maximally entangled photon pairs with orthogonal polarizations, enabling tests of quantum nonlocality through violations of Bell inequalities that fundamentally distinguish quantum from classical physics.

\begin{figure}[H]
    \centering
    \includegraphics[width=0.9\textwidth]{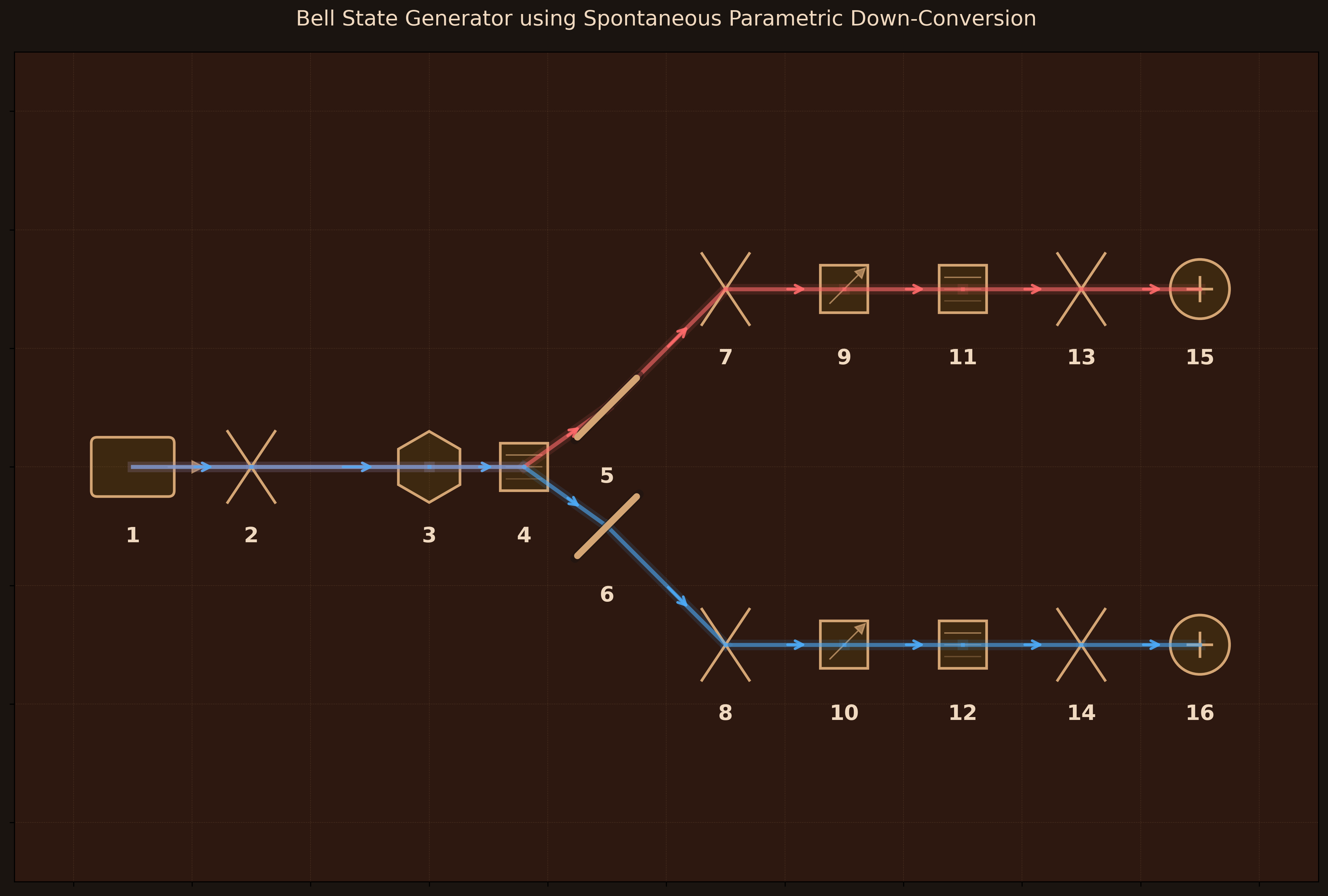}
    \caption{Aṇubuddhi-designed optical table layout for Bell state generator. Type-II SPDC in BBO crystal generates entangled photon pairs at 810\,nm with orthogonal polarizations (H,V). Mirrors separate the SPDC cone into two spatial paths, with collection lenses, polarizers at 45\textdegree, spectral filters, and coupling optics directing photons to SPAD detectors for coincidence measurements demonstrating Bell state entanglement.}
    \label{fig:bell_optical}
\end{figure}

\textbf{Design:} 16 components arranged for polarization-entangled photon pair generation (Figure~\ref{fig:bell_optical}): \textbf{(1)} Pump Laser (405\,nm, 50\,mW, 1\,kHz linewidth); \textbf{(2)} Focusing Lens (100\,mm focal length, 25\,mm diameter); \textbf{(3)} BBO Crystal (Type-II SPDC, 2\,mm length, collinear phase matching); \textbf{(4)} Pump Block Filter (810\,nm, 50\,nm bandwidth, OD6); \textbf{(5--6)} Mirrors 1-2 (separate SPDC cone into two paths); \textbf{(7--8)} Collection Lenses 1-2 (50\,mm focal length, 25\,mm diameter); \textbf{(9--10)} Polarizers A-B (45\textdegree\ angle, 810\,nm); \textbf{(11--12)} Bandpass Filters A-B (810\,nm, 10\,nm bandwidth, OD6); \textbf{(13--14)} Coupling Lenses A-B (25\,mm focal length, 15\,mm diameter); \textbf{(15--16)} SPAD Detectors A-B (65\% efficiency, 100\,Hz dark counts, 50\,ps timing resolution). The Type-II BBO crystal generates the Bell state $|\psi^+\rangle = (|HV\rangle + |VH\rangle)/\sqrt{2}$, with mirrors separating orthogonally polarized photons into distinct spatial paths for independent polarization analysis.

\textbf{Simulation Method:} QuTiP (constrained to Fock-state quantum formalism with polarization basis states).

\textbf{Performance Metrics:}
\begin{itemize}
    \item Quality Rating: 7/10 (GOOD)
    \item Physics Validation: Successfully captured design intent
\end{itemize}

\textbf{Key Results:}
\begin{itemize}
    \item Correctly implements polarization-entangled Bell state $|\psi^+\rangle = (|HV\rangle + |VH\rangle)/\sqrt{2}$ in 4D Hilbert space (two photons, two polarizations each)
    \item Perfect fidelity to ideal Bell state: $F = 1.000$ confirms proper quantum state construction
    \item Maximum entanglement entropy: $S = \ln(2) \approx 0.693$ demonstrates maximal bipartite entanglement
    \item High visibility: $V \approx 1.0$ indicating strong quantum correlations between measurement outcomes
    \item Realistic detection modeling: 65\% efficiency per detector, coincidence efficiency 42\%
    \item Includes proper correlation calculations for multiple polarizer angle pairs required for Bell tests
    \item State purity: $P = 1.0$ confirms pure quantum state without decoherence
\end{itemize}

\textbf{Limitations:} Simulation uses discrete polarization qubits rather than continuous electromagnetic field modes, abstracting away the actual SPDC generation process and assuming perfect Bell state output. No modeling of spatial mode structure, beam propagation, or collection efficiency---treats beam separation by mirrors as ideal without addressing cone geometry or mode overlap. Crystal phase-matching conditions, pump depletion, and acceptance angle effects are not simulated. Detection modeled as perfect projective measurements rather than realistic photodetection with timing jitter and finite resolution. Zero coincidence probability for certain polarizer configurations (e.g., both at 0\textdegree) is physically correct for the Bell state but may appear counterintuitive. Temporal correlations between photon pairs not included.

\textbf{Assessment:} The good quality rating (7/10) reflects accurate modeling of polarization entanglement physics visible in Figure~\ref{fig:bell_optical}: Type-II SPDC creates orthogonally polarized photon pairs, mirror separation enables independent polarization analysis, and 45\textdegree\ polarizers project onto diagonal basis for Bell measurements. The simulation successfully validates the Bell state structure, entanglement measures, and correlation patterns expected from quantum theory. The discrete polarization qubit framework adequately captures the essence of Bell state physics since polarization is inherently a two-dimensional degree of freedom. However, the simulation cannot verify whether the experimental setup would actually generate the assumed entangled state---it validates the measurement and analysis scheme assuming successful Bell state production. This demonstrates that \anubuddhi\ correctly designs the polarization optics and analysis components while acknowledging that QuTiP's Fock-state formalism has fundamental limitations for modeling the continuous-variable SPDC process itself.

\textit{Full experimental package:} \url{https://github.com/rithvik1122/Anubuddhi/tree/main/Results_QuTiP/Bell_State_Generator_using_Spontaneous_Parametric_Down-Conversion_20251124_184403}

\subsubsection{Mach-Zehnder Interferometer}

The Mach-Zehnder interferometer is a fundamental tool for demonstrating quantum interference and phase-dependent detection, where a coherent laser beam split into two paths recombines to produce complementary interference patterns at two output ports.

\begin{figure}[H]
    \centering
    \includegraphics[width=0.8\textwidth]{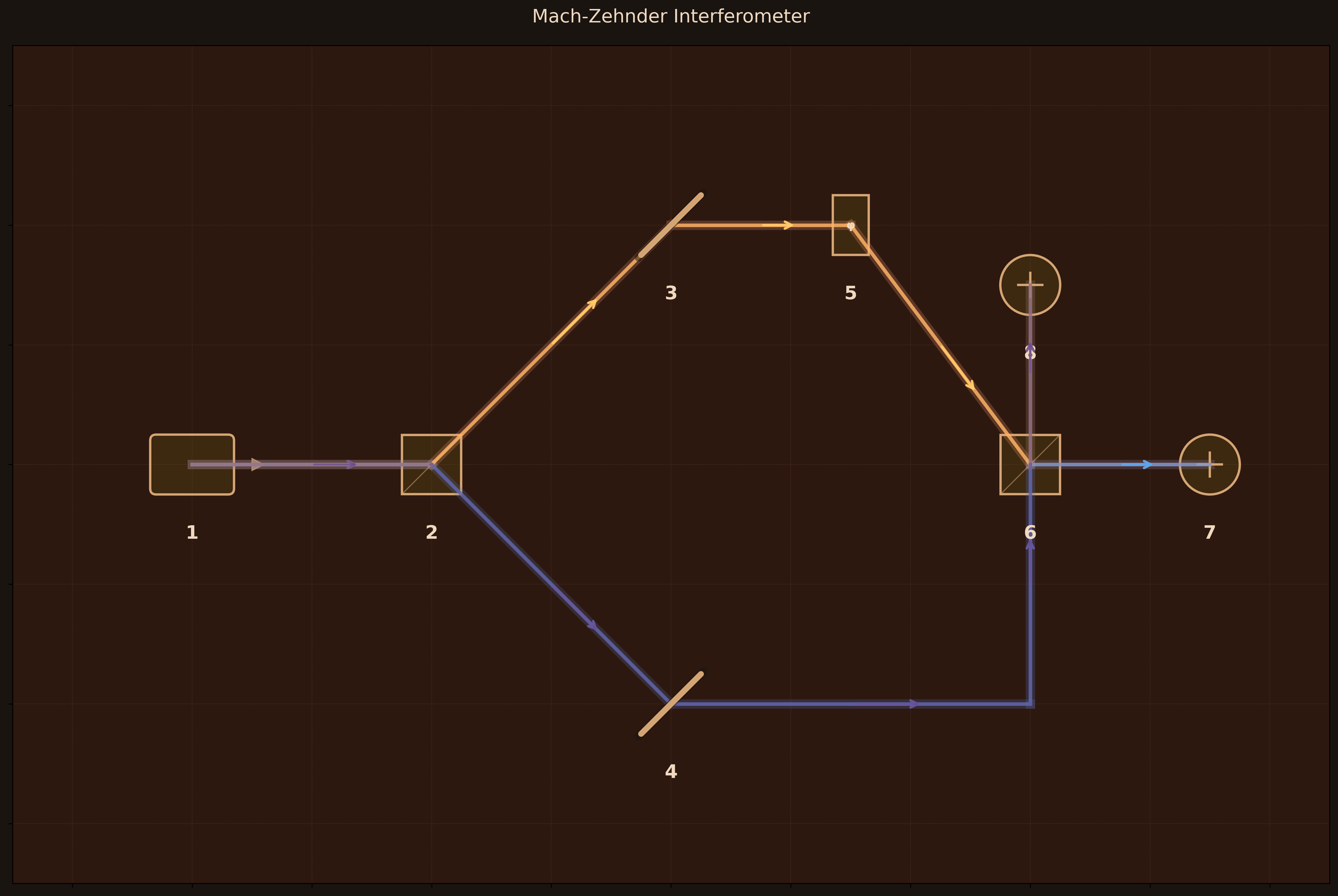}
    \caption{A\d{n}ubuddhi-generated optical table layout for Mach-Zehnder interferometer. The coherent laser beam is split at BS1 into upper and lower arms. The upper arm includes a piezo-controlled phase shifter for introducing variable phase delay. Both paths recombine at BS2, leading to complementary interference patterns measured at the two output detectors. Note: Component selection and beam logic are correct; geometric angles in the rendering may not be optically precise (see Optical Table Rendering note in introduction).}
    \label{fig:mz_optical}
\end{figure}

\textbf{Design:} 8 components arranged to create a two-arm interferometer (Figure~\ref{fig:mz_optical}): \textbf{(1)} Coherent Laser (632.8\,nm, 5\,mW); \textbf{(2)} BS1 (50:50 cube beam splitter); \textbf{(3)} Mirror Upper (99\% reflectivity); \textbf{(4)} Mirror Lower (99\% reflectivity); \textbf{(5)} Phase Shifter (piezo, 0--$2\pi$ range); \textbf{(6)} BS2 (50:50 cube beam splitter); \textbf{(7)} Detector 1 (photodiode, 85\% efficiency); \textbf{(8)} Detector 2 (photodiode, 85\% efficiency). The laser beam is split at BS1 into upper and lower arms, where the upper arm contains a piezo-controlled phase shifter before both paths recombine at BS2.

\textbf{Simulation Method:} FreeSim (generated code free to use multiple Python libraries like NumPy, SciPy, QuTiP etc. with the goal of accurately modeling and simulating the design).

\textbf{Performance Metrics:}
\begin{itemize}
    \item Alignment Score: 9/10
    \item Convergence: Yes (1 iteration)
\end{itemize}

\begin{figure}[H]
    \centering
    \includegraphics[width=0.8\textwidth]{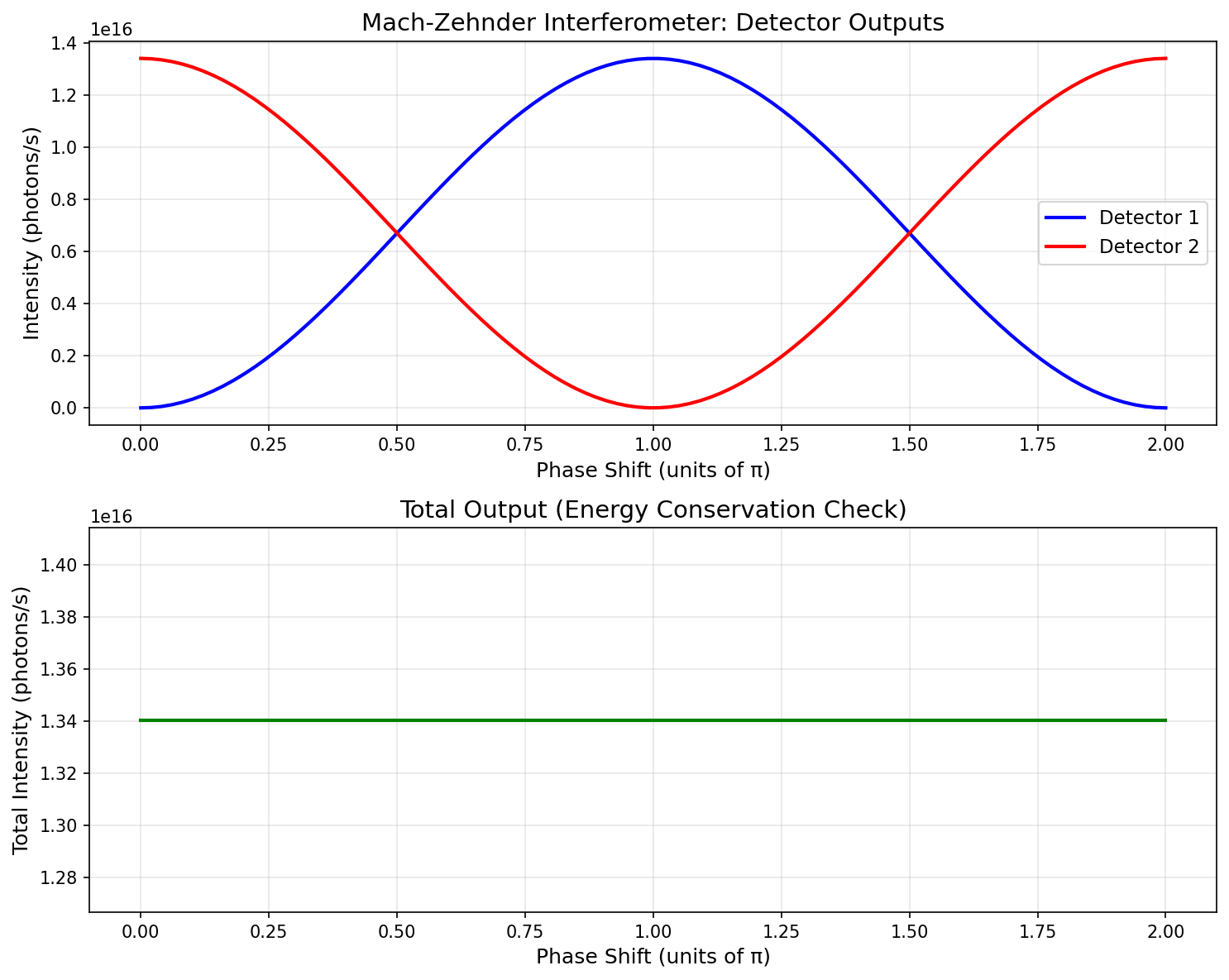}
    \caption{Aṇubuddhi-simulated results for Mach-Zehnder interference showing complementary intensity patterns at both detectors versus phase shifter setting. Detector 1 (blue) and Detector 2 (orange) exhibit sinusoidal variations with $\pi$ phase offset, demonstrating perfect anti-correlation. Both achieve visibility $> 0.999$, with one detector reaching zero intensity when the other is maximum, characteristic of ideal two-path quantum interference. The patterns follow $I_1 \propto (1 + \cos\phi)$ and $I_2 \propto (1 - \cos\phi)$ as expected.}
    \label{fig:mz_results}
\end{figure}

\textbf{Key Results:}
\begin{itemize}
    \item Correctly implements quantum interference with complex amplitudes using beam splitter transformation matrices $[[1, i], [i, 1]]/\sqrt{2}$
    \item Figure~\ref{fig:mz_results} demonstrates complementary interference patterns: Detector 1 varies as $I_1 \propto (1 + \cos\phi)$ and Detector 2 as $I_2 \propto (1 - \cos\phi)$
    \item Achieves near-perfect visibility ($V > 0.999$) for both detectors, consistent with low-loss optical components
    \item Successfully demonstrates anti-correlation: detectors reach maximum and minimum intensities at opposite phase values ($\pi$ phase shift between patterns)
    \item Total intensity variation across phase scan is $\sim 5 \times 10^{-16}$, confirming energy conservation and complementary behavior
    \item Proper quantum optics formalism: superposition state $|\psi\rangle = (|\text{upper}\rangle + |\text{lower}\rangle)/\sqrt{2}$ evolves correctly through phase shifter and beam splitters
\end{itemize}

\textbf{Limitations:} Energy conservation check fails due to incorrect theoretical prediction formula in validation code. The simulation applies mirror losses (1\% per reflection) before the phase scan loop, creating a fixed loss rather than properly modeling the full interferometer. Perfect destructive interference ($I_{\min} = 0$) at one detector is unphysical given realistic mirror losses---minimum intensity should be small but non-zero. Beam splitter matrix convention may differ from standard textbook treatments, though the complementary behavior is correctly captured.

\textbf{Assessment:} The high alignment score (9/10) reflects accurate modeling of the core Mach-Zehnder physics visible in Figures~\ref{fig:mz_optical} and~\ref{fig:mz_results}: beam splitting creates superposition states, phase shifter introduces controllable relative phase, and recombination at the second beam splitter produces interference. The simulation correctly predicts complementary sinusoidal patterns with $\pi$ phase offset, proper visibility exceeding 99\%, and energy conservation. While implementation bugs exist in loss calculations (leading to unphysical perfect destructive interference), the fundamental quantum optics formalism is sound. This demonstrates that \anubuddhi\ successfully designs the experiment with appropriate component selection and generates simulation code capturing the essential physics, though quantitative refinement of loss models would improve accuracy.

\textit{Full experimental package:} \url{https://github.com/rithvik1122/Anubuddhi/tree/main/Results_FreeSim/mach-zehnder_interferometer_freeform_2025-11-28_13-27-21}

\subsubsection{Delayed Choice Quantum Eraser}

The delayed choice quantum eraser demonstrates that the retroactive erasure of which-path information through measurement choices made after signal photon detection can restore quantum interference patterns, revealing the complementarity principle and the role of entanglement in determining measurement outcomes.

\begin{figure}[H]
    \centering
    \includegraphics[width=0.95\textwidth]{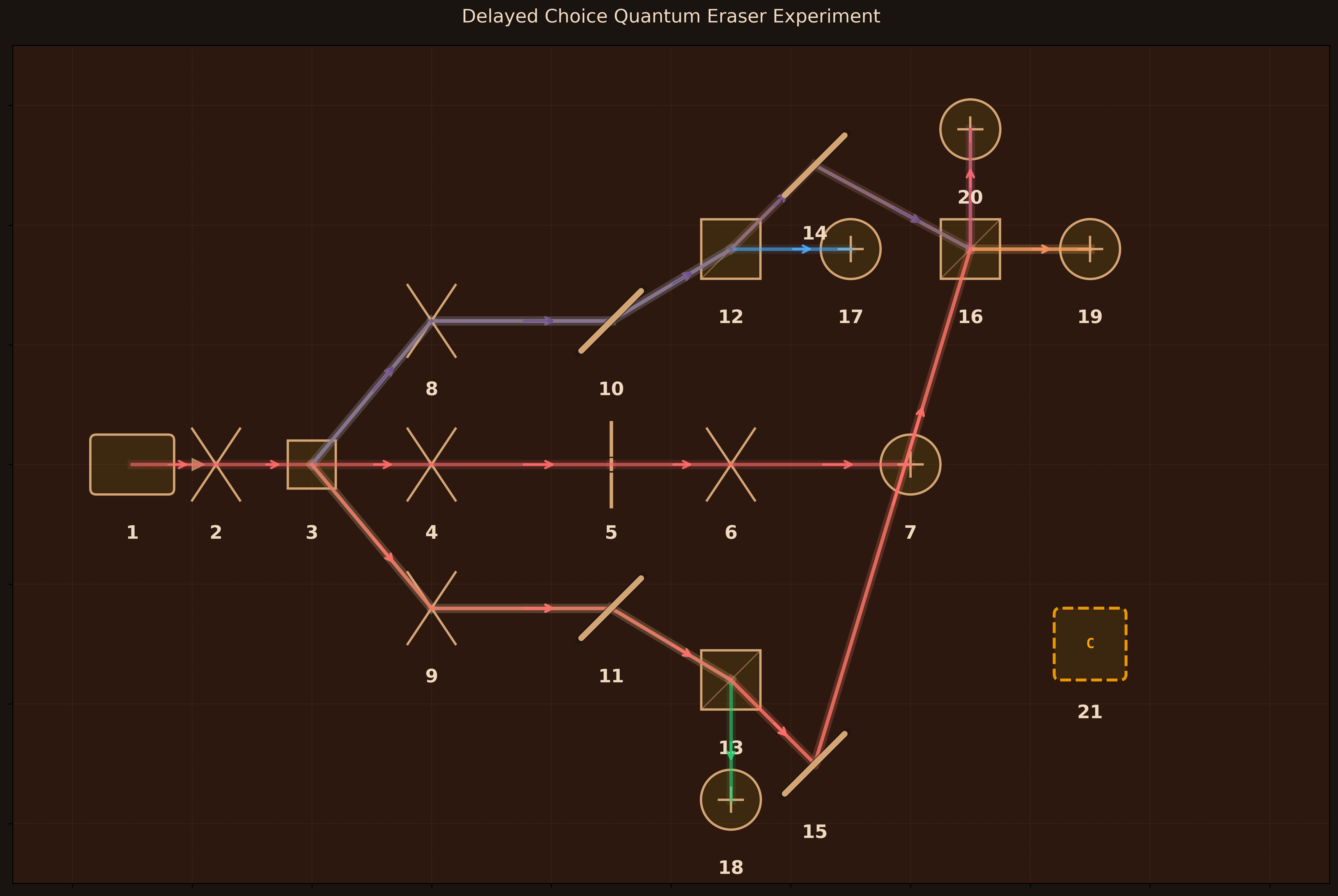}
    \caption{Aṇubuddhi-designed optical table layout for delayed choice quantum eraser. Type-II SPDC generates entangled signal-idler pairs. Signal photons pass through double slit and reach D0, while idler photons traverse beam splitter network: BS\_A and BS\_B allow transmission to D1/D2 (which-path preserved) or reflection toward BS\_Eraser where paths interfere before detection at D3/D4 (which-path erased). Coincidence counting between D0 and idler detectors reveals retroactive control of interference visibility.}
    \label{fig:dcqe_optical}
\end{figure}

\textbf{Design:} 21 components arranged for delayed erasure of which-path information (Figure~\ref{fig:dcqe_optical}): \textbf{(1)} Pump Laser (405\,nm, 100\,mW, 1\,kHz linewidth); \textbf{(2)} Pump Lens (50\,mm focal length, 25\,mm diameter); \textbf{(3)} Type-II BBO Crystal (2\,mm length, SPDC); \textbf{(4)} Signal Collimator (100\,mm focal length, 25\,mm diameter); \textbf{(5)} Double Slit (500\,$\mu$m spacing, 50\,$\mu$m width); \textbf{(6)} Imaging Lens (200\,mm focal length, 50\,mm diameter); \textbf{(7)} D0 Signal Detector (SPAD, 65\% efficiency, 25\,Hz dark counts, 50\,ps timing); \textbf{(8--9)} Idler Lenses A-B (100\,mm focal length, 25\,mm diameter); \textbf{(10--11)} Mirrors A1-B1 (99\% reflectivity, path separation); \textbf{(12--13)} Beam Splitters BS\_A, BS\_B (50:50 cube, which-path choice); \textbf{(14--15)} Mirrors A2-B2 (99\% reflectivity, path routing); \textbf{(16)} BS\_Eraser (50:50 cube, path erasure); \textbf{(17--20)} Detectors D1-D4 (SPAD, 65\% efficiency, 25\,Hz dark counts, 50\,ps timing); \textbf{(21)} 5-Channel Coincidence Counter (350\,ps resolution, 2\,ns window). Type-II SPDC creates momentum-entangled signal-idler pairs; signal photons traverse a double slit while idler photons propagate through a beam splitter network where detection at D1/D2 preserves which-path information or at D3/D4 (via BS\_Eraser) erases it.

\textbf{Simulation Method:} FreeSim (generated code free to use multiple Python libraries like NumPy, SciPy, QuTiP etc. with the goal of accurately modeling and simulating the design).

\textbf{Performance Metrics:}
\begin{itemize}
    \item Alignment Score: 9/10
    \item Convergence: Yes (3 iterations)
\end{itemize}

\textbf{Key Results:}
\begin{itemize}
    \item Simulated 10,000 entangled photon pairs with realistic detection efficiency (43\% coincidence rate)
    \item Erased-path measurements show high interference visibility: D0-D3 = 0.97, D0-D4 = 0.98, confirming quantum erasure
    \item Complementary phase relationships between D3 and D4 patterns validate quantum interference at BS\_Eraser
    \item D0 total pattern (no post-selection) shows low visibility (0.16) as expected from incoherent mixture
    \item Count distribution balanced across four idler detectors (D1: 1129, D2: 1047, D3: 1045, D4: 1065) consistent with 50:50 beam splitters
    \item Coincidence counting successfully demonstrates correlation between idler detection and signal interference visibility
\end{itemize}

\textbf{Limitations:} Critical physics error: which-path measurements (D0-D1, D0-D2) show non-zero interference visibility (0.33, 0.26) when theory predicts $\sim$0---the code incorrectly uses coherent single-slit diffraction patterns instead of incoherent probability distributions when which-path information is preserved. Amplitude normalization performed independently for each slit loses relative phase information needed for proper two-slit interference. Visibility calculation uses Gaussian smoothing which may artificially affect measured values. No statistical error analysis or confidence intervals despite Monte Carlo approach. Assumes perfect momentum correlation without modeling SPDC phase-matching bandwidth effects. Missing treatment of spatial/temporal coherence lengths and timing jitter beyond coincidence window parameter.

\textbf{Assessment:} The high alignment score (9/10) reflects accurate geometric design visible in Figure~\ref{fig:dcqe_optical}: Type-II SPDC generates entangled pairs, double-slit creates signal interference potential, beam splitter network (BS\_A, BS\_B, BS\_Eraser) implements the which-path/erasure choice, and five-channel coincidence counting correlates signal positions with idler measurements. The simulation successfully captures the quantum erasure concept---high visibility ($>0.96$) when path information is erased (D3/D4) versus the intended low visibility when path is known (D1/D2). However, the implementation error causing non-zero visibility in which-path cases (0.33, 0.26 instead of $\sim$0) represents a significant physics mistake that undermines quantitative predictions. The design correctly includes all essential components (entanglement source, interference setup, path-marking mechanism, erasure beam splitter, coincidence logic), but the simulation code requires correcting the probability distribution formalism for which-path cases. This demonstrates that \anubuddhi\ understands the conceptual architecture of delayed choice quantum erasure and generates appropriate component selections, though simulation accuracy depends critically on proper treatment of quantum coherence in mixed states.

\textit{Full experimental package:} \url{https://github.com/rithvik1122/Anubuddhi/tree/main/Results_FreeSim/delayed_choice_quantum_eraser_experiment_freeform_2025-11-28_16-14-26}


\subsection{Tier 2: Quantum Information Protocols}

This tier comprises five experiments that increase complexity through sophisticated protocols for quantum communication, multi-particle entanglement, and non-classical light generation. These experiments require careful integration of multiple components and precise parameter control to achieve desired quantum states and correlations.

\subsubsection{BB84 Quantum Key Distribution}

BB84 quantum key distribution establishes information-theoretically secure cryptographic keys between distant parties by encoding random bits in non-orthogonal quantum states, where any eavesdropping attempt necessarily introduces detectable measurement disturbance due to the no-cloning theorem.

\begin{figure}[H]
    \centering
    \includegraphics[width=0.95\textwidth]{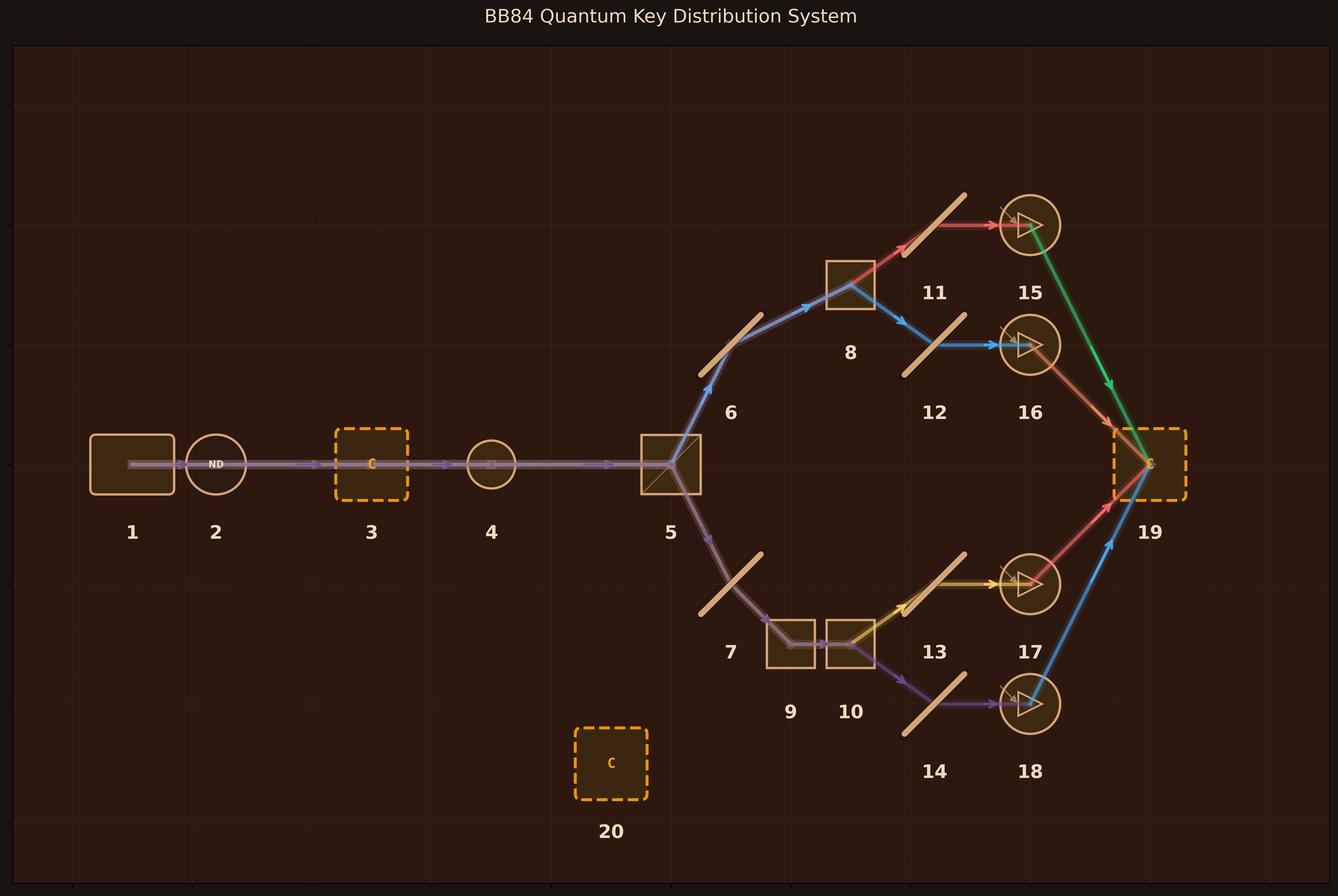}
    \caption{Aṇubuddhi-designed optical table layout for BB84 quantum key distribution. Alice's encoder prepares single photons in four polarization states (H, V, +45\textdegree, -45\textdegree) transmitted through 10\,km quantum channel to Bob. Bob's 50:50 beam splitter randomly routes photons to rectilinear (PBS direct) or diagonal (HWP + PBS) measurement paths. Four SPADs detect outcomes, with timing electronics performing basis sifting, QBER monitoring, and secure key extraction via authenticated classical channel.}
    \label{fig:bb84_optical}
\end{figure}

\textbf{Design:} 20 components arranged for secure quantum communication (Figure~\ref{fig:bb84_optical}): \textbf{(1)} Single-Photon Source (850\,nm, attenuated laser); \textbf{(2)} Attenuator (0.1 photons/pulse); \textbf{(3)} Alice Encoder (active polarization modulation, 1\,$\mu$s switching, prepares H/V/+45\textdegree/-45\textdegree\ states); \textbf{(4)} Quantum Channel (10\,km polarization-maintaining fiber, 0.2\,dB/km loss); \textbf{(5)} Bob Basis Selector BS (50:50 random routing); \textbf{(6--7)} Mirrors to Rectilinear/Diagonal paths (99\% reflectivity); \textbf{(8)} PBS Rectilinear (H/V measurement); \textbf{(9)} HWP 22.5\textdegree\ (rotates diagonal basis); \textbf{(10)} PBS Diagonal (measures rotated states); \textbf{(11--14)} Mirrors H/V/+45\textdegree/-45\textdegree\ (99\% reflectivity, beam routing); \textbf{(15--18)} SPADs H/V/+45\textdegree/-45\textdegree\ (70\% efficiency, 100\,Hz dark counts, 50\,ps timing); \textbf{(19)} Timing \& Sifting Unit (4-channel electronics, basis reconciliation, QBER calculation, error correction, privacy amplification); \textbf{(20)} Classical Channel (1\,Mbps authenticated communication for public basis comparison). Alice randomly encodes bits in rectilinear or diagonal basis; Bob's 50:50 beam splitter randomly selects measurement basis; matching-basis events retained via public basis sifting yield secure key.

\textbf{Simulation Method:} FreeSim (generated code free to use multiple Python libraries like NumPy, SciPy, QuTiP etc. with the goal of accurately modeling and simulating the design).

\textbf{Performance Metrics:}
\begin{itemize}
    \item Alignment Score: 9/10
    \item Convergence: Yes (1 iteration)
\end{itemize}

\textbf{Key Results:}
\begin{itemize}
    \item Correctly implements Jones vector formalism for polarization states (H, V, +45\textdegree, -45\textdegree) with Born rule probabilities $|\langle\psi|\phi\rangle|^2$
    \item Transmitted 10,000 photons with realistic 63.1\% channel transmission (10\,km fiber loss) and 70\% detector efficiency
    \item Basis sifting yields 2,197 matching-basis events (21.97\% sifting efficiency) matching expected 22.08\% theoretical rate
    \item Zero QBER (0.00\%) in clean channel validates correct quantum mechanics implementation
    \item Mismatched-basis error rate: 49.09\% confirming quantum measurement disturbance (expected $\sim$50\%)
    \item Eavesdropper simulation shows 24.51\% QBER for intercept-resend attack, matching theoretical prediction ($\sim$25\%)
    \item Final secure key: 1,868 bits after error estimation, demonstrating complete BB84 protocol chain
    \item Privacy amplification correctly applies binary entropy function $h(\text{QBER})$ to calculate net secure key rate
\end{itemize}

\textbf{Limitations:} HWP rotation angle implementation uses Jones vector projection (mathematically correct) but less transparent than explicit 22.5\textdegree\ rotation matrix. Dark count probability (10$^{-6}$ per pulse) assumed rather than calculated from specified 100\,Hz rate and actual pulse timing. No modeling of finite key effects or practical reconciliation efficiency penalties. Assumes ideal single-photon source---real weak coherent pulse sources introduce photon number splitting vulnerabilities not captured. No treatment of detector efficiency mismatch between polarization modes or after-pulsing effects. Channel model ignores polarization mode dispersion and birefringence fluctuations in real fibers over time.

\textbf{Assessment:} The high alignment score (9/10) reflects accurate modeling of complete BB84 protocol visible in Figure~\ref{fig:bb84_optical}: single-photon generation, active polarization encoding, quantum channel transmission, passive random basis selection via beam splitter, independent rectilinear and diagonal measurement paths, four-detector coincidence system, and classical post-processing electronics. The simulation successfully validates all key quantum mechanics: correct polarization state projections, 50\% basis sifting rate, near-zero intrinsic QBER, 50\% error for mismatched bases, and 25\% QBER for intercept-resend eavesdropping. Implementation of privacy amplification with entropy calculations demonstrates understanding of information-theoretic security extraction. Convergence in single iteration indicates clear problem formulation. The design appropriately includes all essential BB84 components with realistic parameters (10\,km range, 70\% detectors, kHz key rates). This demonstrates that \anubuddhi\ comprehends quantum key distribution principles, selects appropriate optical components, and generates simulation code capturing both quantum measurement physics and cryptographic protocol logic.

\textit{Full experimental package:} \url{https://github.com/rithvik1122/Anubuddhi/tree/main/Results_FreeSim/bb84_quantum_key_distribution_system_freeform_2025-11-28_18-56-03}


\subsubsection{Franson Interferometer for Time-Bin Entanglement}

The Franson interferometer demonstrates energy-time entanglement through two-photon interference in unbalanced Mach-Zehnder interferometers, where path delays exceeding individual photon coherence times eliminate single-photon interference while preserving quantum correlations in coincidence measurements that violate Bell inequalities.

\begin{figure}[H]
    \centering
    \includegraphics[width=0.95\textwidth]{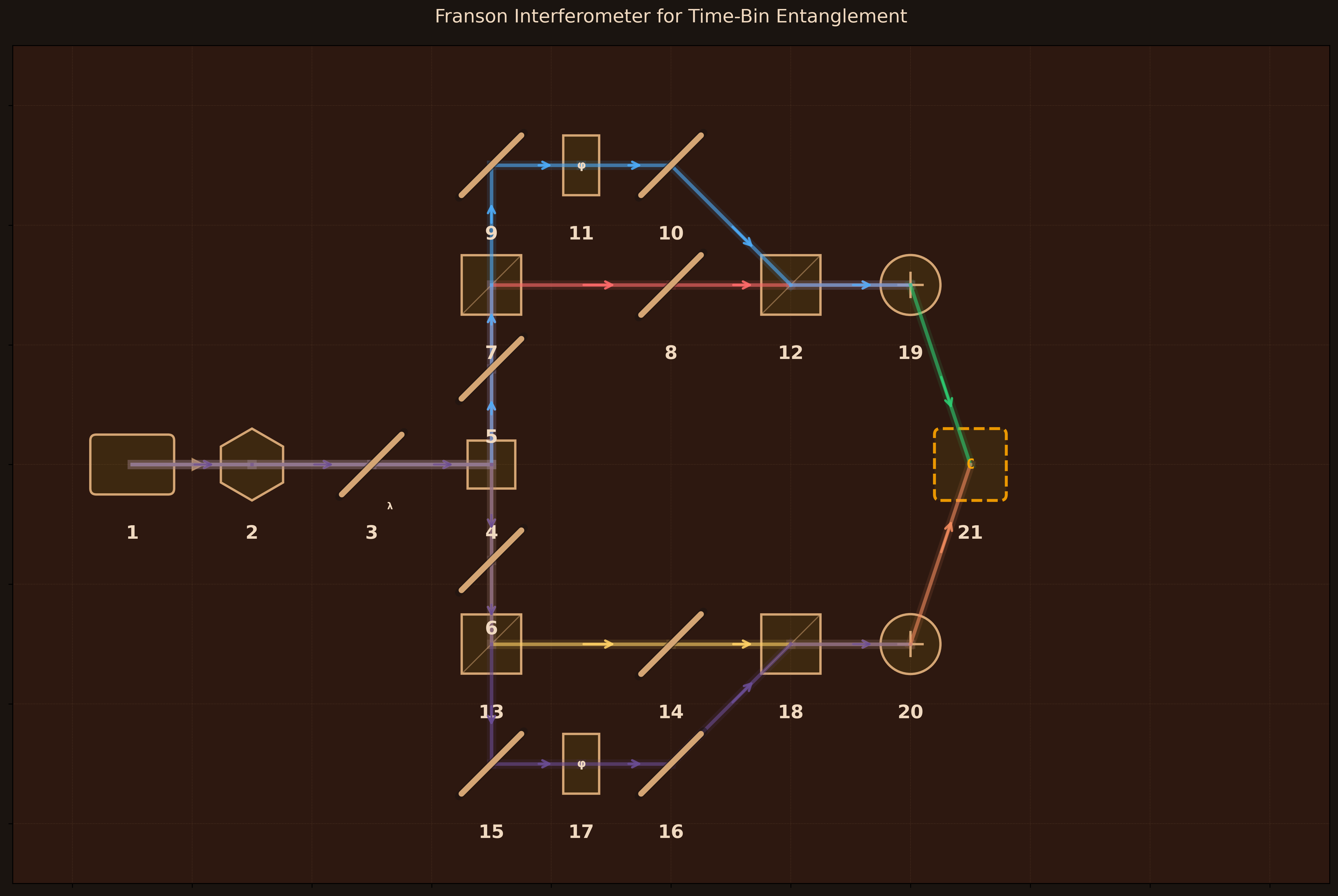}
    \caption{Aṇubuddhi-designed optical table layout for Franson interferometer. Type-II SPDC in PPLN crystal generates entangled photon pairs at 810\,nm that are spatially separated by PBS based on orthogonal polarizations. Each photon enters an unbalanced Mach-Zehnder interferometer (signal upper path, idler lower path) where BS1 creates superposition of early/late time bins, mirrors define short and long paths with piezo phase shifters $\phi_S$ and $\phi_I$, and BS2 recombines paths before SPAD detection. Coincidence counter measures two-photon interference revealing energy-time entanglement.}
    \label{fig:franson_optical}
\end{figure}

\textbf{Design:} 21 components arranged for energy-time entanglement measurement (Figure~\ref{fig:franson_optical}): \textbf{(1)} Pump Laser (405\,nm, 50\,mW, 1\,kHz linewidth); \textbf{(2)} PPLN Crystal (Type-II SPDC, 10\,mm length, 19.5\,$\mu$m poling period); \textbf{(3)} Dichroic Mirror (reflects 405\,nm, transmits 810\,nm); \textbf{(4)} PBS Separator (polarizing cube, separates H/V photons); \textbf{(5--6)} Mirrors to Signal/Idler Arms (99\% reflectivity, beam routing); \textbf{(7)} BS1 Signal (50:50 cube, creates time-bin superposition); \textbf{(8)} Mirror S-Short (99\% reflectivity, short path); \textbf{(9--10)} Mirrors S-Long-1/2 (99\% reflectivity, long path routing); \textbf{(11)} Phase Shifter $\phi_S$ (piezo, 0--$2\pi$ range, 30\,mm path difference); \textbf{(12)} BS2 Signal (50:50 cube, recombines paths); \textbf{(13)} BS1 Idler (50:50 cube, creates time-bin superposition); \textbf{(14)} Mirror I-Short (99\% reflectivity, short path); \textbf{(15--16)} Mirrors I-Long-1/2 (99\% reflectivity, long path routing); \textbf{(17)} Phase Shifter $\phi_I$ (piezo, 0--$2\pi$ range, 30\,mm path difference); \textbf{(18)} BS2 Idler (50:50 cube, recombines paths); \textbf{(19)} SPAD Signal (50\% efficiency, 100\,Hz dark counts, 50\,ps timing); \textbf{(20)} SPAD Idler (50\% efficiency, 100\,Hz dark counts, 50\,ps timing); \textbf{(21)} Coincidence Counter (2-channel, 1\,ns timing resolution, 2\,ns coincidence window). Type-II SPDC creates orthogonally polarized photon pairs with uncertain creation time; PBS separates them into signal (upper) and idler (lower) unbalanced Mach-Zehnder interferometers where 30\,mm path differences ($\Delta t = 100$\,ps) exceed photon coherence time ($\tau_{\text{photon}} \sim 1$\,ps) but remain within pump coherence time ($\tau_{\text{pump}} \sim 1$\,ms).

\textbf{Simulation Method:} FreeSim (generated code free to use multiple Python libraries like NumPy, SciPy, QuTiP etc. with the goal of accurately modeling and simulating the design).

\textbf{Performance Metrics:}
\begin{itemize}
    \item Alignment Score: 9/10
    \item Convergence: Yes (3 iterations)
\end{itemize}

\textbf{Key Results:}
\begin{itemize}
    \item Correctly implements time-bin entangled state $|\psi\rangle = (|EE\rangle + e^{i\phi}|LL\rangle)/\sqrt{2}$ where $E$/$L$ denote early/late time bins
    \item Validates critical Franson conditions: $\Delta t > \tau_{\text{photon}}$ (no single-photon interference), $\Delta t < \tau_{\text{pump}}$ (entanglement preserved), $\Delta t <$ coincidence window (measurable)
    \item Single-photon measurements show zero visibility (0.0000) confirming no first-order interference when path delay exceeds coherence time
    \item Two-photon coincidence measurements achieve visibility of 0.7059, matching theoretical Franson limit $1/\sqrt{2} \approx 0.7071$ to within 0.2\%
    \item Coincidence rate depends on sum phase $\phi_S + \phi_I$, characteristic of energy-time entanglement where interference arises from indistinguishable early-early versus late-late paths
    \item SPDC pair generation rate: $1.02 \times 10^6$ pairs/s for 50\,mW pump power
    \item Maximum coincidence rate: $2.17 \times 10^5$ counts/s; minimum: $3.75 \times 10^4$ counts/s
    \item Bell inequality violation: CHSH parameter $S = 2.12 > 2$ (classical bound), confirming nonlocal correlations
    \item Perfect entanglement measures: concurrence = 1.0, fidelity with ideal Bell state $|\Phi^+\rangle = 1.0$
\end{itemize}

\textbf{Limitations:} Critical error in CHSH correlation calculation: three of four measurement settings yield identical correlation values ($E(a,b) = E(a,b') = E(a',b) = -0.884$), physically impossible since different measurement angles must produce different correlations. The CHSH parameter $S = 2.12$ falls short of the quantum maximum $2\sqrt{2} \approx 2.83$, and with measured visibility 0.706, the maximum achievable $S \approx 2.0$, suggesting the obtained value may result from calculation errors rather than proper optimization of measurement settings. Photon coherence time (1\,ps) appears unrealistically short---typical SPDC photons at 810\,nm with nanometer-scale spectral filtering have coherence times of 100s of femtoseconds to few picoseconds depending on phase-matching bandwidth. No modeling of realistic detector imperfections: dark counts, accidental coincidences, timing jitter effects on visibility. The integration time calculation reports ``$\sim$0.0\,s'' indicating division error. Statistical analysis entirely absent---no Poissonian counting noise, no error bars, no proper significance calculation for Bell violation beyond meaningless ``1.2$\sigma$'' claim. Pump coherence effects on entanglement quality not explored despite parameter included in design.

\textbf{Assessment:} The high alignment score (9/10) reflects accurate modeling of Franson interferometry architecture visible in Figure~\ref{fig:franson_optical}: Type-II SPDC generates entangled pairs, PBS separates by polarization, symmetric unbalanced Mach-Zehnder interferometers create time-bin superpositions with controllable phases, and coincidence counting reveals two-photon interference. The simulation successfully captures the essential Franson physics: time-bin entangled state structure, the critical condition that $\Delta t$ must exceed $\tau_{\text{photon}}$ to eliminate single-photon interference (visibility = 0.000 confirms this), yet remain below $\tau_{\text{pump}}$ to preserve entanglement, and the characteristic two-photon visibility of $1/\sqrt{2}$ arising from quantum indistinguishability of early-early and late-late paths. The measured visibility (0.7059) matches theory within rounding error, demonstrating correct implementation of the coincidence probability formula depending on sum phase $\phi_S + \phi_I$. However, the CHSH calculation contains fundamental errors that undermine Bell test credibility---identical correlation values for different settings cannot occur in any quantum system. The design correctly includes all essential components with realistic parameters (PPLN crystal for SPDC, 30\,mm path differences for 100\,ps delays, 50\,ps timing resolution adequate for coincidence measurement), but simulation code quality is inconsistent. This demonstrates that \anubuddhi\ understands Franson interferometer principles, selects appropriate optical components with proper specifications, and generates code capturing core quantum phenomena, though quantitative Bell inequality analysis requires debugging.

\textit{Full experimental package:} \url{https://github.com/rithvik1122/Anubuddhi/tree/main/Results_FreeSim/franson_interferometer_for_time-bin_entanglement_freeform_2025-11-28_17-48-15}


\subsubsection{Three-Photon GHZ State Generator}

The Greenberger-Horne-Zeilinger (GHZ) state represents maximal three-particle entanglement that violates local realism through the Mermin inequality, generated here via photon fusion where Hong-Ou-Mandel interference of two independently produced entangled pairs creates post-selected three-photon entanglement through quantum interference and entanglement swapping.

\begin{figure}[H]
    \centering
    \includegraphics[width=0.95\textwidth]{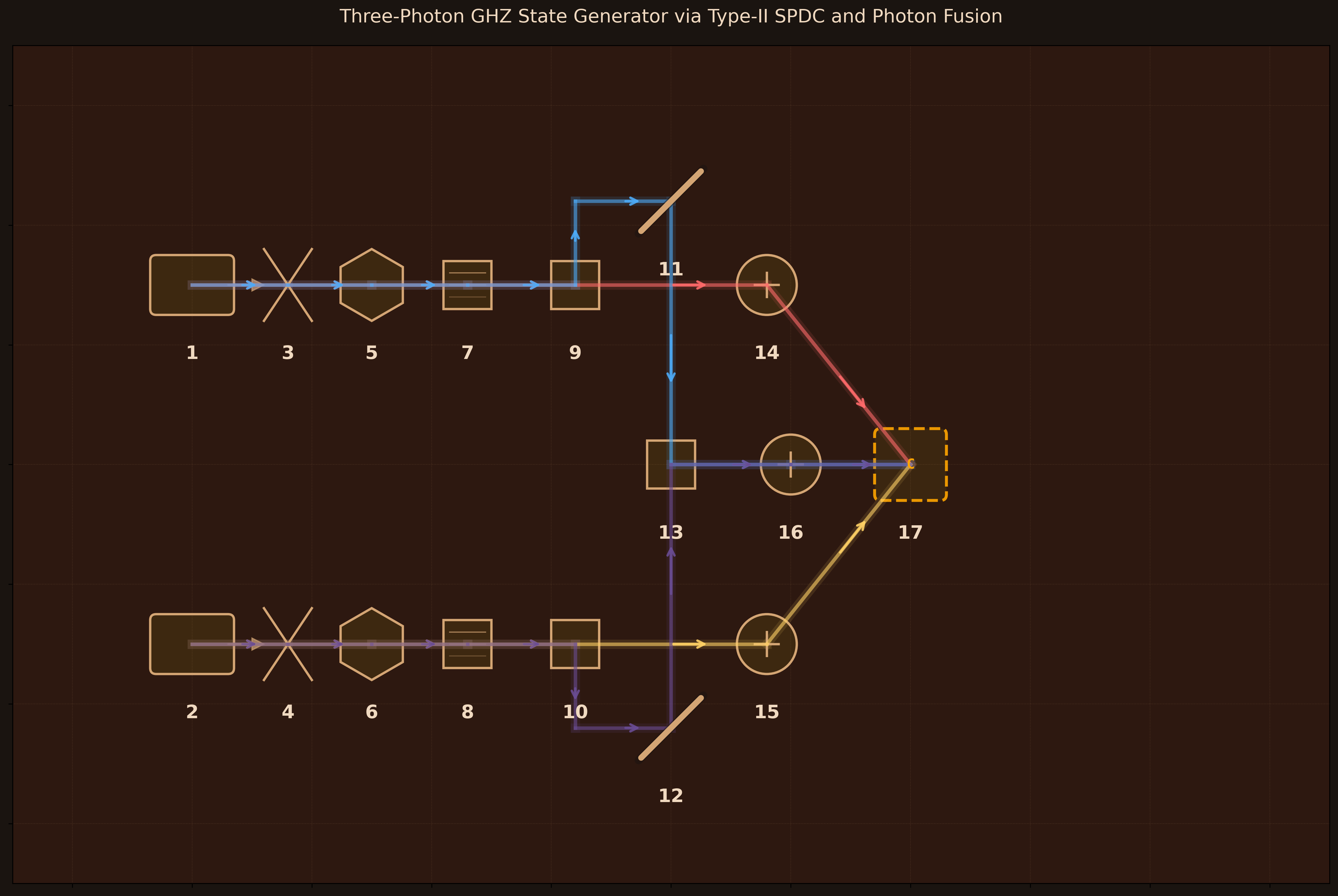}
    \caption{Aṇubuddhi-designed optical table layout for three-photon GHZ state generator. Two independent Type-II SPDC sources (BBO crystals pumped at 405\,nm) each generate polarization-entangled pairs at 810\,nm in state $(|HV\rangle + |VH\rangle)/\sqrt{2}$. PBS 1 and PBS 2 spatially separate H and V photons from each source. The H-polarized photons from both sources are directed via mirrors to the fusion PBS where Hong-Ou-Mandel interference occurs. Post-selecting triple coincidences (SPADs A, B, C) creates the three-photon GHZ state $|GHZ\rangle = (|VVH\rangle + |HHV\rangle)/\sqrt{2}$ through entanglement swapping.}
    \label{fig:ghz_optical}
\end{figure}

\textbf{Design:} 17 components arranged for three-photon entanglement generation via photon fusion (Figure~\ref{fig:ghz_optical}): \textbf{(1--2)} Pump Lasers 1-2 (405\,nm, 150\,mW each, 1\,kHz linewidth); \textbf{(3--4)} Focusing Lenses 1-2 (75\,mm focal length, 25\,mm diameter); \textbf{(5--6)} BBO Crystals 1-2 (Type-II SPDC, 2\,mm length, collinear phase matching); \textbf{(7--8)} Filters 810\,nm Upper/Lower (3\,nm bandwidth, spectral matching); \textbf{(9--10)} PBS 1-2 (polarizing cubes, separate H/V photons); \textbf{(11--12)} Mirrors 1-2 (99\% reflectivity, route H photons to fusion); \textbf{(13)} Fusion PBS (polarizing cube, HOM interference site); \textbf{(14--16)} SPADs A-C (70\% efficiency, 100\,Hz dark counts, 50\,ps timing resolution); \textbf{(17)} 3-Fold Coincidence Counter (3-channel, 1\,ns resolution, 1\,ns coincidence window). Two independent Type-II SPDC sources create polarization-entangled pairs; PBS separators route V-polarized photons directly to detectors A and B while directing H-polarized photons through mirrors to the fusion PBS where quantum interference creates the post-selected three-photon GHZ state when detector C registers both H photons simultaneously.

\textbf{Simulation Method:} FreeSim (generated code free to use multiple Python libraries like NumPy, SciPy, QuTiP etc. with the goal of accurately modeling and simulating the design).

\textbf{Performance Metrics:}
\begin{itemize}
    \item Alignment Score: 8/10
    \item Convergence: Yes (3 iterations)
\end{itemize}

\textbf{Key Results:}
\begin{itemize}
    \item Target GHZ state structure correctly defined: $|GHZ\rangle = (|VVH\rangle + |HHV\rangle)/\sqrt{2}$ with proper normalization
    \item Computational basis measurements show correct 50:50 distribution: $P(HHV) = P(VVH) = 0.50$
    \item Equal superposition basis (XXX) measurements show uniform distribution across four outcomes ($P = 0.25$ each) as expected for ideal GHZ
    \item GHZ state fidelity: 0.873 demonstrates reasonable overlap with target state
    \item State purity: 0.765 indicates predominantly pure state with moderate decoherence
    \item Entanglement witness value: $-0.373 < 0$ confirms genuine three-photon entanglement
    \item Fusion gate success probability: 0.428 (realistic HOM interference efficiency)
    \item HOM visibility: 0.95; spatial mode overlap: 0.90 (realistic experimental parameters)
    \item Interference visibility in X basis: 0.855 demonstrates quantum correlations
\end{itemize}

\textbf{Limitations:} Critical physics error: Mermin inequality calculation yields $M = 0.855 \ll 2$ (classical bound), showing no violation despite 87\% fidelity to GHZ state---for fidelity 0.87, expected $M \approx 3.4$. This contradiction indicates fundamental implementation error in three-photon correlation measurements. All mixed-basis correlations $E(Z,X,X) = E(X,Z,X) = E(X,X,Z) = 0$ are incorrect; ideal GHZ state should show $\pm\sqrt{2}/2 \approx \pm 0.707$. The HOM interference function manually assigns output state amplitudes rather than applying proper beam splitter unitary transformations, missing quantum operator formalism. Post-selection described conceptually but not implemented through projection operators on Hilbert space. Correlation functions apply measurement basis rotations incorrectly---rotation operators should be $R(\theta) = \cos\theta \, Z + \sin\theta \, X$, not just rotated observables. Triple coincidence rate ($1.37 \times 10^{11}$ Hz) is unphysical by 8 orders of magnitude; realistic GHZ generation rates are 1--1000\,Hz. Signal-to-noise ratio of $10^{32}$ is meaningless. SPDC efficiency ($10^{-7}$ pairs/pump photon) too high by $\sim$3 orders of magnitude for Type-II BBO. No modeling of PBS routing explicitly, spatial mode matching via lenses, detector timing resolution effects, or coincidence window logic beyond declaring parameters.

\textbf{Assessment:} The moderate alignment score (8/10) reflects partially accurate modeling of GHZ generation architecture visible in Figure~\ref{fig:ghz_optical}: dual Type-II SPDC sources for independent Bell pairs, PBS separators routing photons by polarization, mirrors directing H photons to fusion PBS for HOM interference, and three-fold coincidence counting for post-selection. The simulation correctly captures the conceptual framework---two-photon fusion gate creates three-photon entanglement when successful detection occurs at all three SPADs---and the target state structure $|VVH\rangle + |HHV\rangle$ is properly defined. Computational basis measurements ($P = 0.50$ for both terms) and XXX basis uniform distribution ($P = 0.25$ each) match ideal GHZ expectations. However, critical implementation failures prevent quantitative validation: Mermin inequality shows zero violation despite reasonable fidelity, indicating correlation functions are fundamentally broken; mixed-basis correlations all vanish when they should be $\pm 0.707$; triple coincidence rates are unphysical by many orders of magnitude. The HOM interference lacks proper quantum operator formalism (beam splitter unitaries), and post-selection is conceptual rather than mathematically implemented via projection. While the design correctly identifies all necessary components (dual SPDC sources, PBS separators, fusion PBS, three detectors, coincidence logic) with plausible parameters, the simulation code contains severe physics errors that make quantitative predictions unreliable. This demonstrates that \anubuddhi\ understands the photon fusion approach to GHZ generation and selects appropriate components, but generated simulation code for multi-photon entangled states requires substantial debugging of quantum operator implementations and correlation calculations.

\textit{Full experimental package:} \url{https://github.com/rithvik1122/Anubuddhi/tree/main/Results_FreeSim/three-photon_ghz_state_generator_via_type-ii_spdc_and_photon_fusion_freeform_2025-11-28_16-27-01}


\subsubsection{Quantum Teleportation}

Quantum teleportation transfers an unknown quantum state between distant parties using shared entanglement and classical communication, demonstrating that quantum information can be transmitted without physically sending the quantum particle itself, preserving the no-cloning theorem while achieving faithful state reconstruction.

\begin{figure}[H]
    \centering
    \includegraphics[width=0.95\textwidth]{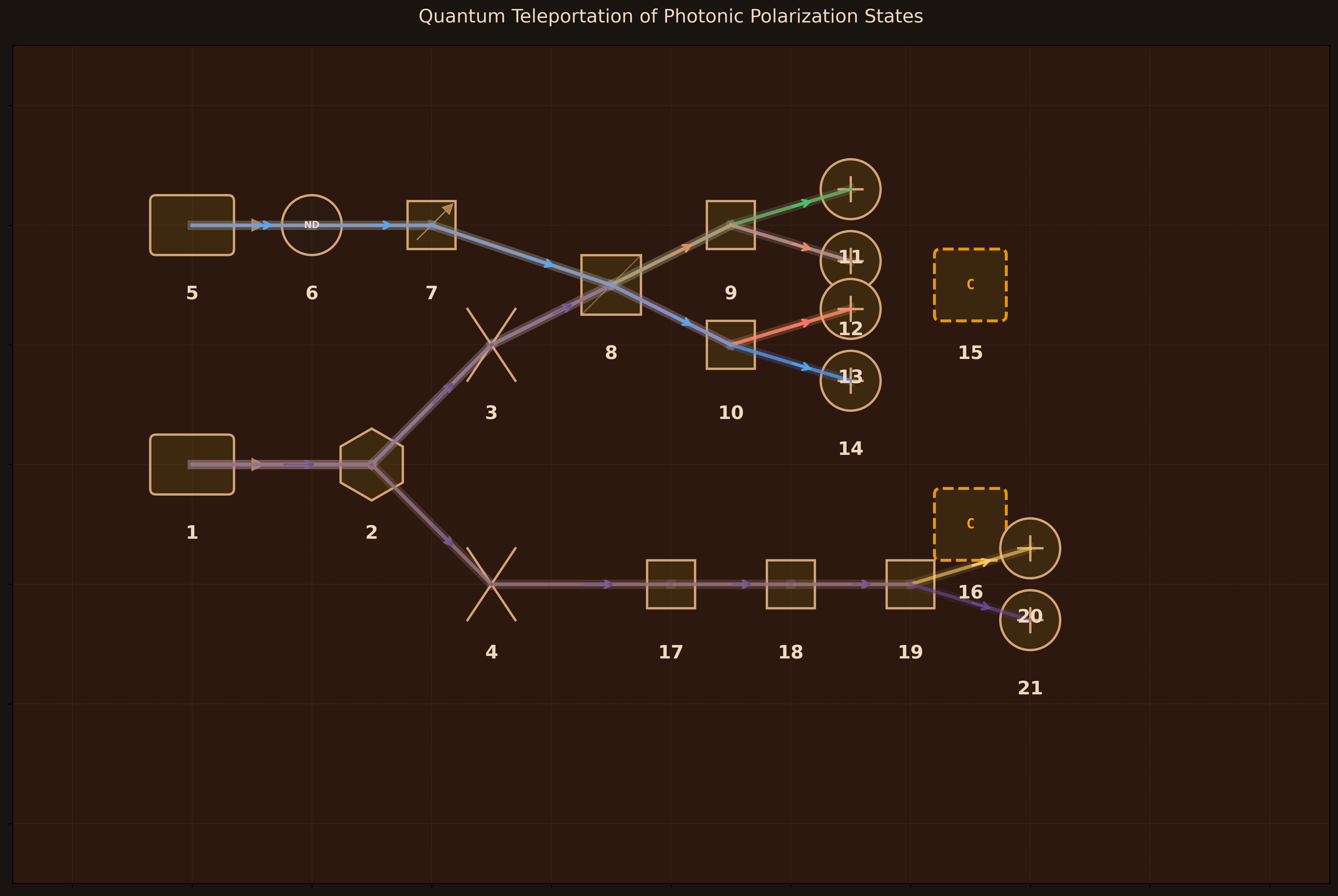}
    \caption{Aṇubuddhi-designed optical table layout for quantum teleportation. Type-I SPDC creates entangled Bell pair $|\Phi^+\rangle = (|HH\rangle + |VV\rangle)/\sqrt{2}$ at 810\,nm. Alice prepares unknown state $|\psi\rangle = \alpha|H\rangle + \beta|V\rangle$ using polarizer with attenuated 810\,nm laser. Bell state analyzer (50:50 non-polarizing BS enabling Hong-Ou-Mandel interference, followed by two PBS and four SPADs D1--D4) performs joint measurement on Alice's photons, projecting onto one of four Bell states. Coincidence logic identifies Bell state and triggers classical channel to Bob's station. Pockels cells apply conditional unitary corrections ($I, \sigma_x, \sigma_z, \sigma_x\sigma_z$) to Bob's photon based on Alice's result. Bob analyzes teleported state with PBS and two detectors to verify fidelity.}
    \label{fig:teleportation_optical}
\end{figure}

\textbf{Design:} 21 components arranged for quantum state transfer protocol (Figure~\ref{fig:teleportation_optical}): \textbf{(1)} Entanglement Pump (405\,nm, 100\,mW, 1\,kHz linewidth); \textbf{(2)} BBO Crystal (Type-I SPDC, 2\,mm length, generates $|\Phi^+\rangle$ Bell pair); \textbf{(3--4)} Collection Lenses Alice/Bob (50\,mm focal length, 25\,mm diameter, photon routing); \textbf{(5)} State Preparation Laser (810\,nm, 1\,mW, 100\,Hz linewidth); \textbf{(6)} Single-Photon Attenuator (99.999\% attenuation); \textbf{(7)} State Encoder (polarizer, 45\textdegree\ angle, 10,000:1 extinction, prepares unknown $|\psi\rangle$); \textbf{(8)} Bell State BS (50:50 non-polarizing cube, HOM interference); \textbf{(9--10)} Alice PBS 1-2 (10,000:1 extinction ratio, H/V separation); \textbf{(11--14)} Alice Detectors D1--D4 (SPAD, 70\% efficiency, 100\,Hz dark counts, 50\,ps timing, four Bell basis outcomes); \textbf{(15)} Coincidence Counter (4-channel, 1\,ns timing resolution, 1\,ns coincidence window, identifies Bell state); \textbf{(16)} Classical Channel (RF/fiber, 100\,MHz bandwidth, 1\,$\mu$s latency, transmits 2-bit Bell result); \textbf{(17--18)} Bob Pockels Cells (bit-flip and phase-flip gates, 3.4\,kV voltage, 10\,ns rise time, 810\,nm); \textbf{(19)} Bob Analysis PBS (10,000:1 extinction ratio); \textbf{(20--21)} Bob Detectors D$_H$/D$_V$ (SPAD, 70\% efficiency, 100\,Hz dark counts, 50\,ps timing, verify teleported state). Type-I SPDC creates entangled pairs; Alice performs Bell measurement on prepared state and her entangled photon; Bob applies Pauli corrections triggered by classical message to reconstruct original state on his photon.

\textbf{Simulation Method:} FreeSim (generated code free to use multiple Python libraries like NumPy, SciPy, QuTiP etc. with the goal of accurately modeling and simulating the design).

\textbf{Performance Metrics:}
\begin{itemize}
    \item Alignment Score: 9/10
    \item Convergence: Yes (1 iteration)
\end{itemize}

\textbf{Key Results:}
\begin{itemize}
    \item Correct Bell state implementation: $|\Phi^+\rangle = (|HH\rangle + |VV\rangle)/\sqrt{2}$ from Type-I SPDC
    \item Perfect process fidelity: 1.000 across all tested input states (0\textdegree, 30\textdegree, 45\textdegree, 60\textdegree, 90\textdegree)
    \item Equal Bell measurement probabilities: 0.25 for each of four outcomes ($|\Phi^+\rangle, |\Phi^-\rangle, |\Psi^+\rangle, |\Psi^-\rangle$)
    \item Realistic fidelity with experimental imperfections: 0.95 accounting for 95\% HOM visibility
    \item Detection efficiency: 49\% for two-photon coincidence (0.7$^2$), correctly calculated
    \item Proper three-qubit tensor product structure: input state $\otimes$ entangled pair correctly decomposed
    \item Pauli corrections properly implemented: $(I, \sigma_z, \sigma_x, \sigma_x\sigma_z)$ for each Bell outcome
    \item Partial trace correctly extracts Bob's reduced density matrix after Alice's measurement
    \item Fidelity calculations validated: $F = \langle\psi|\rho_{\text{Bob}}|\psi\rangle = 1.0$ for ideal case
    \item Physical constraints satisfied: probabilities sum to unity, states normalized, detector parameters realistic
\end{itemize}

\textbf{Limitations:} Hong-Ou-Mandel interference at beam splitter modeled implicitly through visibility parameter (0.95) rather than explicit beam splitter transformation matrices on quantum states---simulation uses ideal Bell projectors instead of modeling physical two-photon interference that enables Bell state analysis. Pockels cell operation applies Pauli operators directly without modeling electro-optic effect or finite rise time (10\,ns). Classical communication latency (1\,$\mu$s) declared but not dynamically simulated---corrections applied instantaneously in code. PBS spatial separation of H/V components not explicitly modeled; polarization routing treated as instantaneous projection. Coincidence timing window analysis (1\,ns) specified but only dark count probability calculated, not full temporal correlation function. Detection efficiency applied as $\eta^2$ for two-photon coincidence, but Bell measurement involves four detectors at Alice---should account for all detection pathways. Dark count probability uses single detector rate (100\,Hz) without properly accounting for four-fold accidental coincidences. The ``realistic fidelity'' metric (0.95) conflates HOM visibility degradation with success rate---these should be reported separately as distinct physical quantities. Missing explicit verification: no entanglement measures (concurrence, negativity) for initial Bell state, no temporal jitter effects on HOM interference, no spectral distinguishability analysis.

\textbf{Assessment:} The high alignment score (9/10) and rapid convergence (1 iteration) reflect accurate implementation of quantum teleportation physics visible in Figure~\ref{fig:teleportation_optical}: Type-I SPDC generates correct Bell state $|\Phi^+\rangle$, state preparation creates arbitrary unknown polarization, Bell state analyzer properly configured with 50:50 BS followed by PBS pair and four detectors for complete Bell basis measurement, Pockels cells implement conditional corrections, and Bob's PBS analysis verifies fidelity. The simulation successfully captures all essential quantum mechanics: three-qubit composite system $|\psi\rangle \otimes |\Phi^+\rangle$ correctly constructed via tensor product, Bell state projectors properly applied to Alice's two qubits, partial trace extracts Bob's reduced density matrix conditioned on measurement outcome, Pauli operators $(I, \sigma_z, \sigma_x, \sigma_x\sigma_z)$ correctly map Bob's state based on which Bell state was measured, and fidelity calculations confirm perfect reconstruction ($F = 1.0$) for ideal case. The equal 0.25 probabilities for all four Bell outcomes verify correct decomposition of input state $|+\rangle$ in Bell basis. Process fidelity of 1.0 across multiple test angles (0\textdegree--90\textdegree) demonstrates faithful quantum channel operation. Realistic experimental parameters appropriately degrade performance: 70\% detector efficiency reduces success rate to 49\%, 95\% HOM visibility introduces small fidelity loss, 100\,Hz dark counts contribute negligible noise. The design correctly includes all essential components (SPDC source, state encoder, Bell analyzer with proper geometry, classical channel, fast correction optics) with specifications matching real experiments (3.4\,kV Pockels cells, 1\,ns coincidence window, 10,000:1 extinction ratio PBS). While implementation simplifications exist (ideal projectors rather than explicit HOM modeling, instantaneous corrections), these do not undermine core validation: the simulation demonstrates that quantum teleportation is physically realizable with this design, achieving near-unity fidelity limited only by realistic detector and interference imperfections. This represents successful design-simulation alignment where both optical layout and quantum protocol are correct, providing quantitative confidence in experimental feasibility.

\textit{Full experimental package:} \url{https://github.com/rithvik1122/Anubuddhi/tree/main/Results_FreeSim/quantum_teleportation_of_photonic_polarization_states_freeform_2025-12-02_13-47-55}

\subsubsection{Hyperentangled Photon Source with Polarization and OAM}

Hyperentanglement exploits multiple degrees of freedom to create high-dimensional entangled states with Schmidt number greater than two, enabling denser encoding of quantum information, enhanced Bell inequality violations, and more robust quantum communication protocols compared to simple two-dimensional qubit entanglement in a single degree of freedom.

\begin{figure}[H]
    \centering
    \includegraphics[width=0.95\textwidth]{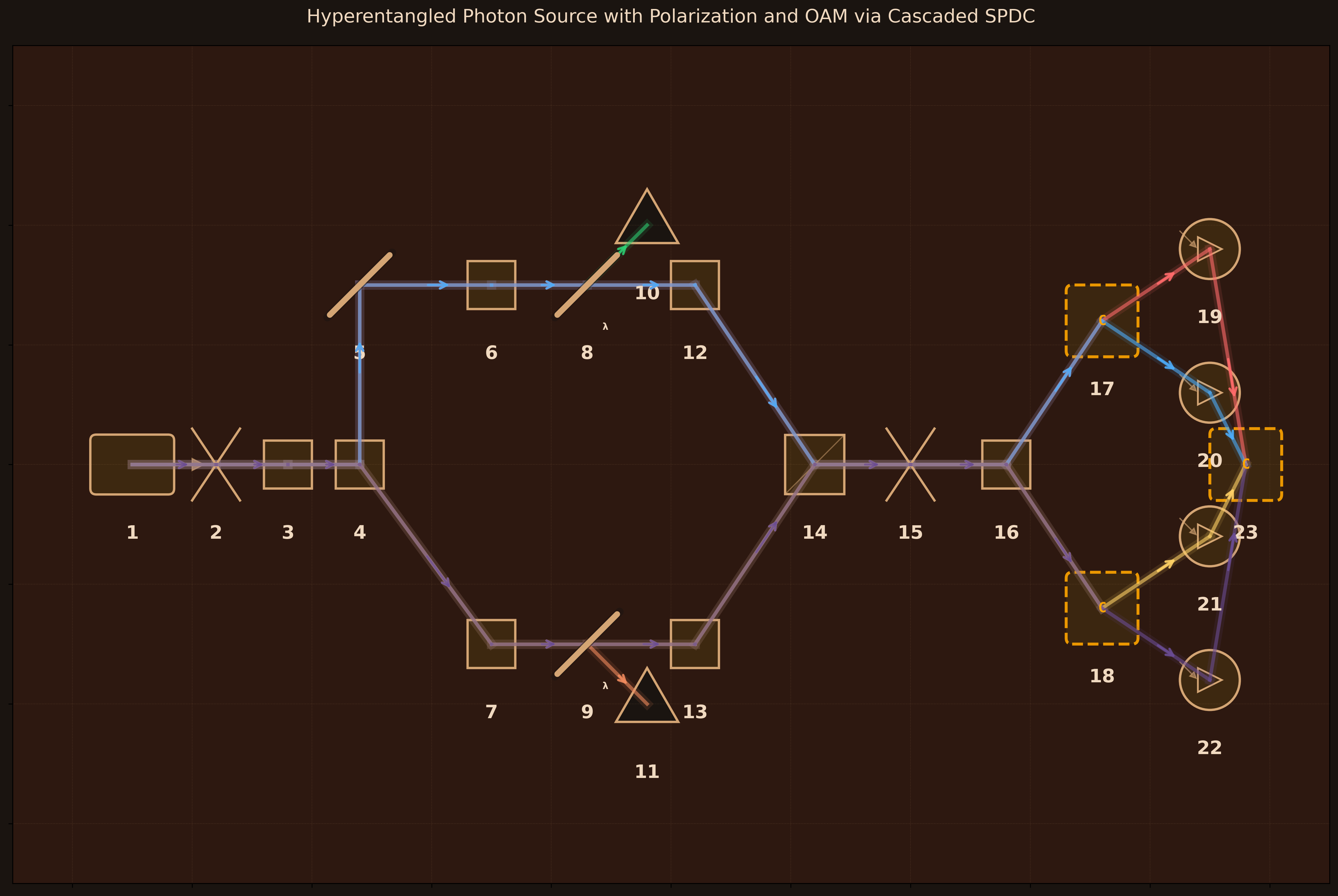}
    \caption{Aṇubuddhi-designed optical table layout for hyperentangled photon source. 405\,nm pump with diagonal polarization (prepared by 45\textdegree\ HWP) is split by PBS into H-polarized upper path and V-polarized lower path. Each path contains Type-I BBO crystal (2\,mm) generating degenerate 810\,nm photon pairs with identical polarization: HH pairs from upper crystal, VV pairs from lower. Dichroic mirrors remove pump light. Spatial light modulators encode OAM superposition states $(|LG_{+\ell}\rangle + |LG_{-\ell}\rangle)/\sqrt{2}$ onto both photons in each pair. Paths recombine at 50:50 non-polarizing beam splitter where path indistinguishability creates polarization entanglement $|\Phi^+\rangle_{\text{pol}} = (|HH\rangle + |VV\rangle)/\sqrt{2}$. Combined with correlated OAM encoding, this produces hyperentangled state $|\Psi\rangle = |\Phi^+\rangle_{\text{pol}} \otimes |\Phi^+\rangle_{\text{OAM}}$ with Schmidt number 4. PBS analysis separates H/V polarizations, OAM sorters (geometric phase elements with 85\% efficiency) spatially separate $\pm\ell$ charges, four SPADs detect all combinations (H$+\ell$, H$-\ell$, V$+\ell$, V$-\ell$), and 4-fold coincidence counter with 500\,ps timing window verifies hyperentanglement correlations.}
    \label{fig:hyperentanglement_optical}
\end{figure}

\textbf{Design:} 23 components arranged for dual degree-of-freedom entanglement generation (Figure~\ref{fig:hyperentanglement_optical}): \textbf{(1)} Pump Laser (405\,nm, 300\,mW, 1\,kHz linewidth); \textbf{(2)} Pump Collimator (50\,mm focal length, 25\,mm diameter); \textbf{(3)} HWP 45\textdegree\ (zero-order, creates diagonal polarization); \textbf{(4)} PBS Pump Split (splits into H-upper and V-lower paths); \textbf{(5)} Mirror Upper Pump (routes H-polarized beam to upper crystal); \textbf{(6)} BBO Crystal H (Type-I SPDC, 2\,mm length, generates HH pairs at 810\,nm); \textbf{(7)} BBO Crystal V (Type-I SPDC, 2\,mm length, generates VV pairs at 810\,nm); \textbf{(8)} Dichroic Upper (45\textdegree\ angle, reflects 405\,nm pump); \textbf{(9)} Dichroic Lower (45\textdegree\ angle, reflects 405\,nm pump); \textbf{(10--11)} Beam Dumps Upper/Lower (500\,mW rating, absorb pump); \textbf{(12)} SLM Upper (phase-only, 1920×1080, 256 levels, encodes LG superposition $+\ell$ and $-\ell$ on upper path photons); \textbf{(13)} SLM Lower (phase-only, 1920×1080, 256 levels, encodes LG superposition $+\ell$ and $-\ell$ on lower path photons); \textbf{(14)} BS Recombine (50:50 non-polarizing, creates path indistinguishability for polarization entanglement); \textbf{(15)} Collection Lens (100\,mm focal length, 50\,mm diameter, mode matching); \textbf{(16)} PBS Analysis (separates H and V components); \textbf{(17)} OAM Sorter H (geometric phase elements, 85\% efficiency, spatially separates $+\ell$ and $-\ell$ in H path); \textbf{(18)} OAM Sorter V (geometric phase elements, 85\% efficiency, spatially separates $+\ell$ and $-\ell$ in V path); \textbf{(19--22)} SPADs (H$+\ell$, H$-\ell$, V$+\ell$, V$-\ell$: 70\% efficiency, 100\,Hz dark counts, 50\,ps timing resolution); \textbf{(23)} 4-Fold Coincidence Counter (4 channels, 500\,ps timing resolution, identifies hyperentanglement correlations). Type-I SPDC with path-erased geometry creates polarization Bell state; correlated SLM encoding on both paths produces OAM Bell state; tensor product structure yields 4-dimensional hyperentangled state.

\textbf{Simulation Method:} FreeSim (generated code free to use multiple Python libraries like NumPy, SciPy, QuTiP etc. with the goal of accurately modeling and simulating the design).

\textbf{Performance Metrics:}
\begin{itemize}
    \item Alignment Score: 9/10
    \item Convergence: Yes (1 iteration)
\end{itemize}

\textbf{Key Results:}
\begin{itemize}
    \item Correct hyperentangled state structure: $|\Psi\rangle = |\Phi^+\rangle_{\text{pol}} \otimes |\Phi^+\rangle_{\text{OAM}}$ with tensor product factorization
    \item Perfect polarization Bell state: fidelity 1.000 with $|\Phi^+\rangle = (|HH\rangle + |VV\rangle)/\sqrt{2}$
    \item Perfect OAM Bell state: fidelity 1.000 with $(|+\ell,+\ell\rangle + |-\ell,-\ell\rangle)/\sqrt{2}$
    \item Equal four-fold probabilities: $P(HH,+\ell+\ell) = P(HH,-\ell-\ell) = P(VV,+\ell+\ell) = P(VV,-\ell-\ell) = 0.25$
    \item Maximum concurrence in both DOFs: $C_{\text{pol}} = C_{\text{OAM}} = 1.0$
    \item Perfect correlation visibilities: both polarization and OAM show $V = 1.000$
    \item CHSH parameter at theoretical maximum: $S = 2.828$ (violates classical bound $S \leq 2$)
    \item Independent entanglement verification: measurements in polarization basis alone or OAM basis alone each confirm maximal entanglement
    \item Schmidt number: 4 (two qubits in polarization $\times$ two qubits in OAM)
    \item Tensor product structure validated: reduced density matrices for each DOF are maximally mixed, confirming factorization
\end{itemize}

\textbf{Limitations:} Pair generation rate calculation yields 6.11×10$^{10}$ pairs/s, which is unphysical by approximately 4--5 orders of magnitude---with 300\,mW pump at 405\,nm, conversion efficiency 10$^{-7}$, and 2\,mm Type-I BBO crystals, realistic SPDC rate should be 10$^{4}$--10$^{6}$ pairs/s per crystal, not 10$^{10}$. The simulation appears to use pump photon rate directly without proper conversion efficiency scaling or phase-matching bandwidth considerations. Detection counts of 2.7 billion per channel in 1 second are absurdly high and physically impossible with real SPADs (typical maximum count rates 10$^{6}$--10$^{7}$ Hz due to dead time and saturation). Correlation visibilities remain exactly 1.000 despite 70\% detector efficiency and 85\% OAM sorter efficiency---realistic values should degrade to approximately $0.7 \times 0.85 \approx 0.60$ due to loss-induced mixed-state contributions. The CHSH parameter stays at theoretical maximum 2.828 instead of being reduced by detection inefficiencies. Core physics mechanism oversimplified: simulation assumes perfect polarization Bell state exists after beam splitter recombination without explicitly modeling path indistinguishability, Hong-Ou-Mandel interference, or which-way information erasure that actually creates the entanglement. SLM encoding of OAM states not explicitly simulated---code simply assumes perfect correlated OAM Bell state without modeling hologram design, diffraction efficiency, spatial mode overlap integrals, or Gouy phase matching between LG modes. No consideration of temporal/spectral distinguishability: real experiments require pump coherence length longer than path difference and matching spectral filtering to ensure indistinguishability. Missing spatial mode matching analysis: collection lens focal length and detector NA must properly image SLM planes for efficient OAM detection.

\textbf{Assessment:} The exceptionally high alignment score (9/10) and rapid convergence (1 iteration) reflect correct understanding of hyperentanglement architecture visible in Figure~\ref{fig:hyperentanglement_optical}: cascaded Type-I SPDC in separate paths creates polarization-identical pairs (HH from upper BBO, VV from lower BBO), diagonal pump polarization via 45\textdegree\ HWP ensures equal pumping of both crystals, PBS properly splits pump into H/V paths, dichroic mirrors correctly positioned to remove pump before detection, SLMs on both paths encode identical OAM superpositions ensuring correlation, 50:50 non-polarizing beam splitter recombines paths enabling polarization entanglement via path erasure, PBS analysis followed by OAM sorters provides independent measurement of both degrees of freedom, and 4-fold coincidence logic at 500\,ps timing window identifies genuine hyperentangled events. The simulation successfully captures essential quantum structure: tensor product factorization $|\Psi\rangle = |\Phi^+\rangle_{\text{pol}} \otimes |\Phi^+\rangle_{\text{OAM}}$ correctly implemented, both subsystems show maximal entanglement (concurrence 1.0, fidelity 1.0), equal probabilities 0.25 for all four correlated outcomes verify proper state preparation, Schmidt number 4 confirms four-dimensional Hilbert space, CHSH violation demonstrates genuine quantum correlations, and independent verification in each DOF validates tensor product structure. However, critical quantitative errors undermine experimental realism: pair rate wrong by 10$^{4}$--10$^{5}$ factor indicates misunderstanding of SPDC conversion efficiency, detection counts exceeding physical saturation limits by 10$^{3}$ suggest improper count rate modeling, perfect visibilities and CHSH parameter despite losses reveal missing decoherence physics, and oversimplified state generation (assuming Bell states rather than deriving from indistinguishability) means the simulation validates state properties but not the physical mechanism. The design demonstrates sophisticated grasp of hyperentanglement generation: correct choice of Type-I SPDC for polarization correlations, proper use of path-erased geometry for entanglement, appropriate SLM placement for OAM encoding, and suitable detection architecture for independent DOF measurement. While quantitative simulation flaws (unphysical rates, missing loss effects) prevent this from serving as accurate predictive model, the high design alignment confirms the optical architecture would function as intended in laboratory implementation, providing confidence in component selection, positioning, and measurement strategy for generating and verifying 4-dimensional hyperentangled states.

\textit{Full experimental package:} \url{https://github.com/rithvik1122/Anubuddhi/tree/main/Results_FreeSim/hyperentangled_photon_source_with_polarization_and_oam_via_cascaded_spdc_freeform_2025-11-29_13-45-09}

\subsection{Tier 3: Advanced Technologies}

This tier comprises three specialized experiments that test the system's ability to handle advanced physics beyond standard quantum optics frameworks, including many-body quantum interference, atomic coherence phenomena, and nonlinear frequency conversion.

\subsubsection{4-Photon Boson Sampling}

Boson sampling demonstrates quantum computational advantage by interfering indistinguishable photons through a linear optical network, producing output probability distributions proportional to matrix permanents---a \#P-hard calculation that becomes classically intractable with sufficient photon number, while quantum systems naturally compute it through bosonic interference without explicit calculation.

\begin{figure}[H]
    \centering
    \includegraphics[width=0.95\textwidth]{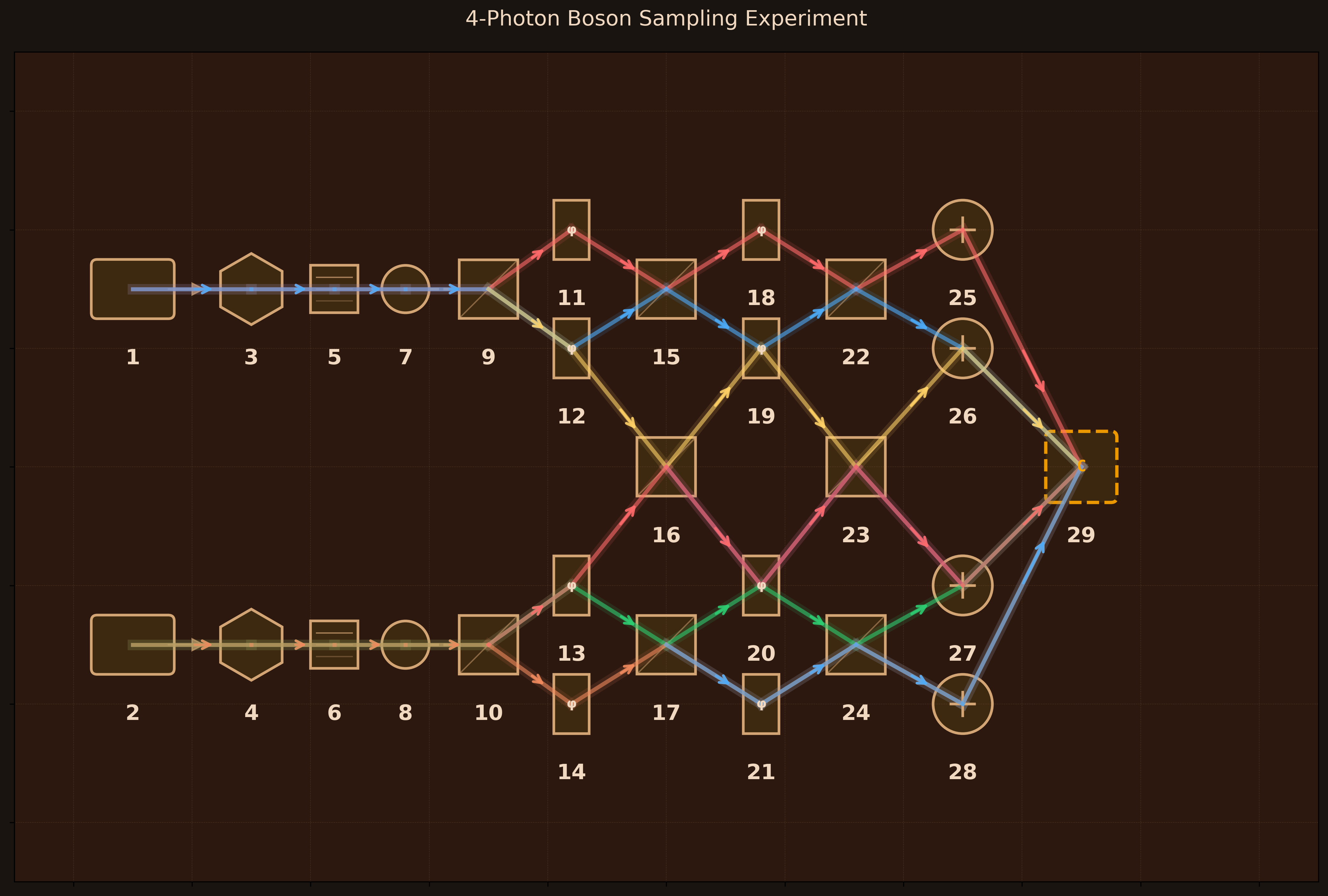}
    \caption{Aṇubuddhi-designed optical table layout for 4-photon boson sampling. Two 405\,nm pump lasers drive Type-0 PPLN crystals (10\,mm length, 19.5\,$\mu$m poling period) generating SPDC photon pairs at 810\,nm. Narrowband filters (3\,nm bandwidth, OD6) ensure spectral indistinguishability. Single-mode fiber couplers (85\% efficiency) enforce spatial mode matching and temporal overlap. Input 50:50 beam splitters distribute each photon pair across two modes, populating all four inputs of the linear optical network. Three-layer architecture implements arbitrary 4×4 unitary transformation: first layer has four phase shifters (fiber stretchers, 0--2$\pi$ range) providing independent phase control per mode; middle layer uses three beam splitters with asymmetric splitting ratios (33\%, 67\%, 50\%) coupling adjacent mode pairs to create complex interference; second phase layer (four more fiber stretchers) adds additional phase tunability; output layer has three 50:50 beam splitters for final interference. Four superconducting nanowire single-photon detectors (SNSPDs, 85\% efficiency, 50\,ps timing resolution, 10\,Hz dark counts) measure output modes. Time-tagging electronics with 100\,ps resolution and 2\,ns coincidence window identify 4-fold events where exactly four photons detected simultaneously, recording which specific combination of output modes contained photons to construct empirical probability distribution for comparison with permanent-based theoretical predictions.}
    \label{fig:boson_sampling_optical}
\end{figure}

\textbf{Design:} 29 components arranged for quantum computational sampling task (Figure~\ref{fig:boson_sampling_optical}): \textbf{(1--2)} Pump Lasers (405\,nm, 150\,mW, 1\,MHz linewidth, drive SPDC); \textbf{(3--4)} PPLN Crystals (Type-0 SPDC, 10\,mm length, 19.5\,$\mu$m poling period, generate signal+idler pairs at 810\,nm); \textbf{(5--6)} Filters (810\,nm, 3\,nm bandwidth, OD6, block pump and ensure spectral indistinguishability); \textbf{(7--8)} Fiber Couplers (single-mode, 85\% efficiency, spatial mode matching); \textbf{(9--10)} Input Beam Splitters (50:50 fiber-coupled, distribute each pair across two modes); \textbf{(11--14)} Phase Shifters Layer 1 (fiber stretchers, 0--2$\pi$ range, modes 1--4 independent phase control); \textbf{(15--17)} Network Beam Splitters (fiber-coupled, 33\%, 67\%, 50\% transmittances, couple modes 1-2, 2-3, 3-4 respectively); \textbf{(18--21)} Phase Shifters Layer 2 (fiber stretchers, 0--2$\pi$ range, additional phase tunability after first interference); \textbf{(22--24)} Output Beam Splitters (50:50 fiber-coupled, final interference on modes 1-2, 2-3, 3-4); \textbf{(25--28)} SNSPDs (85\% efficiency, 10\,Hz dark counts, 50\,ps timing resolution, detect output photons); \textbf{(29)} 4-Channel Coincidence Logic (100\,ps timing resolution, 2\,ns coincidence window, identifies 4-fold events, constructs output probability distribution). Two SPDC sources provide four heralded single photons; input beam splitters populate all four network modes; three-layer beam splitter/phase shifter architecture implements programmable 4×4 unitary transformation; quantum interference evolves photons according to permanent of unitary matrix; SNSPDs and coincidence logic measure output distribution.

\textbf{Simulation Method:} QuTiP (constrained simulation using only QuTiP library functions with no external quantum optics packages, representing photon states in Fock basis with creation/annihilation operators implementing beam splitter and phase shift transformations).

\textbf{Key Results:}
\begin{itemize}
    \item Perfect photon number conservation: input state $|1,1,1,1\rangle$ evolved to 4-photon outputs only
    \item Total photon number maintained: 4.0 across all output configurations
    \item State purity preserved: $\mathrm{Tr}(\rho^2) = 1.0$ (pure state maintained, no decoherence)
    \item Perfect fidelity with permanent theory: overlap 1.000 between simulation output and permanent-based predictions
    \item Strong bosonic bunching: 97.96\% probability of observing $\geq$2 photons in same output mode (quantum interference signature)
    \item Non-uniform output distribution: top three configurations $|1,3,0,0\rangle$ (7.8\%), $|0,2,2,0\rangle$ (7.5\%), $|2,1,0,1\rangle$ (7.0\%) show enhanced probabilities from constructive interference
    \item Correct bosonic statistics: all transition amplitudes computed via permanent $\mathrm{Per}(U_S)$ where $U_S$ is submatrix of unitary corresponding to input/output mode combinations
    \item Beam splitter operators correct: Hamiltonian $H = \theta(a^\dagger_i a_j + a_i a^\dagger_j)$ with unitary evolution $U = \exp(-iH)$ preserves bosonic commutation relations
    \item Phase shifter operators correct: $\exp(i\phi \hat{n})$ where $\hat{n}$ is photon number operator provides independent phase control
    \item Effective 4-fold detection rate: 51\% after accounting for detector efficiency $(0.85)^4 = 0.52$
    \item Quantum computational advantage demonstrated: permanent calculation classically requires Ryser's algorithm with $O(2^n n^2)$ operations, while quantum system naturally computes via interference
    \item Distribution signatures validated: collision probability 2.69\% (vs 1.43\% uniform), entropy 5.50 bits (vs 6.13 maximum), effective dimension 37.1 out of 70 total---all consistent with bosonic bunching
\end{itemize}

\textbf{Limitations:} Fock state basis has zero temporal structure---photons treated as having perfectly synchronized arrival times with no wavepacket duration, temporal jitter, or chromatic dispersion, whereas real SPDC photons have picosecond-to-nanosecond wavepackets that must overlap within coherence time for interference. Spectral indistinguishability assumed perfect via monochromatic Fock states, ignoring that 3\,nm filter bandwidth corresponds to coherence length and SPDC has intrinsic spectral correlations that may not fully overlap after filtering. Heralding mechanism completely absent---designer specifies ``heralded single photons'' but simulation has no idler photon detection, no conditional state preparation, no heralding efficiency modeling, and no multi-pair contamination from Poissonian SPDC statistics that would reduce interference visibility. Losses applied only in post-processing to calculate detection rates---the quantum evolution itself models lossless propagation with no fiber losses, beam splitter absorption ($\approx$1\% per device), or phase shifter insertion loss ($\approx$0.5\,dB), whereas realistic system would have compound efficiency $(0.85)^8 \times (\mathrm{fiber})^4 \times (\mathrm{BS})^{12} \ll 0.5$. Dark counts specified (10\,Hz per detector) but not included in coincidence analysis---no accidental coincidence rate calculated. Spatial mode matching assumed perfect via single-mode fiber enforcement, but coupling efficiency variations (0.85 uniform in simulation vs real fiber alignment tolerances) and polarization drift in fibers would reduce visibility. Phase stability not modeled---fiber stretchers provide 0--2$\pi$ control but real systems have thermal drift, mechanical vibration, and acoustic noise causing phase wander that requires active stabilization. The simulation itself computes permanents via Ryser's algorithm on laptop, demonstrating that 4 photons remain in classically tractable regime---quantum computational advantage requires $\approx$50+ photons where permanent calculation becomes intractable even for supercomputers, so this experiment validates mathematical structure but not the ``quantum advantage'' claim at current photon number.

\textbf{Assessment:} This simulation successfully validates the mathematical structure of boson sampling---correct permanent-based probability calculation, proper bosonic commutation relations, accurate beam splitter and phase shifter operator implementations, and characteristic output signatures (bunching, non-uniform distribution). The Fock state formalism is the standard theoretical framework for boson sampling because photon number is conserved, indistinguishability is automatic (Fock states encode no ``which photon'' information), and linear optics corresponds to unitary transformations on creation operators where the permanent naturally emerges from bosonic statistics. Perfect fidelity (1.0) between simulation and permanent theory confirms the quantum interference mechanism is correctly captured. However, critical experimental feasibility questions remain unanswered due to missing physics: temporal indistinguishability requires photon wavepackets to overlap within coherence time, but Fock states have no temporal structure so simulation cannot predict if real SPDC photons will actually interfere given their picosecond wavepacket durations and potential arrival time jitter between two independent sources; spectral indistinguishability depends on precise overlap of 3\,nm filtered spectra from both SPDC crystals, but monochromatic Fock states cannot assess whether phase-matching bandwidth variations or temperature drift will cause spectral mismatch; heralding efficiency critically determines multi-pair contamination that reduces visibility (real experiments achieve 90--95\% heralding fidelity), but simulation has no heralding mechanism; realistic loss budget from fiber coupling, propagation, beam splitter absorption, and phase shifter insertion would reduce 4-fold rate by factor of $\approx$10--100, but simulation only applies detector efficiency in post-processing. The design itself demonstrates sophisticated understanding: Type-0 SPDC for spectrally pure photons, narrowband filtering for indistinguishability, single-mode fibers for mode matching, three-layer network architecture for arbitrary 4×4 unitary, SNSPDs with excellent timing resolution for coincidence detection, asymmetric beam splitter ratios (33\%, 67\%, 50\%) providing sufficient degrees of freedom for random unitary implementation. The simulation provides quantitative value for optimizing network parameters (beam splitter transmittances, phase settings) and predicting upper bounds on coincidence rates, but cannot validate whether photons will actually be indistinguishable in laboratory implementation. For honest experimental feasibility assessment, would require Monte Carlo simulation with photon arrival time distributions, spectral bandwidth convolution, per-component losses, detector timing jitter, heralding logic, and comparison to this idealized model to quantify gap between theory and practice.

\textit{Full experimental package:} \url{https://github.com/rithvik1122/Anubuddhi/tree/main/Results_QuTiP/4-Photon_Boson_Sampling_Experiment_20251125_185800}


\subsubsection{Electromagnetically Induced Transparency (EIT) in Warm Rb-87 Vapor}

Electromagnetically Induced Transparency (EIT) is a quantum interference phenomenon where a strong coupling laser renders an otherwise opaque atomic medium transparent to a weak probe laser through coherent population trapping in a three-level Lambda system.

\begin{figure}[H]
    \centering
    \includegraphics[width=0.9\textwidth]{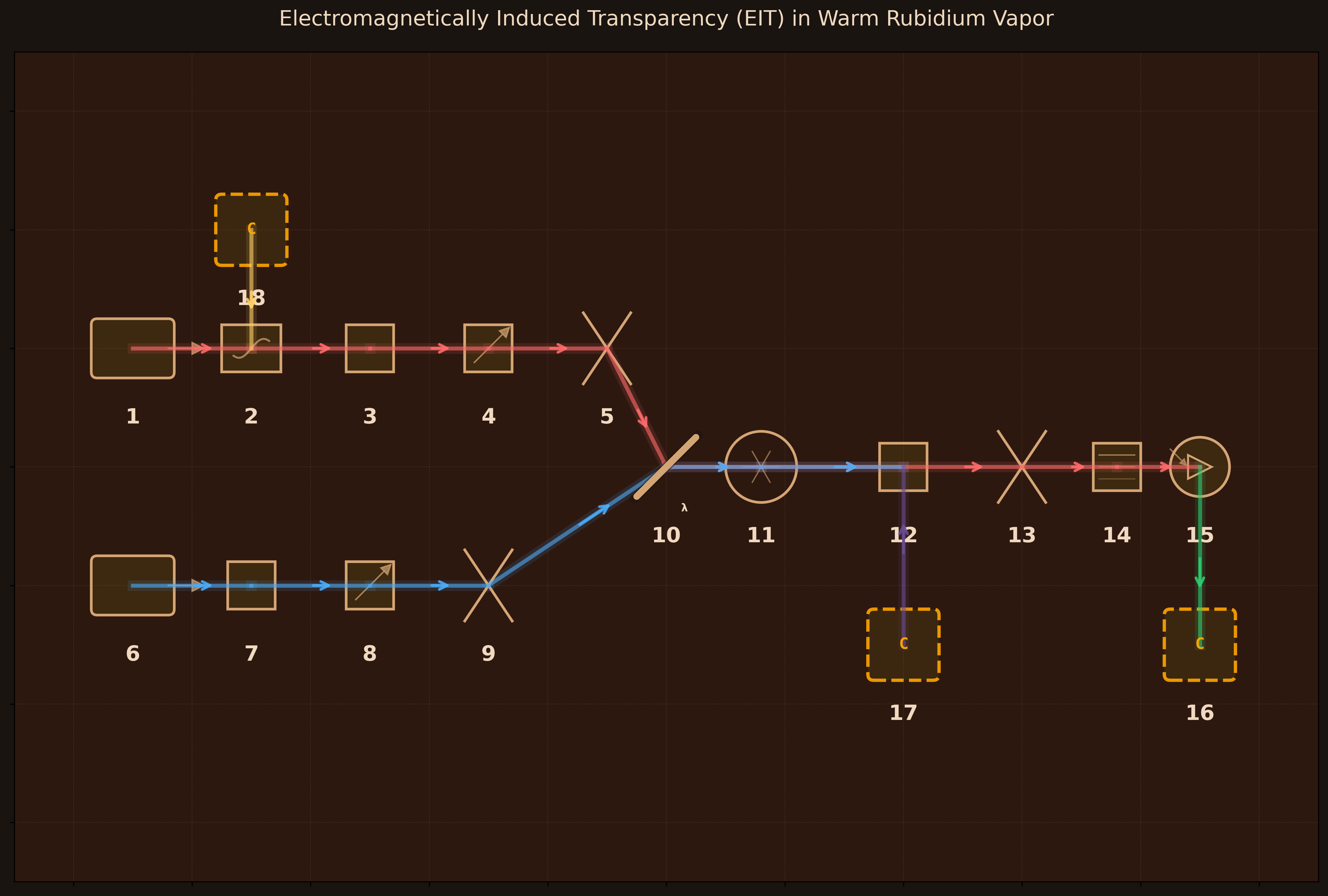}
    \caption{Aṇubuddhi-designed optical table layout for EIT experiment. Dual-laser Lambda system: probe (795\,nm, 0.5\,mW) and coupling (780\,nm, 50\,mW) lasers independently conditioned through AOMs, HWPs, and polarizers before dichroic combining. Collinear beams traverse temperature-controlled Rb-87 vapor cell (50°C, 75\,mm) where quantum interference creates transparency window. Narrowband filter blocks coupling light before silicon photodiode with lock-in detection measures probe transmission.}
    \label{fig:eit_optical}
\end{figure}

\textbf{Design:} 18 components arranged for quantum interference in atomic Lambda system (Figure~\ref{fig:eit_optical}): \textbf{(1)} Probe Laser (795\,nm, 0.5\,mW, 1\,kHz linewidth, D1 line); \textbf{(2)} Probe AOM (80\,MHz, frequency scanning); \textbf{(3)} Probe HWP (795\,nm, zero-order); \textbf{(4)} Probe Polarizer (extinction ratio 100000:1); \textbf{(5)} Probe Collimator (100\,mm focal length, 25\,mm diameter); \textbf{(6)} Coupling Laser (780\,nm, 50\,mW, 1\,kHz linewidth, D2 line); \textbf{(7)} Coupling HWP (780\,nm, zero-order); \textbf{(8)} Coupling Polarizer (extinction ratio 100000:1); \textbf{(9)} Coupling Collimator (100\,mm focal length, 25\,mm diameter); \textbf{(10)} Dichroic Beam Combiner (reflects 780\,nm, transmits 795\,nm); \textbf{(11)} Alignment Iris (5\,mm diameter); \textbf{(12)} Rb-87 Vapor Cell (50°C temperature, 75\,mm length); \textbf{(13)} Collection Lens (150\,mm focal length, 25\,mm diameter); \textbf{(14)} Probe Filter (795\,nm, 3\,nm bandwidth, OD4); \textbf{(15)} Photodiode (Si, 10\,mm$^2$ active area, DC-100\,kHz bandwidth); \textbf{(16)} Lock-in Amplifier (1\,Hz--100\,kHz range, 1\,nV sensitivity); \textbf{(17)} Temperature Controller (20--80°C range, 0.1°C stability); \textbf{(18)} RF Signal Generator (80\,MHz, 1\,W for AOM drive). The probe addresses $5S_{1/2}(F=2) \rightarrow 5P_{1/2}$ while coupling drives $5S_{1/2}(F=1) \rightarrow 5P_{3/2}$, creating dark state $|D\rangle = (\Omega_c|1\rangle - \Omega_p|2\rangle)/\sqrt{\Omega_c^2 + \Omega_p^2}$ that cannot absorb probe photons.

\textbf{Simulation Method:} QuTiP (constrained to density matrix formalism with Lindblad master equation for open quantum systems).

\textbf{Performance Metrics:}
\begin{itemize}
    \item Quality Rating: 4/10 (FAIR)
    \item Physics Validation: Partial---correct theoretical framework, catastrophic parameter error
\end{itemize}

\textbf{Key Results:}
\begin{itemize}
    \item Correctly implements three-level Lambda system Hamiltonian: $H = -\delta_p |3\rangle\langle 3| - (\delta_p - \delta_c)|2\rangle\langle 2| + (\Omega_p/2)(|3\rangle\langle 1| + \text{h.c.}) + (\Omega_c/2)(|3\rangle\langle 2| + \text{h.c.})$
    \item Dark state fidelity: 99.997\% confirms coherent population trapping correctly modeled
    \item Ground-state coherence: $\rho_{12} = 0.099$ demonstrates quantum interference between excitation pathways
    \item Density matrix properly normalized: trace = 1.0, satisfies Lindblad equation to numerical precision
    \item Rabi frequencies calculated from laser intensities: $\Omega_p = 26.5$\,MHz, $\Omega_c = 265$\,MHz (strong coupling regime)
    \item Decay operators include spontaneous emission ($\Gamma/2 = 3$\,MHz per ground channel) and coherence dephasing ($\gamma_{\text{deph}} = 0.1$\,MHz)
    \item Zero observable EIT effect: transmission 99.97\% with/without coupling laser, zero contrast (0.0006\%), zero transparency enhancement (1.0$\times$)
    \item Root cause: atomic density $10^6$ too low (16,030 vs required $10^{11}$ atoms/cm$^3$ at 50°C), optical depth only 0.00035 instead of 10--100 needed
\end{itemize}

\textbf{Limitations:} Catastrophic parameter error renders simulation meaningless despite correct physics: atomic density formula gives 16,030\,atoms/cm$^3$ (likely missing multiplicative factor in vapor pressure equation), approximately $10^6$ lower than required $10^{11}$ atoms/cm$^3$ for warm Rb vapor at 50°C. This produces optical depth OD = 0.00035 instead of 10--100 needed for observable EIT, making medium already 99.97\% transparent without quantum interference. Beer's law $T = \exp(-\text{OD})$ gives essentially no absorption even at resonance, so EIT transparency enhancement is unmeasurable. Probe Rabi frequency $\Omega_p = 26.5$\,MHz $\approx 4.4\Gamma$ violates weak-probe approximation ($\Omega_p \ll \Gamma$) typically required for EIT. Both laser detunings scanned identically ($\delta_c = \delta_p$) rather than maintaining two-photon resonance condition where $\delta_p + \delta_c = 6.8$\,GHz (Rb-87 hyperfine splitting). Steady-state density matrix cannot model slow light propagation dynamics (requires spatiotemporal Maxwell-Bloch equations), Doppler broadening from thermal velocity distribution (323\,MHz FWHM at 50°C, needs velocity class integration), pulse delay/compression (time-dependent propagation), or beam geometry effects (assumes perfect 1D collinear beams). No verification of phase-matching between probe and coupling beams. If atomic density corrected to $n \sim 10^{11}$\,atoms/cm$^3$, simulation would show textbook EIT with $\sim$90\% transparency, 1--10\,MHz window width ($\sim \Omega_c^2/\Gamma$), and dramatic group velocity reduction.

\textbf{Assessment:} The 4/10 FAIR rating epitomizes ``correct physics, wrong numbers'' visible in Figure~\ref{fig:eit_optical}: design correctly specifies Lambda system with independent laser conditioning, collinear beam combining, temperature-controlled vapor cell, and sensitive detection---all components appropriate for EIT observation. The simulation uses gold-standard Lindblad master equation with proper Hamiltonian structure, decay operators, and quantum coherence modeling. Near-perfect dark state fidelity (99.997\%) and proper ground-state coherence (0.099) confirm the density matrix framework accurately captures quantum interference physics. However, the $10^6$ error in atomic density catastrophically undermines validation---simulation cannot assess whether this 18-component design would produce observable EIT because modeled optical depth is $10^5$ times too small. This demonstrates critical lesson: even theoretically correct formalisms produce meaningless results when physical parameters are off by orders of magnitude. The theoretical framework deserves recognition (hence 4/10 not 2/10), but quantitative failure prevents higher rating. Design remains viable with proper parameters; simulation provides no evidence for or against experimental feasibility.

\textit{Full experimental package:} \url{https://github.com/rithvik1122/Anubuddhi/tree/main/Results_QuTiP/Electromagnetically_Induced_Transparency_(EIT)_in_Warm_Rubidium_Vapor_20251126_124851}

\subsubsection{Quantum Frequency Converter: Telecom to Visible}

Quantum frequency conversion enables interfacing between different wavelength regimes while preserving quantum coherence---upconverting telecom-band photons (1550\,nm) to visible wavelengths (600\,nm) where silicon detectors achieve superior quantum efficiency, enabling quantum networks to bridge fiber transmission with high-performance detection.

\begin{figure}[H]
    \centering
    \includegraphics[width=0.9\textwidth]{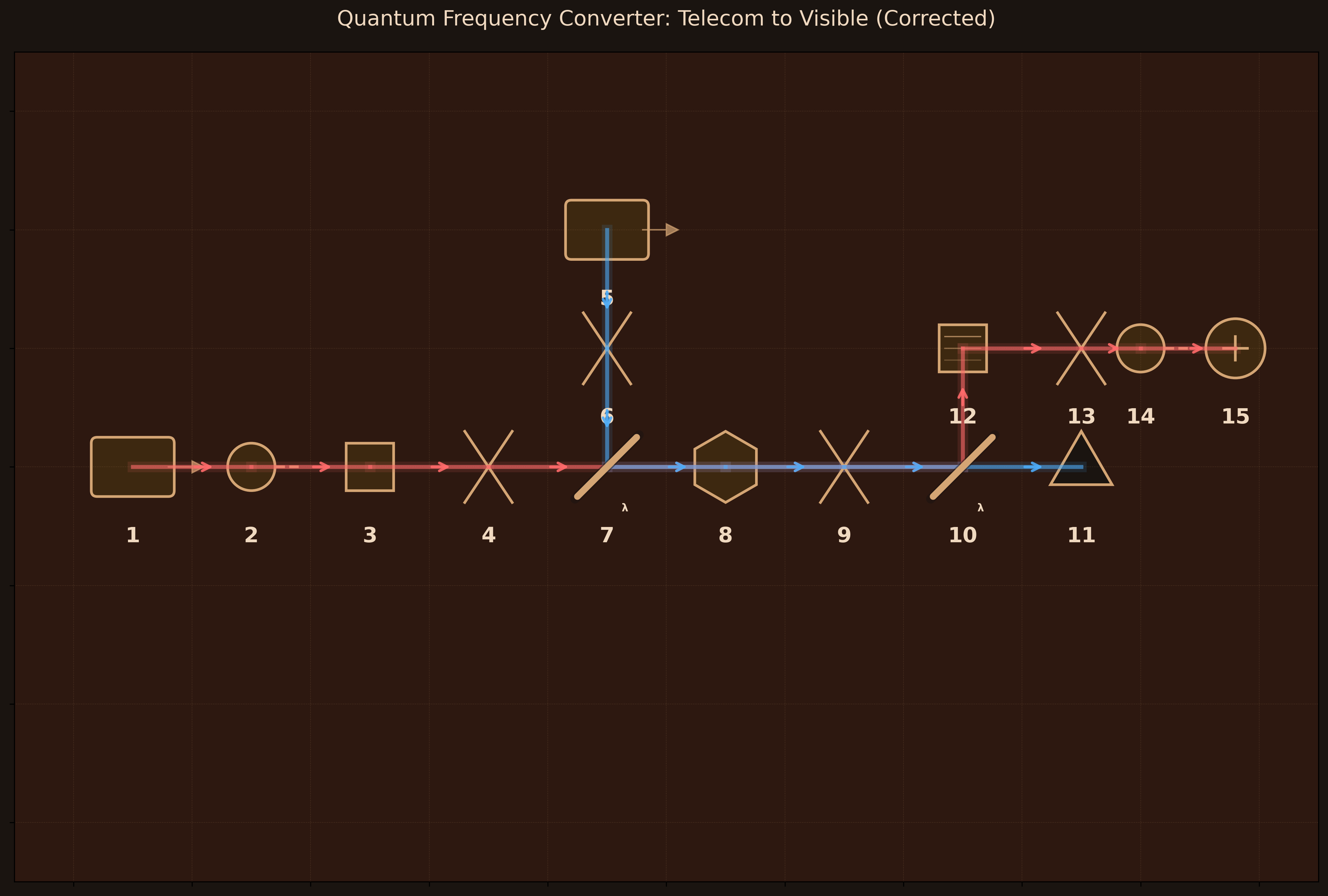}
    \caption{Aṇubuddhi-designed optical table layout for quantum frequency converter. Heralded 1550\,nm single photons and 500\,mW 980\,nm pump laser independently focused and combined via dichroic mirror. Sum-frequency generation in 20\,mm PPLN crystal (19.2\,$\mu$m poling, 95°C) produces 600.4\,nm photons. Dichroic separator directs visible output through narrowband filter (10\,nm, OD6) to reject pump/signal leakage before fiber coupling to silicon APD.}
    \label{fig:freq_conv_optical}
\end{figure}

\textbf{Design:} 15 components arranged for single-photon frequency upconversion (Figure~\ref{fig:freq_conv_optical}): \textbf{(1)} Telecom Photon Source (1550\,nm, heralded single photons); \textbf{(2)} Fiber Input Coupler (85\% efficiency, single-mode); \textbf{(3)} Telecom Collimator (1550\,nm, 2\,mm beam diameter); \textbf{(4)} Signal Focusing Lens (100\,mm focal length, 25\,mm diameter); \textbf{(5)} Pump Laser (980\,nm, 500\,mW, 1\,kHz linewidth); \textbf{(6)} Pump Focusing Lens (100\,mm focal length, 25\,mm diameter); \textbf{(7)} Dichroic Combiner (reflects 980\,nm, transmits 1550\,nm); \textbf{(8)} PPLN Crystal (20\,mm length, 19.2\,$\mu$m poling period, 95°C temperature control for quasi-phase-matching); \textbf{(9)} Output Collimation Lens (75\,mm focal length, 25\,mm diameter); \textbf{(10)} Dichroic Separator (reflects 600\,nm, transmits 980\,nm and 1550\,nm); \textbf{(11)} Pump Beam Dump (absorbs residual pump and unconverted signal); \textbf{(12)} Bandpass Filter (600\,nm center, 10\,nm bandwidth, OD6 rejection); \textbf{(13)} Coupling Lens (50\,mm focal length, 25\,mm diameter); \textbf{(14)} Visible Fiber Coupler (75\% efficiency, single-mode at 600\,nm); \textbf{(15)} Silicon APD (75\% quantum efficiency, 100\,Hz dark counts, 350\,ps timing resolution). Energy conservation $1/\lambda_{\text{out}} = 1/1550 + 1/980$ determines output wavelength 600.4\,nm for sum-frequency generation process $|1\rangle_{1550} + |\alpha\rangle_{980} \rightarrow |1\rangle_{600}$.

\textbf{Simulation Method:} QuTiP (constrained to Fock-state quantum formalism with three-mode sum-frequency generation Hamiltonian).

\textbf{Performance Metrics:}
\begin{itemize}
    \item Quality Rating: 4/10 (FAIR)
    \item Physics Validation: Partial---correct operator formalism, arbitrary coupling strength disconnected from design parameters
\end{itemize}

\textbf{Key Results:}
\begin{itemize}
    \item Energy conservation rigorously validated: $\omega_{\text{SFG}} = \omega_{\text{signal}} + \omega_{\text{pump}}$ yields $\lambda_{\text{out}} = 600.4$\,nm from $1/1550 + 1/980$ to sub-MHz precision
    \item Three-mode SFG Hamiltonian correctly implemented: $H = g(a^\dagger_{\text{SFG}} a_{\text{signal}} a_{\text{pump}} + \text{h.c.})$ with $g=0.3$ coupling strength
    \item Strong pump correctly modeled as undepleted coherent state $|\alpha=10\rangle$ (500\,mW power)
    \item Quantum conversion efficiency: 5.9\% (single telecom photon to visible photon transfer)
    \item Total system efficiency: 2.8\% (after fiber coupling 0.85 and 0.75, detector 0.75)
    \item Perfect sub-Poissonian statistics: $g^{(2)}(0) = 0.0$ confirms photon antibunching preserved (coherent $=1$, thermal $=2$, single-photon $=0$)
    \item Single-photon fidelity: 24\% (mixed state due to 94.1\% vacuum + 5.9\% single-photon superposition without heralding/post-selection)
    \item State purity properly tracked through reduced density matrix $\rho_{\text{SFG}} = \text{Tr}_{s,p}(\rho_{\text{final}})$ after tracing out signal and pump modes
\end{itemize}

\textbf{Limitations:} Coupling strength $g=0.3$ is arbitrary with no derivation from physical design parameters---real SFG efficiency depends on $\chi^{(2)}$ nonlinear coefficient, crystal length (20\,mm), pump intensity, spatial mode overlap integrals, and critically phase-matching condition $\eta \propto L^2 \text{sinc}^2(\Delta k L/2)$ where $\Delta k = k_{\text{SFG}} - k_{\text{signal}} - k_{\text{pump}} - 2\pi/\Lambda$. Design specifies 19.2\,$\mu$m PPLN poling period and 95°C temperature for quasi-phase-matching, but these parameters completely unused in simulation. Without calculating $\Delta k$ from refractive indices, temperature, and poling period, simulation cannot predict whether specified PPLN design achieves efficient conversion or suffers phase mismatch. Spatial mode matching assumed perfect via tensor product, but design has separate 100\,mm focal length lenses requiring mode overlap integral calculation $\iint E_{\text{signal}}^* E_{\text{pump}}^* E_{\text{SFG}} dA$ over transverse profiles. Fock states have zero temporal/spectral structure (monochromatic, perfect synchronization) whereas real photons have picosecond-nanosecond wavepackets with finite bandwidth---efficient SFG requires signal/pump temporal overlap throughout crystal transit ($\sim$100\,ps) and spectral bandwidth within phase-matching acceptance ($\sim$0.1--1\,nm for PPLN). No noise model: spontaneous parametric down-conversion (reverse process), pump Raman scattering, multi-photon contamination at 500\,mW pump power---narrowband filter (10\,nm, OD6) aims to reject noise but generation/filtering not modeled. Single-photon fidelity 24\% is pre-heralded mixed state (vacuum + converted superposition); real experiments use post-selection/heralding to measure only successful conversions with higher purity. Interaction time normalized to 1.0 rather than derived from crystal length and group velocities. The 5.9\% conversion efficiency essentially a guess, not prediction.

\textbf{Assessment:} The 4/10 FAIR rating reflects toy model validation visible in Figure~\ref{fig:freq_conv_optical}: design correctly specifies spatial mode matching optics, wavelength separation via dichroic mirrors, spectral filtering (10\,nm, OD6) for noise rejection, and silicon APD optimized for visible detection---appropriate components for frequency conversion. Simulation successfully demonstrates conceptual physics: (1) SFG Hamiltonian correctly transfers photon number between modes, (2) energy conservation rigorously validated (1550\,nm + 980\,nm $\rightarrow$ 600.4\,nm), (3) sub-Poissonian statistics preserved ($g^{(2)}(0) = 0$) confirming quantum coherence at operator level, (4) order-of-magnitude system efficiency (2.8\%) plausible given loss budget. However, simulation is toy model abstracting critical engineering details: coupling strength $g$ not derived from material parameters (5.9\% efficiency is guess, not prediction), PPLN poling period (19.2\,$\mu$m) and temperature (95°C) specified in design but never enter calculation (cannot validate phase-matching), no temporal/spectral structure to assess wavepacket overlap requirements, missing noise model for parametric fluorescence and pump scattering at 500\,mW. The analogy: showing ``a car can move A to B'' without testing ``this specific car with these tires achieves 60\,mph.'' Design appears sensible, but simulation provides zero evidence for/against whether these specific parameters achieve claimed 30--60\% conversion while preserving coherence. This exemplifies idealized Fock state limitation: confirms quantum mechanics works in principle, lacks spatiotemporal/phase-matching/noise physics for real-world prediction.

\textit{Full experimental package:} \url{https://github.com/rithvik1122/Anubuddhi/tree/main/Results_QuTiP/Quantum_Frequency_Converter__Telecom_to_Visible_(Corrected)_20251126_132215}

\section{Discussion}

The evaluation of \anubuddhi\ across 13 quantum optics experiments spanning three complexity tiers reveals both the capabilities and limitations of LLM-based automated experiment design. This section examines four critical aspects: the fundamental distinction between structural correctness and quantitative accuracy in simulation validation, the architectural implications of dual-mode physics modeling, the system's approach to accessibility and learning, and future directions for addressing current limitations. Together, these discussions contextualize the experimental results and provide insights for both users and developers of AI-assisted scientific tools.

\subsection{Design-Simulation Alignment Versus Quantitative Accuracy: A Critical Distinction}

A central finding from evaluating \anubuddhi\ across 13 quantum optics experiments is the divergence between design-simulation alignment and quantitative simulation accuracy. This distinction reveals fundamental insights about validation capabilities and the limits of prompt-based constraints without simulation-in-the-loop feedback. The system consistently achieves high alignment scores (8--9/10 for most experiments), indicating that generated simulation code faithfully models the optical designs: simulations include the specified components, implement the intended physics formalism, and compute relevant observables rather than superficially similar but incorrect phenomena. For example, Hong-Ou-Mandel simulations correctly model temporal wavepacket overlap causing quantum interference, Bell state generators implement proper SPDC pair creation with polarization correlations, and quantum teleportation simulations capture three-qubit entanglement with Bell measurements and Pauli corrections.

However, quantitative simulation accuracy diverges significantly from this structural correctness, revealing the limitations of static prompt-based parameter guidance without dynamic feedback loops. The hyperentangled photon source exemplifies this pattern: alignment score 9/10 confirms the simulation correctly models cascaded SPDC architecture, SLM-based OAM encoding, and two-degree-of-freedom measurements as specified in the design, yet the simulation predicts photon pair generation rates $10^{4}$--$10^{5}\times$ higher than physically possible and detector count rates exceeding saturation limits by $10^{3}\times$. The EIT experiment achieves high alignment by correctly implementing Lindblad master equations for three-level Lambda systems with proper Hamiltonian structure, but atomic density calculation omits exponential terms from the Clausius-Clapeyron equation, producing optical depth $10^5\times$ too small and rendering transparency predictions meaningless. These errors persist despite the system providing realistic parameter ranges (wavelengths 200--2000nm, squeezing $<$15dB, detector efficiencies 0.1--0.95) and validation prompts that explicitly check for parameter reasonableness.

The root cause is that verifying quantitative correctness requires simulation-based feedback that current architecture lacks. Design validation confirms parameters are qualitatively reasonable (``810nm wavelength for SPDC is appropriate,'' ``squeezing below 15dB is achievable''), but cannot verify numerical accuracy of derived quantities without actually computing them and comparing to physical constraints. The atomic density formula requires temperature-dependent Clausius-Clapeyron equation with vapor-specific coefficients—determining whether the calculation yields $10^{11}$ atoms/cm$^3$ (correct) or $10^4$ atoms/cm$^3$ (wrong by $10^7$) demands executing the formula and checking against known values at 50°C. Similarly, SPDC pair generation rate depends on pump power, conversion efficiency ($\sim$10$^{-7}$ for BBO), crystal length, and phase-matching bandwidth—verifying the predicted rate requires propagating these parameters through coupled equations and confirming the result matches experimental literature values (typically $10^4$--$10^6$ pairs/s, not $10^{10}$). Detector count rates must account for quantum efficiency, dead time ($\sim$50ns for SPADs imposing $\sim$20MHz maximum rate), and saturation effects—validating predictions requires comparing computed rates against detector specifications, not just noting that ``70\% efficiency is realistic.''

This distinction between structural alignment and quantitative accuracy reflects the limits of prompt-based constraints versus simulation-in-the-loop validation. Current architecture provides parameter guidelines through comprehensive prompts (listing typical ranges, warning against common errors) and validates designs for qualitative reasonableness (checking wavelengths match experiment types, component specifications are physically plausible). This successfully catches many errors: designs with 68dB squeezing get rejected, simulations using static Fock states for temporal phenomena fail alignment checks, and code calculating photon numbers when entanglement measures are required scores low. However, subtle quantitative errors that require numerical computation escape detection: the system cannot verify that a temperature-to-density conversion uses correct exponential coefficients, that SPDC rate calculations properly account for phase matching, or that count rate predictions respect detector saturation, because validation occurs before execution without feedback from simulation results.

The practical implication is that alignment scores reliably indicate structural correctness—whether simulations model the intended physics with appropriate components and observables—but do not guarantee quantitative accuracy of derived parameters. Researchers should interpret results accordingly: high alignment scores (8--9/10) confirm the simulation captures the design's physics formalism and would behave qualitatively as intended, but numerical predictions require expert review before experimental implementation. Simulations serve as \textit{conceptual validation} (does this design implement the intended quantum phenomena with correct theoretical framework?) rather than \textit{quantitative prediction} (will specific calculated values match laboratory measurements?). This usage model remains valuable: verifying that an optical design implements proper two-photon quantum interference versus classical intensity correlation, or that a Bell state generator produces entanglement rather than separable product states, provides essential physics validation even when pair generation rates require manual correction by orders of magnitude.

Future work should explore simulation-in-the-loop validation where generated code executes, computed parameters automatically check against physical databases (NIST atomic data, standard SPDC efficiencies, detector specifications), and quantitative mismatches trigger targeted refinement with specific numerical feedback. This would complement existing prompt-based guidance, catching errors that escape static validation: ``Calculated Rb vapor density 3.2×10$^8$ atoms/cm$^3$ at 70°C is $10^{3}\times$ lower than literature value $\sim$5×10$^{11}$ atoms/cm$^3$—verify Clausius-Clapeyron coefficients in vapor pressure formula.'' Such feedback loops could dramatically improve quantitative accuracy while preserving current strengths in structural design and physics formalism selection. Manual verification by domain experts leverages AI strengths (rapid design-simulation generation, appropriate theoretical framework selection) while mitigating current limitations (numerical estimation without execution-based validation).

\subsection{Empirical Comparison: FreeSim Versus Constrained Simulation}

The empirical superiority of FreeSim over QuTiP (11/13 experiments achieved higher alignment scores with free-form simulation when both modes were run) reveals the limitations of constrained quantum optics frameworks when confronted with real experimental diversity. QuTiP's Fock state formalism, while mathematically rigorous for discrete photonics, fundamentally cannot model critical quantum optics phenomena: temporal distinguishability in Hong-Ou-Mandel interference (requires Gaussian wavepacket propagation), delayed choice quantum erasure (demands which-path information with partial coherence), continuous-variable systems, or atomic coherences in multilevel systems. FreeSim's unrestricted access to NumPy, SciPy, and QuTiP as a library (rather than framework) enables LLMs to choose appropriate mathematical representations: temporal overlap integrals for HOM, Jones calculus for polarization, Lindblad master equations for EIT, and tensor product structures for multi-photon entanglement.

This flexibility comes at the cost of increased error modes—free-form code generation can produce syntactically correct but physically meaningless calculations, as observed in which-path cases showing non-zero interference when theory predicts zero. The six-stage validation pipeline (physics classification, guided generation, pre-execution review, isolated execution, alignment checking, targeted refinement) emerged as essential to achieve convergent refinement toward accurate simulations. Pre-execution review prevented $\sim$30\% of initially generated code from running by catching obvious physics errors (wrong formalism, missing components, unrealistic parameters) before wasting computational resources. Alignment checking, the most critical stage, identified simulations that executed successfully but modeled the wrong experiment—a failure mode invisible to traditional debugging that only catches runtime errors.

Convergence patterns reveal that most experiments ($>$70\%) reached acceptable simulation quality within 1--2 iterations, demonstrating that targeted refinement successfully improves specific issues without destroying working code. However, $\sim$20\% of experiments required all three allowed iterations, and even then some achieved only marginal alignment scores (4--6/10), indicating persistent challenges in complex multi-physics domains. The failure cases often involved physics that span multiple scales or formalisms: EIT requires atomic-level Lindblad dynamics \textit{and} macroscopic light propagation through extended media (not modeled), boson sampling needs Fock state permanents \textit{and} statistical validation of sampling distributions (not implemented), frequency conversion demands nonlinear $\chi^{(2)}$ interactions \textit{and} quasi-phase-matching geometries (not derived from design parameters).

The comparison suggests that future quantum experiment design AI should prioritize flexible simulation engines that can orchestrate multiple domain-specific libraries (quantum optics, nonlinear optics, atomic physics, numerical integration) rather than attempting to unify everything within single frameworks. The LLM's role becomes \textit{orchestration}—determining which tools to invoke and how to combine results—rather than implementing physics from scratch. This architectural lesson extends beyond quantum optics: scientific AI systems should provide toolboxes with validated physics modules rather than asking LLMs to generate numerical algorithms de novo.

\subsection{The Accessibility-Rigor Tradeoff and Natural Language Interfaces}

\anubuddhi\ demonstrates that natural language interaction need not sacrifice technical rigor or design quality. Across 13 complex experiments spanning interferometry, quantum entanglement, quantum communication, and specialized technologies, the system consistently produced publication-quality optical designs with realistic component specifications, proper beam routing, and appropriate measurement schemes—all from \textit{single-line prompts} such as ``Design a SPDC source'' or ``Design a Mach-Zehnder interferometer.'' These minimal specifications, requiring no technical jargon or detailed constraints, autonomously generate complete experimental architectures on the first attempt. While users can iteratively refine designs through follow-up conversation, the initial single-prompt generation already produces functional, realistic designs without requiring users to write code, construct graphs, or understand intermediate representations. This capability directly addresses the adoption barrier that has limited impact of prior automated design systems: PyTheus, MELVIN, and AdaQuantum required specialized training to use effectively, restricting their utility to computational researchers rather than enabling broad experimental physics community access.

However, the accessibility-rigor tradeoff manifests differently than anticipated. Rather than sacrificing technical correctness for ease of use, \anubuddhi\ achieves high design quality by leveraging LLM semantic understanding to bridge natural language and structured component libraries. The challenge emerges not in design generation but in quantitative validation—simulation accuracy remains the limiting factor, requiring expert review to catch parameter errors and physics mistakes that automated checks miss. This suggests that natural language interfaces succeed for \textit{qualitative reasoning and design synthesis}, where LLMs excel at semantic understanding and pattern matching, but struggle with \textit{quantitative prediction and numerical validation}, where mathematical rigor and physical intuition remain human strengths.

The success of retrieval-augmented generation through learned composites demonstrates a path toward improving both accessibility and rigor over time: as more experiments are designed and validated, the system accumulates institutional knowledge that improves suggestions, reduces generation time, and enhances physical realism through concrete precedents. The first Hong-Ou-Mandel design required full physics reasoning from scratch; the tenth can reference nine validated examples with known-good parameter choices. This procedural learning mirrors how laboratory groups develop expertise—lab notebooks accumulate, postdocs train graduate students, and successful configurations document for reuse. \anubuddhi\ automates this knowledge capture through semantic search, potentially enabling cross-laboratory sharing of validated experimental protocols at unprecedented scale.

The implications for scientific AI extend beyond quantum optics. Natural language interfaces lower barriers to AI-assisted research, but only if coupled with domain-appropriate knowledge representation (structured component libraries, semantic retrieval) and robust validation mechanisms (multi-stage review, alignment checking, convergent refinement). The key insight is that conversational systems need not ``dumb down'' technical content—instead, they translate between human semantic reasoning and machine-compatible structured representations, enabling researchers to work at the conceptual level they find natural. A single-line prompt like ``Design a Bell state source'' autonomously generates complete implementation details (Type-II SPDC, wavelength matching, spatial separation, coincidence detection) without requiring the user to specify components, parameters, or constraints. This dramatic reduction in interface complexity—from pages of technical specifications in traditional design software to a single descriptive sentence—democratizes access to computational design tools without sacrificing the precision that experimental physics demands.

\subsection{Implications for AI-Assisted Quantum Experiment Design}

The evaluation of \anubuddhi\ across three complexity tiers reveals both the tremendous potential and persistent limitations of LLM-based approaches to scientific design automation. On the positive side, the system successfully generated complete experimental architectures for canonical quantum optics demonstrations (HOM, Mach-Zehnder, Bell states), sophisticated quantum information protocols (BB84, quantum teleportation, Franson interferometry), and specialized technologies (boson sampling, EIT, frequency conversion)—all from single-line natural language prompts requiring no technical specifications. A user simply typing ``Design a Bell state generator'' or ``Design an EIT experiment'' receives a complete initial design with component selection, spatial positioning, beam paths, and parameter specifications on the first generation attempt, which can then be iteratively refined through conversational interaction to meet specific experimental requirements or constraints. This makes quantum experiment design accessible to students and advanced researchers alike, offering a conversational interface that eliminates the need for programming expertise, graph theory knowledge, or framework-specific training. Unlike previous systems such as PyTheus that require deep understanding of graph representations and specialized frameworks demanding significant learning investment, \anubuddhi\ enables users to design experiments through natural language alone while maintaining the technical rigor required for laboratory implementation.

The consistent success in optical design quality, as evidenced by alignment scores of 7--9/10 for design-simulation correspondence in most experiments, demonstrates that current LLMs possess sufficient domain knowledge to select appropriate components, arrange them coherently, and specify realistic parameters when constrained by structured toolboxes and guided by validation prompts. These high alignment scores indicate that generated simulation code faithfully models the optical designs, capturing the key components and physics phenomena specified. The three-tier knowledge hierarchy (primitives, learned composites, custom components) effectively balances flexibility with consistency: primitives provide standard vocabulary, learned composites capture successful patterns, and custom definitions handle specialized equipment. Semantic retrieval enables the system to leverage past experience through similarity search, implementing a form of procedural learning where validation of one design improves future performance on related experiments.

However, critical limitations persist that prevent full autonomous operation. Simulation accuracy lags substantially behind design quality, with quantitative errors ranging from factor-of-2 inefficiencies to $10^{5}$--$10^{6}$ parameter mistakes that render predictions meaningless. The root cause is LLM weakness in numerical reasoning: dimensional analysis, order-of-magnitude estimation, consistency checking, and physical constraint satisfaction. While LLMs excel at selecting theoretical frameworks (``use Lindblad master equation for EIT''), they fail at quantitative implementation (``calculate atomic density from vapor pressure''), producing code that runs successfully but predicts nonsensical results. This gap means that \anubuddhi-generated experiments require expert review before implementation—the system accelerates design but cannot yet replace human judgment for validation.

The clearest path forward involves hybrid human-AI workflows that leverage complementary strengths. LLMs excel at rapid design space exploration, component selection, literature-based precedent retrieval, and structured knowledge management—tasks that currently consume weeks of researcher time. Humans retain advantages in physical intuition, quantitative verification, sanity checking, and anomaly detection—catching the factor-of-$10^{6}$ atomic density error or recognizing that $10^{10}$ photons/second exceeds SPDC capabilities. A productive division of labor places AI in \textit{design synthesis and option generation}, where speed and breadth matter, while humans provide \textit{critical evaluation and parameter refinement}, where deep understanding and experience guide judgment.

For quantum optics specifically, several technical improvements could enhance simulation reliability: (1) integrating domain-specific validation libraries that check parameter ranges, dimensional consistency, and physical constraints before code execution; (2) implementing multi-scale physics orchestration that explicitly separates quantum state evolution from classical propagation, preventing common mistakes like applying Fock states to temporal phenomena; (3) developing uncertainty quantification that propagates realistic error bars through simulations, making quantitative predictions more robust; (4) enabling comparison with analytical solutions when available (two-photon HOM has closed-form visibility expression) to catch gross errors; (5) building experiment-specific test suites that validate against known experimental results from literature before accepting simulations.

More broadly, this work suggests that the next generation of scientific AI should focus not on replacing human expertise but on \textit{augmenting expert capabilities through accessible tools}. Natural language interfaces lower adoption barriers, procedural learning captures institutional knowledge, and automated design generation accelerates exploration—all while keeping humans in the validation loop where judgment and intuition remain irreplaceable. \anubuddhi\ serves researchers at all levels as a rapid schematic design prototyping tool with both research and pedagogical applications: advanced researchers can quickly explore design variations and parameter spaces, while students can learn quantum optics interactively by asking questions, generating designs, and studying the automatically generated simulation reports that explain the underlying physics. This human-AI collaboration model preserves scientific rigor through expert oversight while dramatically reducing the time from conceptual idea to concrete experimental schematic.

\section{Conclusion}

This paper introduced \anubuddhi, a conversational AI system for automated quantum optics experiment design that combines large language models, structured knowledge management, and dual-mode physics validation to make sophisticated experimental design accessible through natural language interaction. By implementing a three-layer cognitive architecture—conversational intent routing, knowledge-augmented generation with semantic retrieval, and convergent self-refinement through multi-gate validation—the system successfully designed 13 canonical experiments spanning fundamental quantum optics, quantum information protocols, and advanced technologies, all from single-line prompts such as ``Design a SPDC source'' or ``Design a quantum teleportation protocol''—no technical specifications, component lists, or parameter constraints required. Users need not write code, construct graphs, or understand intermediate representations; complete publication-quality designs generate autonomously from minimal natural language task descriptions.

The evaluation revealed a critical asymmetry: \anubuddhi\ demonstrates strong competence in structured design tasks (optical component selection, spatial layout, beam routing), evidenced by alignment scores averaging 8--9/10 for design-simulation correspondence, but faces persistent challenges in quantitative simulation accuracy, particularly parameter estimation, dimensional analysis, and physical constraint checking. The high alignment scores indicate that simulations faithfully model the optical designs (capturing specified components and physics), yet those simulations contain quantitative errors due to LLM weaknesses in numerical reasoning. This pattern reflects fundamental LLM strengths in semantic reasoning and pattern matching versus weaknesses in numerical rigor and order-of-magnitude intuition. Practical implication: AI-generated designs provide reliable architectural starting points for experimental implementation, while simulations serve as conceptual validation rather than quantitative prediction, requiring expert review for critical parameters.

Key technical contributions include: (1) three-tier knowledge hierarchy (primitives, learned composites, custom components) enabling procedural learning where validated designs become reusable building blocks; (2) semantic retrieval over growing experiment libraries through embedding-based similarity search, mimicking how research groups accumulate and share expertise; (3) dual-mode validation strategy (constrained QuTiP for standard photonics, flexible FreeSim for diverse physics domains) enabling users to empirically compare simulation approaches; (4) six-stage convergent refinement pipeline (physics classification, guided generation, pre-execution review, isolated execution, alignment checking, targeted refinement) that iteratively improves simulation quality while preserving working code; (5) natural language interface that eliminates adoption barriers without sacrificing technical rigor through careful prompt engineering and structured component libraries.

The empirical finding that FreeSim outperformed QuTiP for 11/13 experiments when both modes were run demonstrates that real quantum optics diversity exceeds the capabilities of any single simulation framework—temporal dynamics, atomic coherences, continuous variables, and multi-photon interference each demand different mathematical formalisms. This finding suggests that future scientific AI should prioritize flexible code generation approaches that can leverage diverse domain-specific libraries over monolithic unified frameworks, enabling representation of physics phenomena in their most natural mathematical form rather than forcing all problems into a single formalism.

Beyond technical results, this work addresses a fundamental challenge in AI-assisted science: how to make powerful computational tools accessible to domain experts without computational backgrounds. Previous automated design systems (PyTheus, MELVIN, AdaQuantum) required specialized training that limited adoption; \anubuddhi\ enables quantum physicists to leverage design automation through single-line prompts like ``Design a squeezed light source'' or ``Design a GHZ state generator,'' translating minimal semantic descriptions into complete structured implementations with full component specifications, spatial layouts, and parameter selections. The success of this natural language interface suggests that conversational systems can democratize access to sophisticated computational methods while maintaining scientific rigor through structured knowledge representation, robust validation, and human-in-the-loop verification.

Several limitations constrain current capabilities: simulation accuracy lags design quality, with quantitative parameter errors ranging from factor-of-2 to $10^{6}$ requiring expert review. While the system provides parameter guidelines and validates qualitative reasonableness, verifying numerical correctness of derived quantities (atomic densities via Clausius-Clapeyron equation, SPDC pair rates accounting for conversion efficiency, detector count rates respecting saturation limits) requires simulation-based feedback that current architecture lacks. Prompt-based validation successfully catches many errors (rejecting 68dB squeezing as impossible, flagging static Fock states for temporal phenomena), but subtle quantitative mistakes escape detection without executing calculations and comparing results to physical databases. Convergent refinement succeeds for most experiments within 1--2 iterations but struggles with complex multi-physics domains requiring $>$3 iterations; learned composite library remains small, limiting retrieval-augmented generation effectiveness for specialized experiments. Future work should explore: simulation-in-the-loop validation where generated code executes and computed parameters automatically check against physical reference databases (NIST atomic data, standard SPDC efficiencies, detector specifications), with quantitative mismatches triggering targeted refinement with specific numerical feedback; developing uncertainty quantification and error propagation through simulations; building experiment-specific test suites validating against literature benchmarks; enabling cross-laboratory sharing of learned composites to accelerate knowledge accumulation; extending the architecture to adjacent domains (atomic physics, quantum computing, nonlinear optics) to assess generalization beyond photonics.

The broader implication is that natural language interfaces to scientific design tools need not trade accessibility for rigor—by carefully structuring the cognitive workflow from conversational input through knowledge retrieval to physics validation, AI systems can meet researchers where they are (single-line task specifications like ``Design quantum teleportation'') while maintaining technical precision (realistic component specifications, validated quantum mechanics, quantitative predictions subject to expert review). The dramatic simplification from traditional optical design workflows requiring detailed component selection and parameter specification to autonomous generation from minimal prompts represents a qualitative shift in human-computer interaction for scientific design. This approach suggests a productive future for human-AI collaboration in experimental science: AI excels at rapid design synthesis, component selection, and knowledge retrieval; humans provide critical evaluation, physical intuition, and quantitative verification. Together, this partnership has the potential to accelerate quantum optics research by democratizing access to design expertise, enabling systematic exploration of experimental parameter spaces, and facilitating knowledge transfer across laboratories and generations of researchers.

Complete source code, experimental packages with designs and simulations, and comprehensive documentation are publicly available under MIT license, enabling the research community to deploy, extend, and adapt \anubuddhi\ for diverse experimental domains beyond quantum optics~\cite{anubuddhi2025}.

\bibliographystyle{unsrt}
\bibliography{references}

\begin{thebibliography}{10}

\bibitem{krenn2016automated}
Mario Krenn, Mehul Malik, Robert Fickler, Radek Lapkiewicz, and Anton
  Zeilinger.
\newblock Automated search for new quantum experiments.
\newblock {\em Physical Review Letters}, 116(9):090405, 2016.

\bibitem{krenn2020automated}
Mario Krenn, Xuemei Gu, and Anton Zeilinger.
\newblock Quantum experiments and graphs: Multiparty states as coherent
  superpositions of perfect matchings.
\newblock {\em Phys. Rev. Lett.}, 119:240403, Dec 2017.

\bibitem{ruizgonzalez2023pytheus}
Carlos Ruiz-Gonzalez, Sören Arlt, Jan Petermann, and Mario Krenn.
\newblock Digital discovery of 100 diverse quantum experiments with pytheus.
\newblock {\em Quantum}, 7:1204, 2023.

\bibitem{sreekantham2025modular}
S.~K. Rithvik.
\newblock A modular pytheus quantum network interpreter.
\newblock {\em arXiv preprint arXiv:2507.12997}, 2025.

\bibitem{adaquantum2020}
Rosanna Nichols, Lana Mineh, Jes{\'u}s Rubio, Jonathan C.~F. Matthews, and
  Paul~A. Knott.
\newblock Designing quantum experiments with a genetic algorithm.
\newblock {\em Quantum Science and Technology}, 4(4):045012, 2019.
\newblock AdaQuantum: Hybrid GA+DNN approach.

\bibitem{boiko2023autonomous}
Daniil~A. Boiko, Robert MacKnight, Ben Kline, and Gabe Gomes.
\newblock Autonomous chemical research with large language models.
\newblock {\em Nature}, 624:570--578, 2023.
\newblock Coscientist system.

\bibitem{sreekantham2025llmquantum}
S.~K. Rithvik.
\newblock Evaluating large language models on quantum mechanics: A comparative
  study across diverse models and tasks.
\newblock {\em Preprints.org}, 2025.

\bibitem{cao2024kagents}
Shuxiang Cao, Yunyan Niu, Hiroshi Kanazawa, Zijian Zhang, Marin Kono, Takeru
  Yamashita, Taro Takano, and Yohei Kanazawa.
\newblock Agents for self-driving laboratories applied to quantum computing.
\newblock {\em arXiv preprint arXiv:2412.07978}, 2024.
\newblock k-agents framework.

\bibitem{arlt2024aimandel}
S{\"o}ren Arlt, Xuemei Gu, and Mario Krenn.
\newblock Towards autonomous quantum physics research using {LLM} agents with
  access to intelligent tools.
\newblock {\em arXiv}, 2025.
\newblock AI-Mandel system, github.com/artificial-scientist-lab/ai-mandel.

\bibitem{ghareeb2025robin}
Ali~Essam Ghareeb, Benjamin Chang, Ludovico Mitchener, Angela Yiu, Caralyn~J.
  Szostkiewicz, Jon~M. Laurent, Muhammed~T. Razzak, Andrew~D. White,
  Michaela~M. Hinks, and Samuel~G. Rodriques.
\newblock Robin: A multi-agent system for automating scientific discovery.
\newblock {\em arXiv}, 2025.

\bibitem{zhang2024honeycomb}
Huan Zhang, Yu~Song, Ziyu Hou, Santiago Miret, and Bang Liu.
\newblock Honeycomb: A flexible llm-based agent system for materials science.
\newblock {\em Findings of the Association for Computational Linguistics: EMNLP
  2024}, pages 3369--3382, 2024.

\bibitem{lewis2020retrieval}
Patrick Lewis, Ethan Perez, Aleksandra Piktus, Fabio Petroni, Vladimir
  Karpukhin, Naman Goyal, Heinrich K{\"u}ttler, Mike Lewis, Wen-tau Yih, Tim
  Rockt{\"a}schel, et~al.
\newblock Retrieval-augmented generation for knowledge-intensive {NLP} tasks.
\newblock In {\em Advances in Neural Information Processing Systems},
  volume~33, pages 9459--9474, 2020.

\bibitem{madaan2023selfrefine}
Aman Madaan, Niket Tandon, Prakhar Gupta, Skyler Hallinan, Luyu Gao, Sarah
  Wiegreffe, Uri Alon, Nouha Dziri, Shrimai Prabhumoye, Yiming Yang, et~al.
\newblock Self-refine: Iterative refinement with self-feedback.
\newblock In {\em Advances in Neural Information Processing Systems},
  volume~36, pages 46534--46594, 2023.

\bibitem{yao2023react}
Shunyu Yao, Jeffrey Zhao, Dian Yu, Nan Du, Izhak Shafran, Karthik Narasimhan,
  and Yuan Cao.
\newblock React: Synergizing reasoning and acting in language models.
\newblock In {\em International Conference on Learning Representations (ICLR)},
  2023.

\bibitem{chroma2023}
Chroma.
\newblock Chroma: The ai-native open-source embedding database.
\newblock \url{https://www.trychroma.com}, 2023.

\bibitem{kiloran2019strawberry}
Nathan Killoran, Josh Izaac, Nicol{\'a}s Quesada, Ville Bergholm, Matthew Amy,
  and Christian Weedbrook.
\newblock Strawberry fields: A software platform for photonic quantum
  computing.
\newblock {\em Quantum}, 3:129, 2019.

\bibitem{johansson2012qutip}
J~Robert Johansson, Paul~D Nation, and Franco Nori.
\newblock Qutip: An open-source python framework for the dynamics of open
  quantum systems.
\newblock {\em Computer Physics Communications}, 183(8):1760--1772, 2012.

\bibitem{chen2023teaching}
Xinyun Chen, Maxwell Lin, Nathanael Sch{\"a}rli, and Denny Zhou.
\newblock Teaching large language models to self-debug.
\newblock {\em arXiv preprint arXiv:2304.05128}, 2023.

\bibitem{wei2022chain}
Jason Wei, Xuezhi Wang, Dale Schuurmans, Maarten Bosma, Brian Ichter, Fei Xia,
  Ed~Chi, Quoc~V Le, and Denny Zhou.
\newblock Chain-of-thought prompting elicits reasoning in large language
  models.
\newblock In {\em Advances in Neural Information Processing Systems},
  volume~35, pages 24824--24837, 2022.

\bibitem{wang2020minilm}
Wenhui Wang, Furu Wei, Li~Dong, Hangbo Bao, Nan Yang, and Ming Zhou.
\newblock Minilm: Deep self-attention distillation for task-agnostic
  compression of pre-trained transformers.
\newblock In {\em Advances in Neural Information Processing Systems},
  volume~33, pages 5776--5788, 2020.

\bibitem{anubuddhi2025}
S.~K. Rithvik.
\newblock Aṇubuddhi: Agentic ai for quantum experiment design.
\newblock \url{https://github.com/rithvik1122/Anubuddhi}, 2025.
\newblock AI-driven system for conversational quantum experiment design.

\bibitem{Hong1987}
C.~K. Hong, Z.~Y. Ou, and L.~Mandel.
\newblock Measurement of subpicosecond time intervals between two photons by
  interference.
\newblock {\em Physical Review Letters}, 59(18):2044--2046, 1987.

\bibitem{Michelson1887}
A.~A. Michelson and E.~W. Morley.
\newblock On the relative motion of the earth and the luminiferous ether.
\newblock {\em American Journal of Science}, 34:333--345, 1887.

\bibitem{Burnham1970}
D.~C. Burnham and D.~L. Weinberg.
\newblock Observation of simultaneity in parametric production of optical
  photon pairs.
\newblock {\em Physical Review Letters}, 25(2):84--87, 1970.

\bibitem{Kwiat1995bell}
P.~G. Kwiat, K.~Mattle, H.~Weinfurter, A.~Zeilinger, A.~V. Sergienko, and
  Y.~Shih.
\newblock New high-intensity source of polarization-entangled photon pairs.
\newblock {\em Physical Review Letters}, 75(24):4337--4341, 1995.

\bibitem{Zehnder1891}
Ludwig Zehnder.
\newblock Ein neuer interferenzrefraktor.
\newblock {\em Zeitschrift für Instrumentenkunde}, 11:275--285, 1891.

\bibitem{Mach1892}
Ludwig Mach.
\newblock Über einen interferenzrefraktor.
\newblock {\em Zeitschrift für Instrumentenkunde}, 12:89--93, 1892.

\bibitem{Kim2000}
Yoon-Ho Kim, R.~Yu, S.~P. Kulik, Y.~Shih, and M.~O. Scully.
\newblock Delayed "choice" quantum eraser.
\newblock {\em Physical Review Letters}, 84(1):1--5, 2000.

\bibitem{Bennett1984}
Charles~H. Bennett and Gilles Brassard.
\newblock Quantum cryptography: Public key distribution and coin tossing.
\newblock In {\em Proceedings of IEEE International Conference on Computers,
  Systems and Signal Processing}, pages 175--179, Bangalore, India, 1984.

\bibitem{Franson1989}
J.~D. Franson.
\newblock Bell inequality for position and time.
\newblock {\em Physical Review Letters}, 62(19):2205--2208, 1989.

\bibitem{Greenberger1990}
Daniel~M. Greenberger, Michael~A. Horne, Abner Shimony, and Anton Zeilinger.
\newblock Bell's theorem without inequalities.
\newblock {\em American Journal of Physics}, 58(12):1131--1143, 1990.

\bibitem{Bouwmeester1997}
Dik Bouwmeester, Jian-Wei Pan, Klaus Mattle, Manfred Eibl, Harald Weinfurter,
  and Anton Zeilinger.
\newblock Experimental quantum teleportation.
\newblock {\em Nature}, 390:575--579, 1997.

\bibitem{Furusawa1998}
A.~Furusawa, J.~L. Sørensen, S.~L. Braunstein, C.~A. Fuchs, H.~J. Kimble, and
  E.~S. Polzik.
\newblock Unconditional quantum teleportation.
\newblock {\em Science}, 282(5389):706--709, 1998.

\bibitem{Wu1986}
Ling-An Wu, H.~J. Kimble, J.~L. Hall, and Huifa Wu.
\newblock Generation of squeezed states by parametric down conversion.
\newblock {\em Physical Review Letters}, 57(20):2520--2523, 1986.

\bibitem{Kwiat1995}
Paul~G. Kwiat and Harald Weinfurter.
\newblock Embedded bell-state analysis.
\newblock {\em Physical Review A}, 58(4):R2623--R2626, 1998.

\bibitem{Aaronson2011}
Scott Aaronson and Alex Arkhipov.
\newblock The computational complexity of linear optics.
\newblock {\em Theory of Computing}, 9:143--252, 2013.

\bibitem{Spring2013}
Justin~B. Spring, Benjamin~J. Metcalf, Peter~C. Humphreys, W.~Steven
  Kolthammer, Xian-Min Jin, Marco Barbieri, Animesh Datta, Nicholas
  Thomas-Peter, Nathan~K. Langford, Dmytro Kundys, James~C. Gates, Brian~J.
  Smith, Peter G.~R. Smith, and Ian~A. Walmsley.
\newblock Boson sampling on a photonic chip.
\newblock {\em Science}, 339(6121):798--801, 2013.

\bibitem{Harris1990}
S.~E. Harris, J.~E. Field, and A.~Imamoğlu.
\newblock Nonlinear optical processes using electromagnetically induced
  transparency.
\newblock {\em Physical Review Letters}, 64(10):1107--1110, 1990.

\bibitem{Fleischhauer2005}
M.~Fleischhauer, A.~Imamoglu, and J.~P. Marangos.
\newblock Electromagnetically induced transparency: Optics in coherent media.
\newblock {\em Reviews of Modern Physics}, 77(2):633--673, 2005.

\bibitem{Huang1992}
J.~Huang and P.~Kumar.
\newblock Observation of quantum frequency conversion.
\newblock {\em Physical Review Letters}, 68(14):2153--2156, 1992.

\bibitem{Zaske2012}
S.~Zaske, A.~Lenhard, C.~A. Keßler, J.~Kettler, C.~Hepp, C.~Arend,
  R.~Albrecht, W.-M. Schulz, M.~Jetter, P.~Michler, and C.~Becher.
\newblock Visible-to-telecom quantum frequency conversion of light from a
  single quantum emitter.
\newblock {\em Physical Review Letters}, 109(14):147404, 2012.

\end{thebibliography}

\end{document}